\documentclass[12pt,onecolumn,journal,	draftclsnofoot]{IEEEtran}
\ifCLASSINFOpdf
 \usepackage[pdftex]{graphicx}
\else
\fi
 \usepackage{amsmath}
\usepackage{subfig}
\usepackage{pgfplots}
 \usepackage{amsfonts} 
\usepackage{rotating}
\usepackage{array}
\usepackage{epstopdf}
\usepackage{tikz} 
\usepackage{booktabs}
\usepackage{float}
\usepackage[final]{changes}
\usepackage{soul}
\usetikzlibrary{plotmarks}
\pgfplotsset{ 
	compat=1.7, 
	legend style = {font=\small\sffamily}, 
	label style = {font=\small\sffamily} 
}
\newcolumntype{C}{ >{\centering\arraybackslash} m{2cm} }
\newcolumntype{D}{ >{\centering\arraybackslash} m{0.2cm} }

\newlength\figureheight
\newlength\figurewidth

\begin{document}
\definechangesauthor[name={Milind Gide}, color=red]{MG}
\setremarkmarkup{(#2 -- #1)}
%
\title{A Locally Weighted Fixation Density-Based Metric for Assessing the Quality of Visual Saliency Predictions}
%
%
%

\author{Milind~S.~Gide       
        and~Lina~J.~Karam,~\IEEEmembership{Fellow,~IEEE}
\thanks{M.~S.~Gide and L.~J.~Karam are with the School
	of Electrical, Computer and Energy Engineering, Arizona State University, Tempe, AZ, 85287-5706 USA~(e-mail:~mgide,~karam@asu.edu)}
}

%
%

\markboth{Final version published in IEEE Transactions on Image Processing 2016}%
{Shell \MakeLowercase{\textit{et al.}}: Bare Demo of IEEEtran.cls for Journals}
%



\maketitle

\begin{abstract}
With the increased focus on visual attention (VA) in the last decade, a large number of computational visual saliency methods have been developed over the past few years. These models are traditionally evaluated by using performance evaluation metrics that quantify the match between predicted saliency and fixation data obtained from eye-tracking experiments on human observers. Though a considerable number of such metrics have been proposed in the literature, there are notable problems in them. In this work, we discuss shortcomings in existing metrics through illustrative examples and propose a new metric that uses local weights based on fixation density which overcomes these flaws. To compare the performance of our proposed metric  at assessing the quality of saliency prediction with other existing metrics, we construct a ground-truth subjective database in which saliency maps obtained from 17 different VA models are evaluated by 16 human observers on a 5-point categorical scale in terms of their visual resemblance with corresponding ground-truth fixation density maps obtained from eye-tracking data. The metrics are evaluated by correlating metric scores with the human subjective ratings. The correlation results show that the proposed evaluation metric outperforms all other popular existing metrics. Additionally, the constructed database and corresponding subjective ratings provide an insight into which of the existing metrics and future metrics are better at estimating the quality of saliency prediction and can be used as a benchmark. 
\end{abstract}


%


%
\IEEEpeerreviewmaketitle

\section{Introduction}
%
%
%
%
\IEEEPARstart{V}{isual} attention (VA) is the broad area of research that aims to explain the mechanisms by which the human visual system (HVS) filters the vast amount of visual information captured by the retina.
VA has applications in a large number of diverse areas like object recognition, image segmentation, compression, selective reduced-power visual processing, to name a few. As a result, there has been a lot of focus recently on developing computational VA models.  The VA mechanism is considered to be a combination of instantaneous pre-attentive, bottom-up processes that depend on low-level cues, and much slower, top-down, cognitive processes that depend on high-level precepts. Most of the existing models are based on bottom-up concepts and output what is known as a saliency map that gives the visual importance of each pixel location. Given the large number of VA models to choose from, it is necessary to evaluate these models. VA models are traditionally evaluated by comparing the saliency maps with eye-tracking data that is obtained from human observers. Several performance metrics that objectively quantify the match between the predicted saliency map and eye-tracking data have been introduced over the past decade for evaluating VA models (see ~\cite{borjireview} for a detailed review). A number of studies like those by Toet~\cite{Toet}, Borji et al.~\cite{borjieval}, and Judd et al.~\cite{judd2012benchmark} have evaluated state-of-the-art VA models using a subset of these metrics. However, none of these studies have evaluated the effectiveness of existing metrics in assessing the quality of VA models and ignore certain notable flaws in them.  The motivation for the proposed metric is to provide a more accurate assessment of the quality of visual saliency prediction than existing metrics, which can aid in a better comparative evaluation of VA models.  The proposed metric can also act as an improved measure of cost for training VA models that use machine learning.  Yet another application for the proposed metric is faithful objective comparison of eye-tracking equipments. 
Given the importance of accurate evaluation of VA models, there have been a few papers in recent years that discuss metrics. LeMeur and Baccino~\cite{LeMeur2013} gave an overview of existing performance metrics in literature and discussed their strengths and weaknesses. Riche et al.~\cite{riche2013saliency} provided a taxonomy for existing metrics and also studied the correlation between the metrics. They showed that each metric alone is not sufficient to evaluate a VA model and suggested the use of a combination of metrics to get a better estimate of performance.  Recently, Gide et al.~\cite{MilindSamTPAMI} discussed known flaws in existing metrics  through examples and proposed a metric, $sNSS$, that resolves the center-bias problem in the Normalized Scanpath Saliency ($NSS$) metric~\cite{NSS} through shuffling.  However, none of these works provide a common benchmark to compare the performance metrics. 

The key contribution of this paper is to propose a novel metric that assigns locally adaptive weights to fixation points based on local fixation density and thus gives more importance to the visually relevant fixations in the ground-truth eye-tracking data.  We also address the problem of a lack of a benchmark for evaluating existing and future performance metrics by constructing a subjective database in which ratings on a 5-point categorical scale by human observers are used to rate saliency maps of several VA models based on their visual resemblance to ground-truth saliency maps. The average ratings or mean opinion scores (MOS) are then correlated with the performance metric scores to evaluate the metrics.

This paper is organized as follows. In Section~\ref{sec:shortcomings} we highlight the known problems in existing popular metrics~\cite{mit-saliency-benchmark} through illustrative examples.  We then propose a new metric that uses locally adaptive weights for fixation points  in  Section~\ref{sec:proposedmetrics}. The details of the subjective study are presented in Section~\ref{sec:subjectivestudy}, and the correlation results for the existing and proposed metrics are presented in Section~\ref{sec:results}. Finally, we conclude the paper in Section~\ref{sec:conclusion} and also provide directions for future research. 

\section{Existing metrics and their shortcomings}
\label{sec:shortcomings}
\begin{table}
	\centering
	\footnotesize
	\caption{Evaluated Metrics.}
	\begin{tabular}{|c|c|c|}
		\hline
		Metric Name & Category & Ground-truth \\
		\hline
		$AUC_{Borji}$~\cite{borjieval} & Location-based & Fixation Points \\
		$AUC_{Judd}$~\cite{judd2012benchmark}  & Location-based & Fixation Points\\
		$sAUC$~\cite{borjieval} & Location-based & Fixation Points \\
		$WF_\beta$~\cite{margolin2014evaluate} & Location-based & Fixation Density Map \\
		$NSS$~\cite{borjieval}&Value-based & Fixation Points \\
		$sNSS$~\cite{MilindSamTPAMI}&Value-based & Fixation Points \\
		$CC$~\cite{borjieval}& Distribution-based & Fixation Density Map \\
		$SIM$~\cite{judd2012benchmark}& Distribution-based & Fixation Density Map\\
		$EMD$~\cite{judd2012benchmark}& Distribution-based & Fixation Density Map \\     
		\added[id=MG]{$MAE$~\cite{MAE}} & \added[id=MG]{Distribution-based}& \added[id=MG]{Fixation Density Map}\\
		\hline
	\end{tabular}
	\label{tab:existingmetrics}
\end{table}

Existing metrics can be classified into the following three major categories: value-based, location-based, and distribution-based, depending on the type of similarity measure used to compare the predicted saliency map to the eye-tracking data~\cite{riche2013saliency}. The value-based metrics focus on the predicted saliency map values at fixation points, the location-based metrics focus on how well the salient regions in the predicted saliency maps match with the locations of the fixation points, and the distribution-based metrics focus on the differences in the statistical distributions of the predicted saliency maps and fixation points. In addition to these categories, metrics can also be  classified based on the type  of ground-truth used. Some metrics use only the fixation locations from the eye-tracking data whereas others use a ground-truth saliency map (GSM) which is obtained by convolving a 2D Gaussian with the fixations and normalizing the resulting map. \added[id=MG]{Several recent studies have used different types metrics to benchmark VA models.  Toet~\cite{Toet} evaluated several VA models by using the Spearman's correlation coefficient. More recently, Borji et al.~\cite{borjieval} used the $AUC_{Borji}$, $CC$ and $NSS$ measures and Judd et al.~\cite{judd2012benchmark} used the $AUC_{Judd}$, $Similarity$ and $EMD$ metrics to evaluate several VA models. The MIT saliency benchmark project~\cite{mit-saliency-benchmark} is an up-to-date online benchmarking resource that lists the performance of all the recent state-of-the-art VA models using seven popular evaluation metrics that are a combination of those used in~\cite{borjieval} and~\cite{judd2012benchmark}.} The metrics used by the MIT Saliency Benchmark~\cite{mit-saliency-benchmark} along with recently proposed metrics  $WF_{\beta}$~\cite{margolin2014evaluate} and $sNSS$~\cite{MilindSamTPAMI}  \added[id=MG]{in addition to a baseline metric $MAE$~\cite{MAE}} are listed in Table~\ref{tab:existingmetrics} along with  the categories they belong to, as well as the type of ground-truth used.  
\begin{figure}[t]
	\centering
	\footnotesize
	\begin{tabular}{cccc}
		\includegraphics[width=0.2\textwidth]{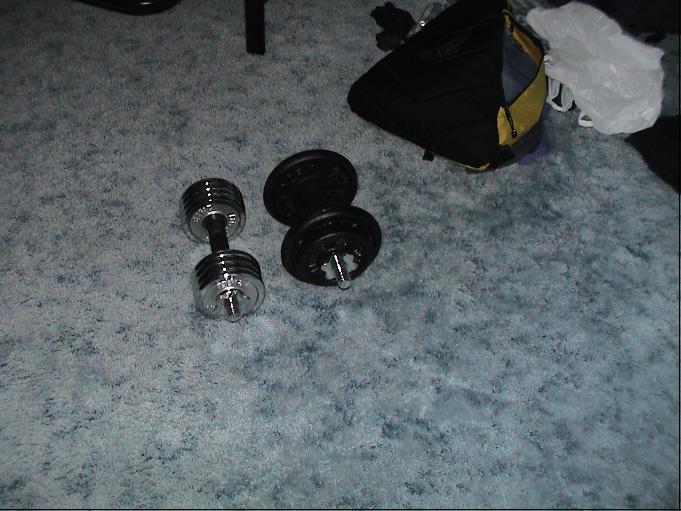} &
		\includegraphics[width=0.2\textwidth]{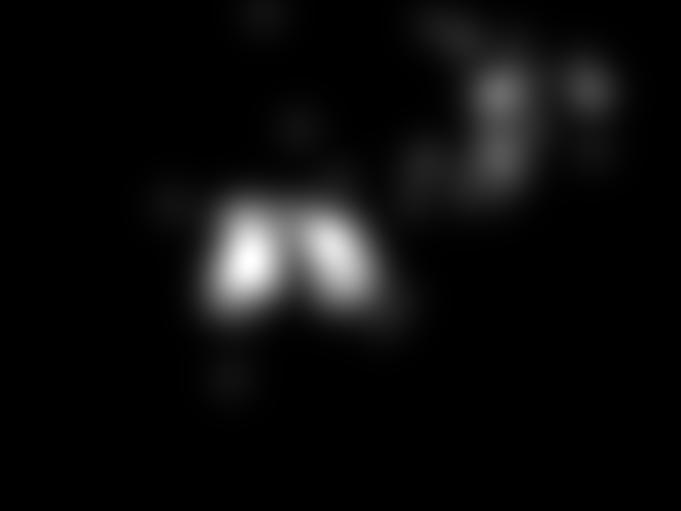} &
		\includegraphics[width=0.2\textwidth]{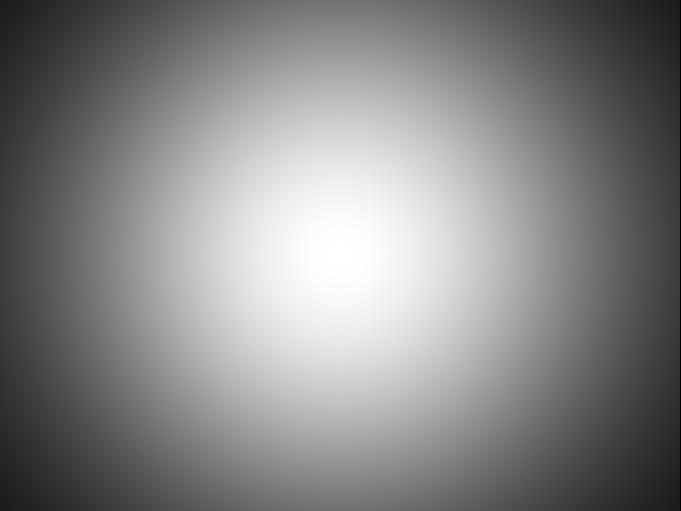} & 
		\includegraphics[width=0.2\textwidth]{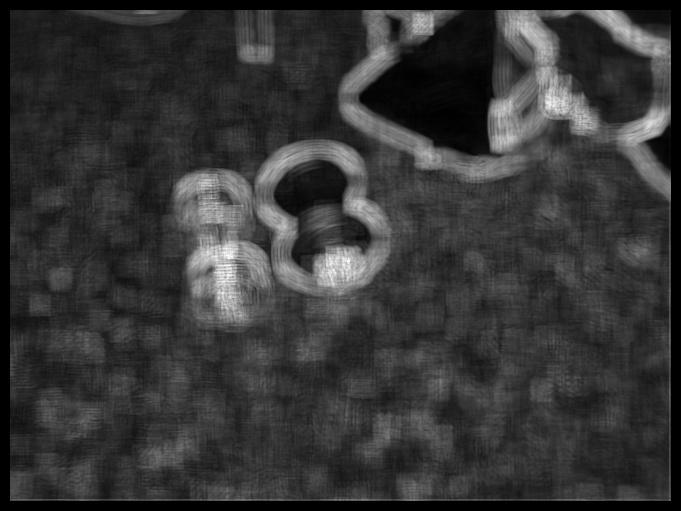} \\
		\textbf{(a)} Original Image &
		\textbf{(b)} Ground Truth & 
		\textbf{(c)} Centered Gaussian &
		\textbf{(d)} AIM~\cite{AIM} 
	\end{tabular}
	\footnotesize
	\begin{tabular}{|c|c c|}
		\hline
		Non-shuffled Metrics & (c) & (d) \\ \hline
		$AUC_{Borji}$~\cite{borjieval} & 0.8195 & 0.7509 \\	
		$AUC_{Judd}$~\cite{Judd_2012} & 0.8264 & 0.7760 \\
		$CC$~\cite{Judd_2012} & 0.4522 & 0.4359 \\ 
		$SIM$~\cite{Judd_2012} & 0.3415 &  0.3398 \\
		$EMD$~\cite{Judd_2012} &  3.2479 & 3.5294 \\
		$NSS$~\cite{NSS} & 1.2347  & 1.3745 	 \\
		\hline
		Shuffled metrics & (c) & (d) \\	\hline		
		$sAUC$~\cite{borjieval} & 0.5979 & 0.6689 \\
		$sNSS$~\cite{MilindSamTPAMI} &  0.4853 & 0.8911\\
		Proposed $sWNSS$ &  0.2144  & 0.9104  \\
		
		\hline			
	\end{tabular}
	\caption{Center bias problem in existing metrics that is rectified by the shuffled metrics. For $EMD$, a lower score indicates better performance; for the other metrics, a higher score indicates better performance.}
	\label{fig:CenterBias}
\end{figure}

\added[id=MG]{The first notable and well-analyzed problem with existing metrics is the problem of center-bias~\cite{borjireview,MilindSamTPAMI,LeMeur2013}. This problem arises due to an inherent tendency of images and photographs to contain objects of interest in central regions as compared to peripheral regions.  Most metrics that do not factor the center-bias in their formulation tend to incorrectly reward models that independent of content assign higher importance to central regions and lower importance to peripheral regions. One way of tackling this issue is through ``shuffling'' in which ground-truth fixations for all other images in the dataset are randomly sampled and high saliency predictions at such locations are penalized. Consequently, models that blindly reward central regions are penalized to a greater extent by the shuffling process, and receive a much lower score than more discriminative models~\cite{borjireview}. An illustration of the effect of center-bias on shuffled and non-shuffled metrics is shown in Figure~\ref{fig:CenterBias}. As shown in Figure~\ref{fig:CenterBias}, the}
\deleted[id=MG]{The} non-shuffled metrics like $AUC_{Borji}$, $AUC_{Judd}$, $CC$, $EMD$, $SIM$, and  $NSS$ tend to give higher scores to models that assign higher saliency to central regions as compared to the boundaries \deleted[id=MG]{(see Figure~\ref{fig:CenterBias})}. As a result, these incorrectly result in higher performance scores for a centered Gaussian blob (Figure~\ref{fig:CenterBias}(c)) as compared to a saliency map from a VA model (Figure~\ref{fig:CenterBias}(d)). \deleted[id=MG]{One solution to this problem is shuffling the metrics~\cite{SUN}. As seen in Figure~\ref{fig:CenterBias}} \added[id=MG]{On the other hand, the shuffled metrics} \deleted[id=MG]{$sAUC$ assigns}\added[id=MG]{assign} a better score to the AIM~\cite{AIM} saliency map over the centered Gaussian map. 
\begin{figure}[t]
	\centering
	\footnotesize
	\begin{tabular}{cccc}
		\includegraphics[width=0.2\textwidth]{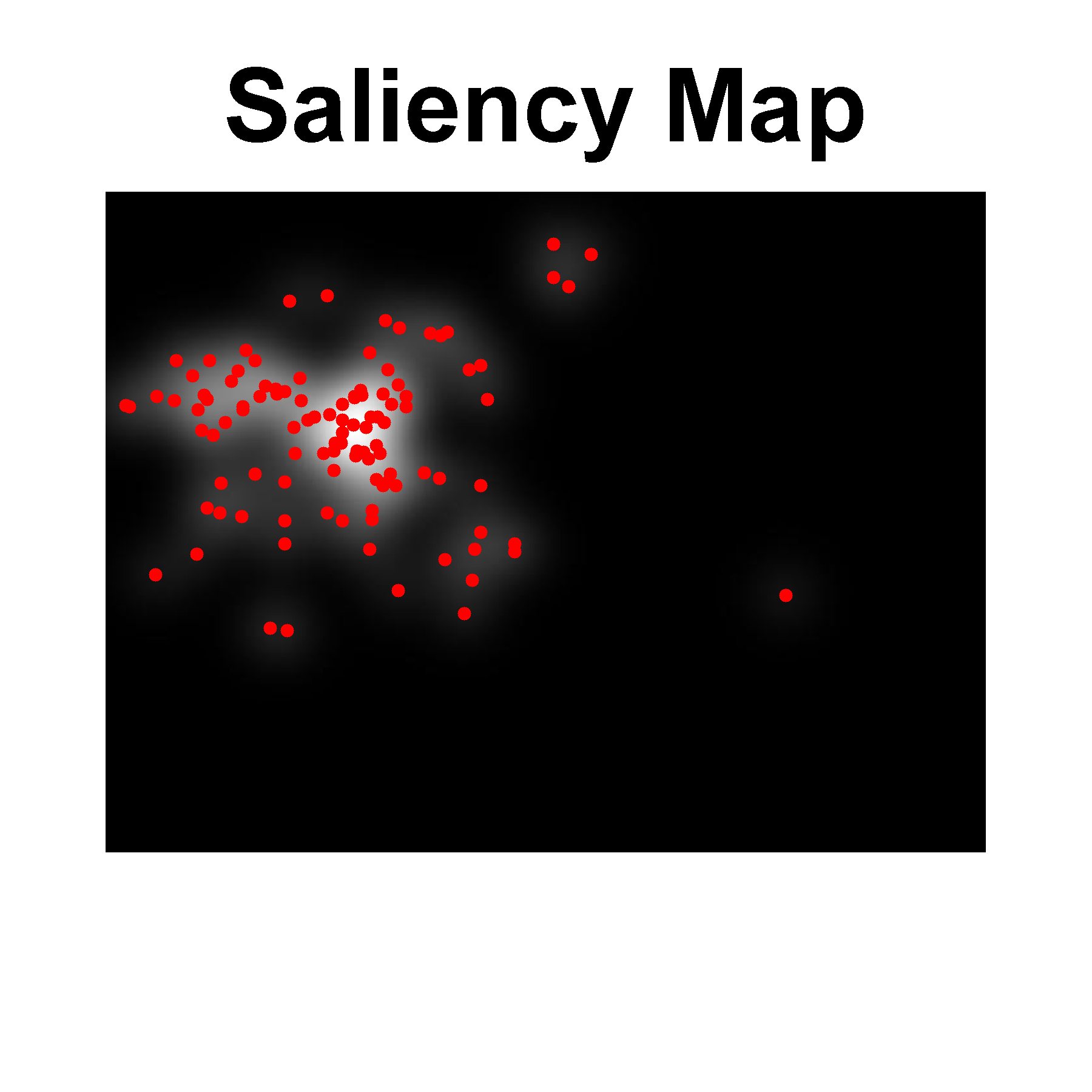} & 
		\includegraphics[width=0.2\textwidth]{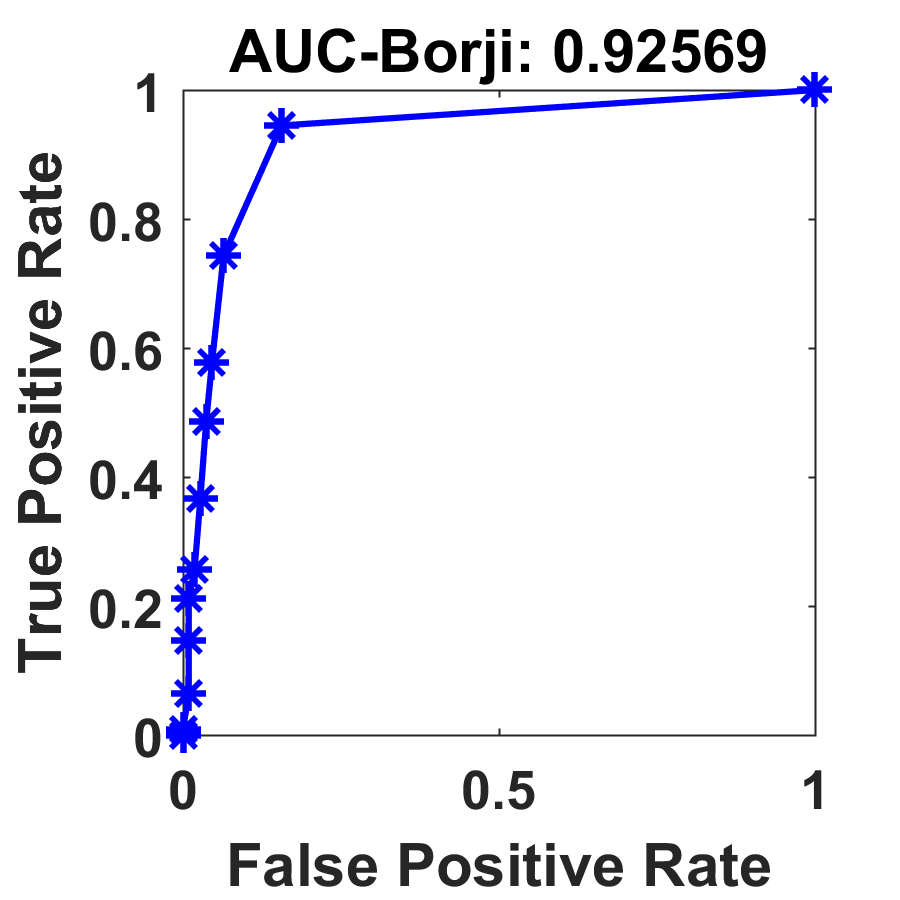} & 
		\includegraphics[width=0.2\textwidth]{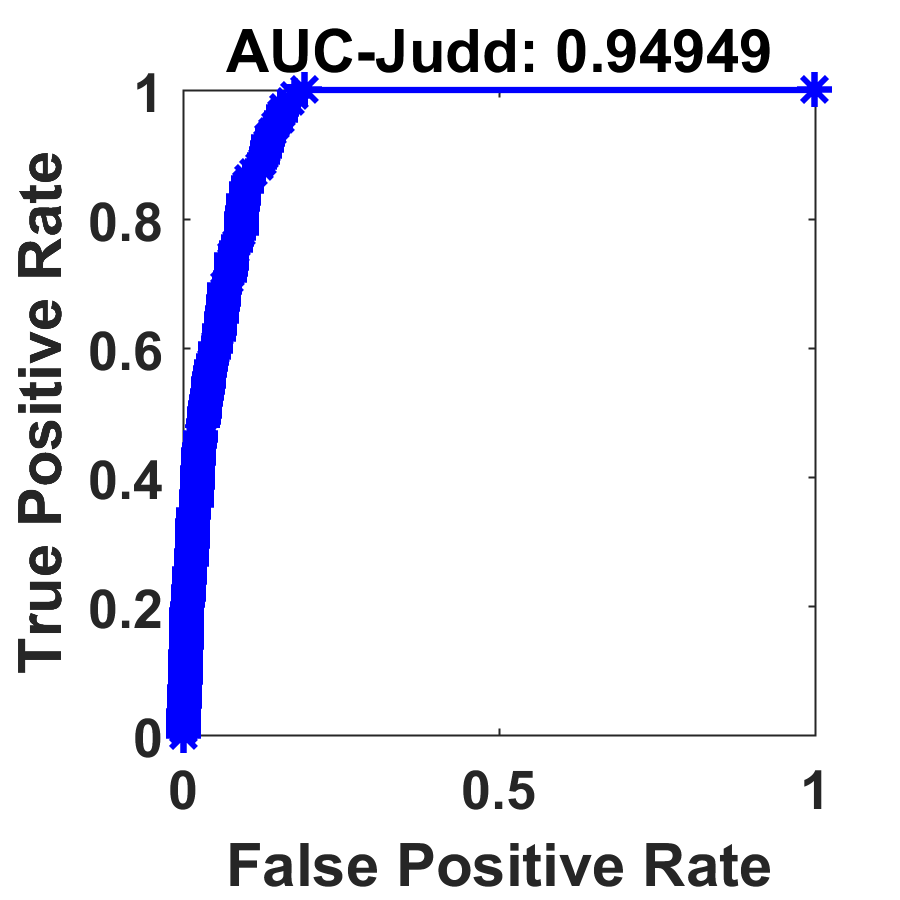} & 
		\includegraphics[width=0.2\textwidth]{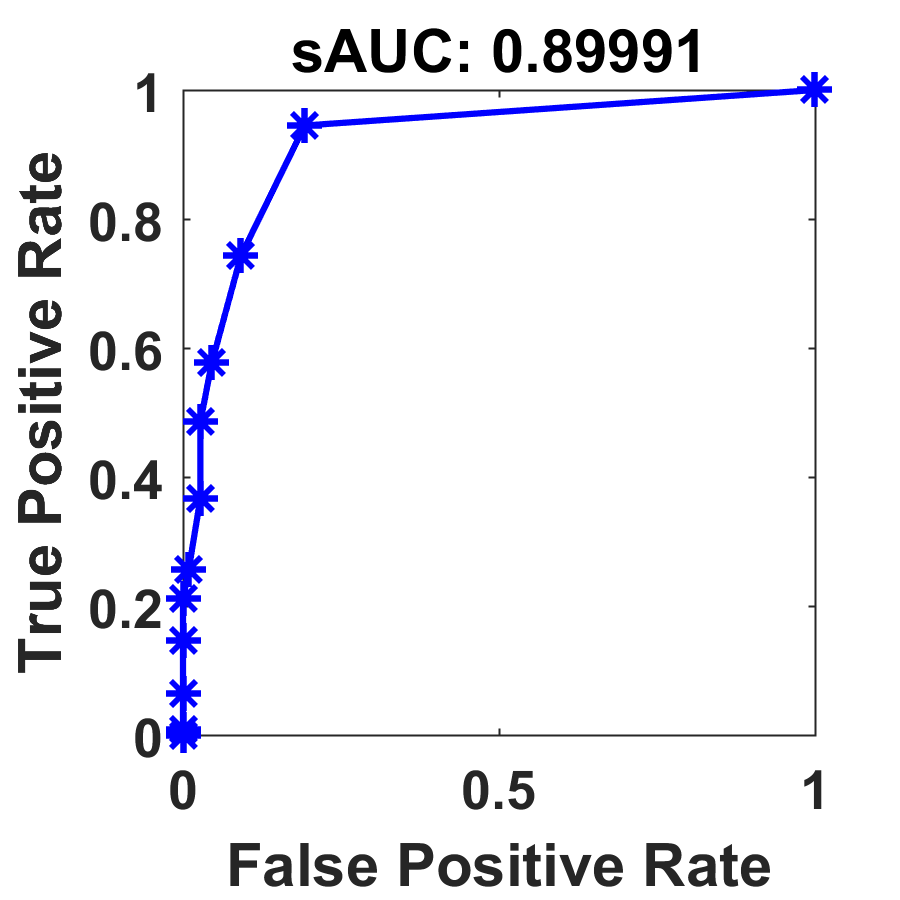}\\
		\includegraphics[width=0.2\textwidth]{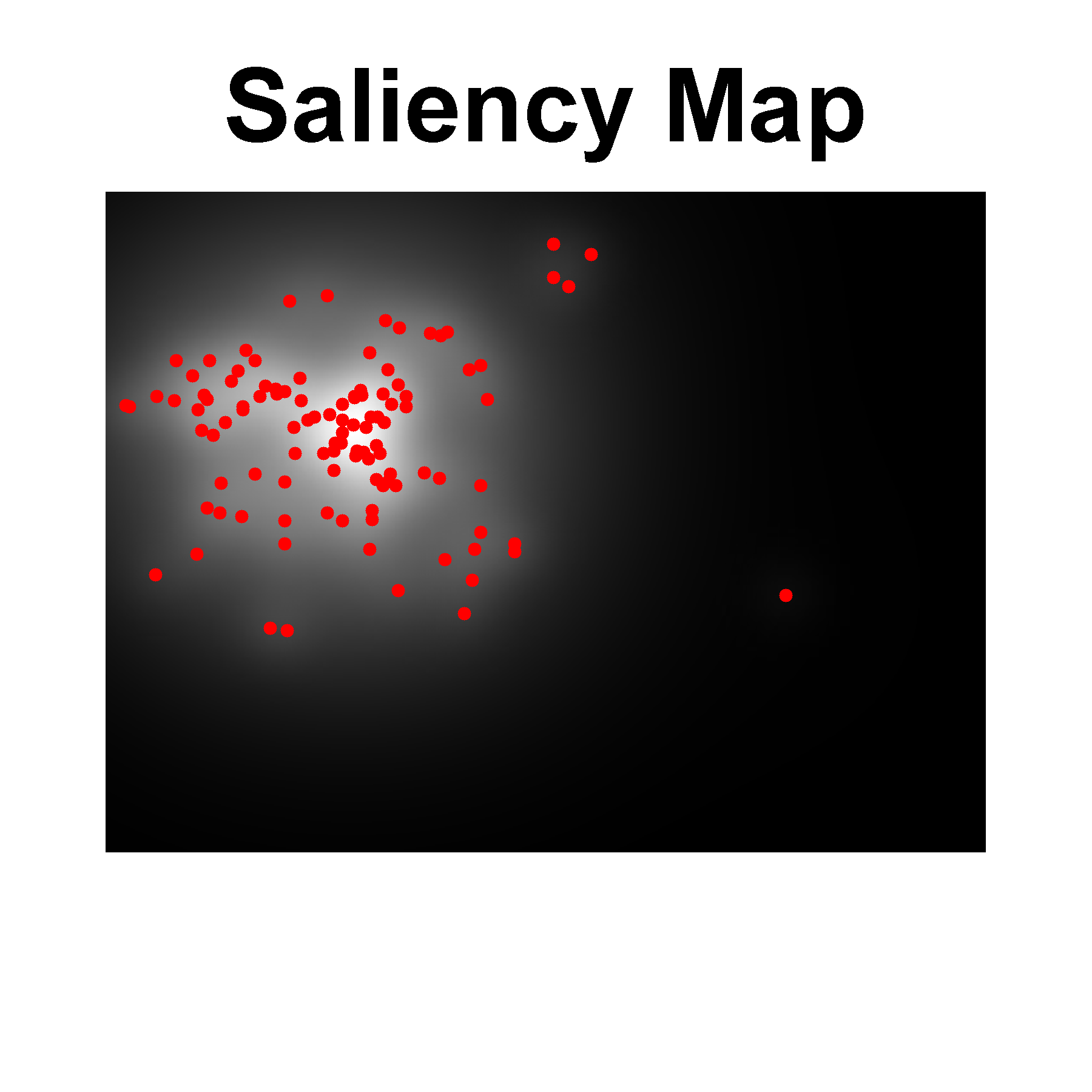} &
		\includegraphics[width=0.2\textwidth]{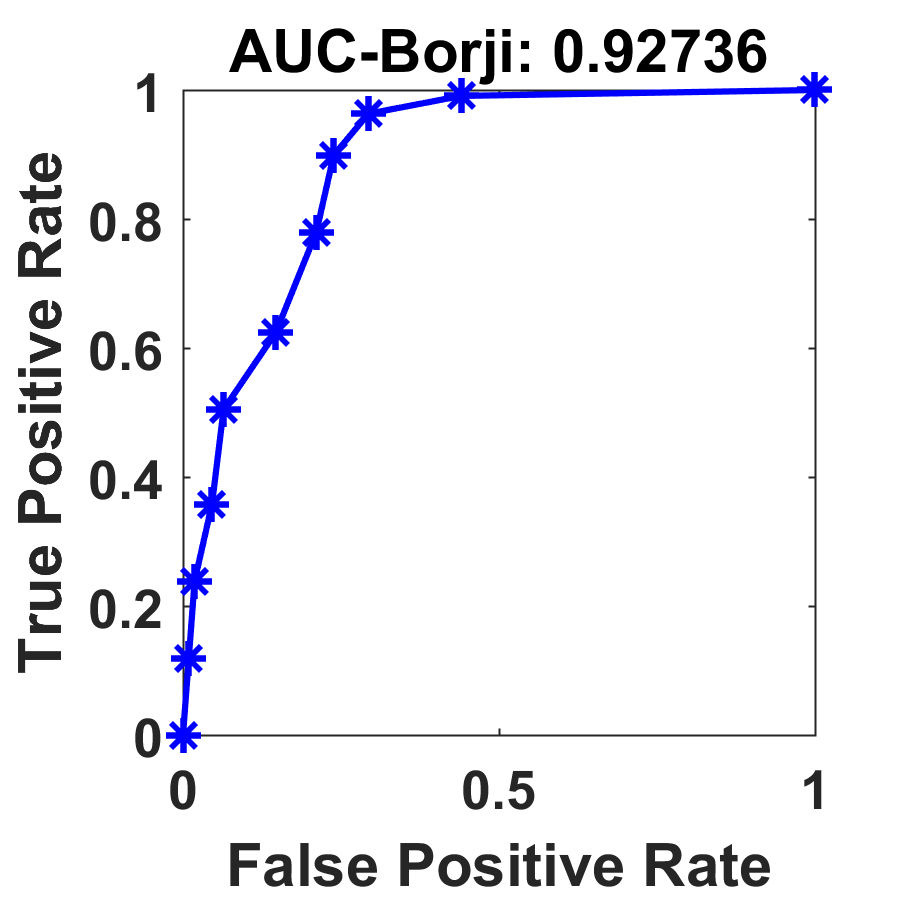} &
		\includegraphics[width=0.2\textwidth]{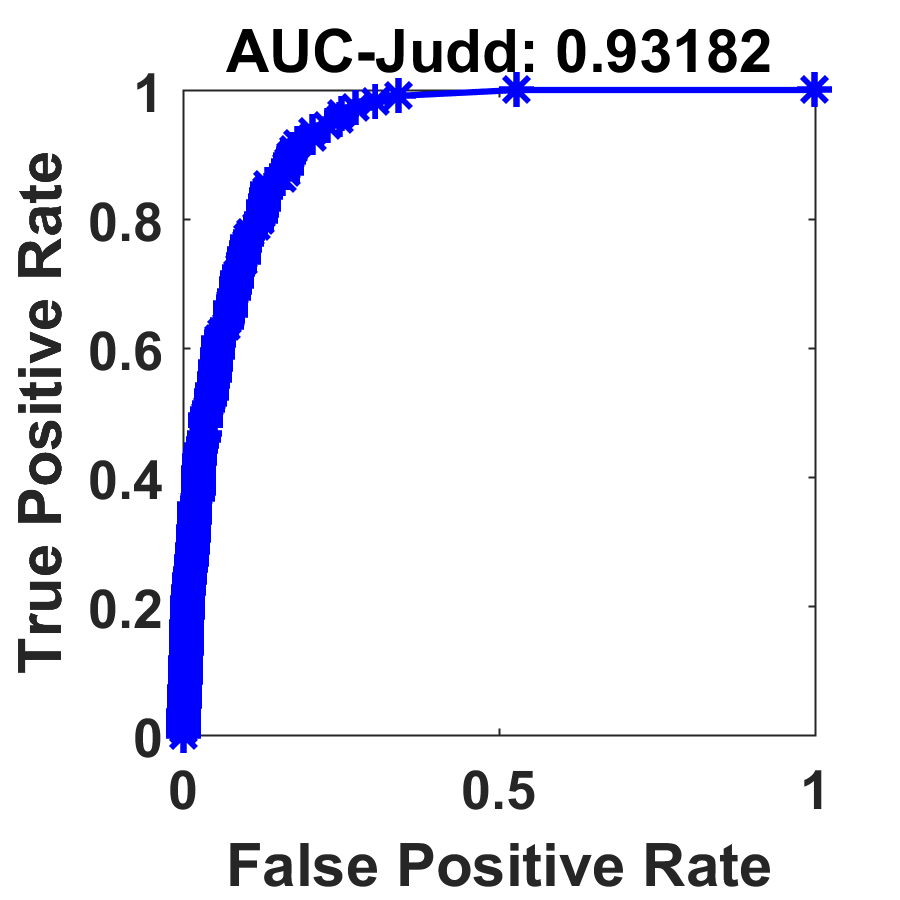} &
		\includegraphics[width=0.2\textwidth]{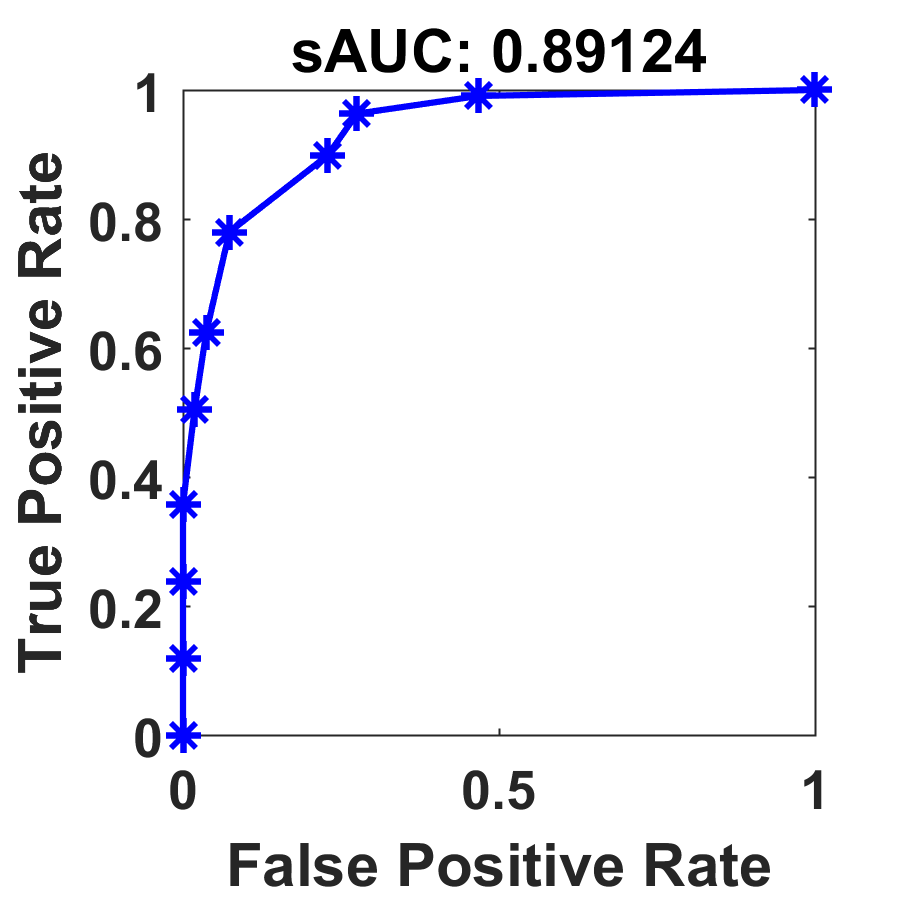} \\
		
	\end{tabular}
	\caption{AUC Interpolation flaw. Top Row: (Left to Right) Ground-truth saliency map with ground-truth fixations and corresponding ROC curves for $AUC_{Borji}$, $AUC_{Judd}$ and $sAUC$. Bottom Row: (Left to Right) Fuzzier version of the ground-truth saliency map with ground-truth fixations and corresponding ROC curves for $AUC_{Borji}$, $AUC_{Judd}$ and $sAUC$.}
	
	\label{fig:interpolationflaw}
\end{figure}

\begin{figure}[t]
	\centering
	\footnotesize
	\begin{tabular}{cc}
		\includegraphics[width=0.3\textwidth]{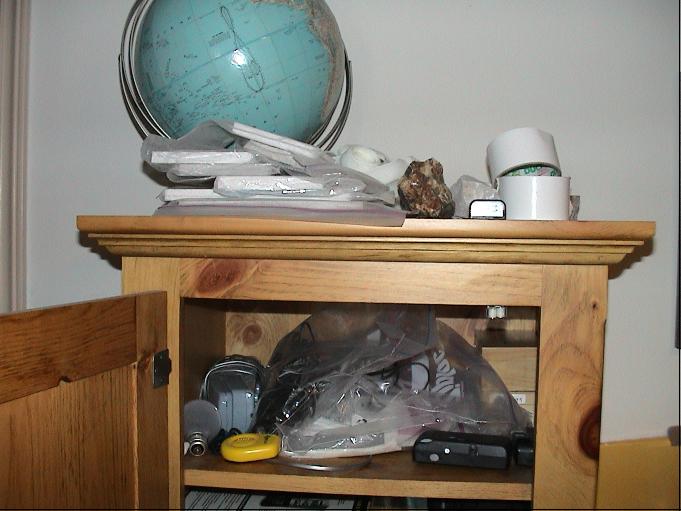} & 
		\includegraphics[width=0.3\textwidth]{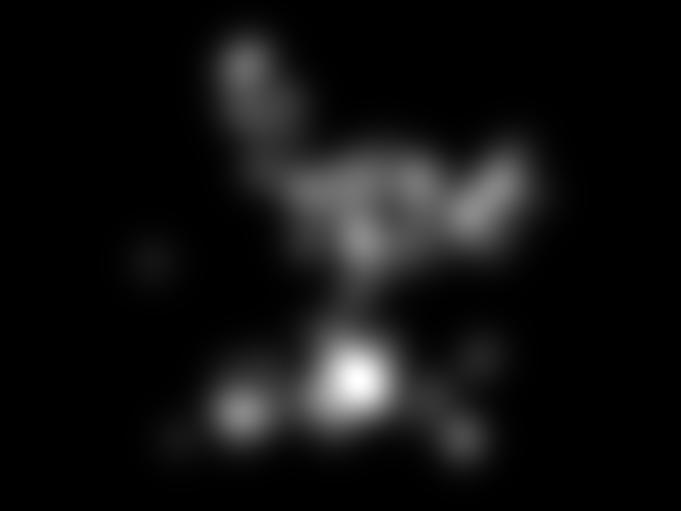} \\
		\textbf{(a)}Image &
		\textbf{(b)}Ground-truth	
	\end{tabular}
	\begin{tabular}{ccc}
		\includegraphics[width=0.3\textwidth]{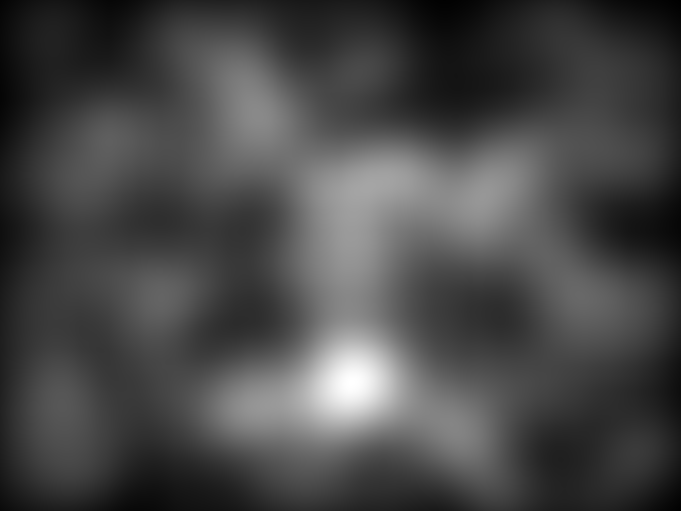} & 
		\includegraphics[width=0.3\textwidth]{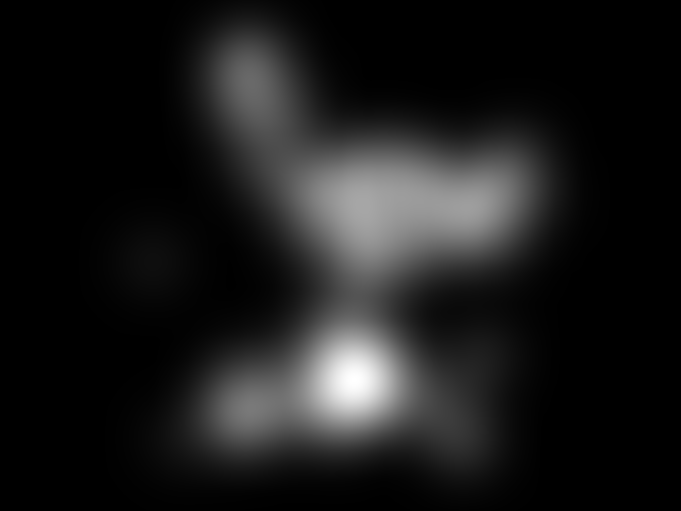} &
		\includegraphics[width=0.3\textwidth]{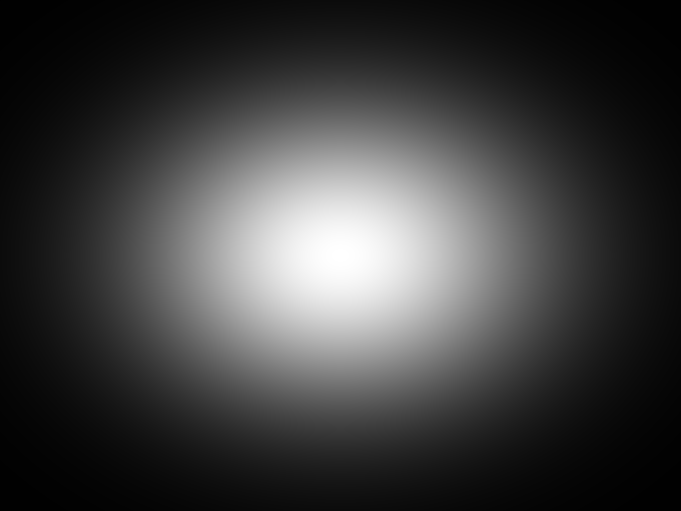} \\
		\textbf{(c)} &
		\textbf{(d)} & 
		\textbf{(e)} 
	\end{tabular}
	\footnotesize
	\begin{tabular}{| c | c c c |}
		\hline
		& (c) & (d) & (e)\\ 
		\hline
		$AUC_{Borji}$~\cite{borjieval} & 0.9104 &  0.9167 &  0.7955 \\
		$AUC_{Judd}$~\cite{Judd_2012} & 0.9164 & 0.9250 & 0.8030 \\
		$sAUC$~\cite{borjieval} & 0.8212 & 0.8113 &  0.4667 \\
		$NSS$~\cite{NSS}& 1.9630 & 2.2586 & 1.0794 \\
		Proposed $WNSS$ &     2.7984     &   3.1348   &    0.9889      \\
		$sNSS$~\cite{MilindSamTPAMI} &  1.3573 &  1.5864 & -0.1247 \\
		Proposed $sWNSS$ & 2.1901 & 2.5390  &  -0.0273 \\
		\hline		 			
	\end{tabular}
	\caption{The two saliency maps (c) and (d) have nearly identical $sAUC$, $AUC_{Borji}$ and $AUC_{Judd}$ scores, however it is clear that (d) is a much ``better'' saliency map.  $NSS$, $sNSS$  and the proposed $WNSS$ and $sWNSS$ metrics do not have this problem as they assign a significantly higher score to (d) than (c). A centered Gaussian blob (e) will perform well using $NSS$ and the proposed $WNSS$ metrics, however using the $sNSS$ and the proposed shuffled $sWNSS$ the same Gaussian blob receives a low score as expected.}
	\label{fig:snss}
\end{figure}

However, the AUC metrics including the $sAUC$ suffer from \added[id=MG]{another notable flaw} \deleted[id=MG]{what is} known as the interpolation flaw (described in detail in~\cite{margolin2014evaluate}). As seen in Figure~\ref{fig:interpolationflaw}, $AUC_{Borji}$, $AUC_{Judd}$ and $sAUC$ are less sensitive to false-positives. \added[id=MG]{As a result, a ``fuzzy'' ground-truth saliency map created by increasing the background activity in the neighborhood of a true-positive peak incorrectly gets higher or almost similar scores than the actual ground-truth saliency when using the AUC-based metrics.} \deleted[id=MG]{introduced by making the saliency map more fuzzy.}  
The other metrics $NSS$~\cite{NSS}, $CC$~\cite{borjieval}, $EMD$~\cite{mit-saliency-benchmark} and $SIM$~\cite{mit-saliency-benchmark} do not exhibit the interpolation flaw but do suffer from the center-bias problem as seen in Figure~\ref{fig:CenterBias}. 
Of these metrics, only $NSS$ is a viable candidate to be shuffled to tackle the center-bias issue as suggested in~\cite{MilindSamTPAMI}.  This metric termed Shuffled NSS or $sNSS$ for short is given by
\begin{equation}
sNSS = NSS(p) - NSS(r)
\label{eq:SNSS}
\end{equation}
\noindent where $p$ and $r$ denote, respectively, the ground-truth fixation points for the image and the randomly sampled non-fixation points from the set of fixation points for other images in the dataset and
\begin{equation}
NSS(x) = \frac{1}{N}\sum\limits_{x\in X}\frac{S(x) - \mu_s}{\sigma_s}.
\label{eq:NSS}
\end{equation}
In (\ref{eq:NSS}), $\mu_s$ and $\sigma_s$ are, respectively, the mean and standard deviation of the predicted saliency map $S$ and $N$ is the number of points in the set $X$. The random sampling for the non-fixation points $r$ is repeated a number of times, typically 100, and the final result is the average of scores obtained for each of these trials. The $sNSS$ metric improves on the $sAUC$ scores by correctly assigning a better score to the saliency map in Figure~\ref{fig:snss}(d) as compared to the one in Figure~\ref{fig:snss}(c). It also improves upon $NSS$ by giving the centered Gaussian map in Figure~\ref{fig:snss}(e) a low score. 

\begin{figure}[t]
	\centering
	\footnotesize
 \resizebox{\textwidth}{!}{
	\begin{tabular}{ccccc}
		\includegraphics[width=0.18\textwidth]{figures/AIMSAUC_failcase.jpg} & 
		\includegraphics[width=0.18\textwidth]{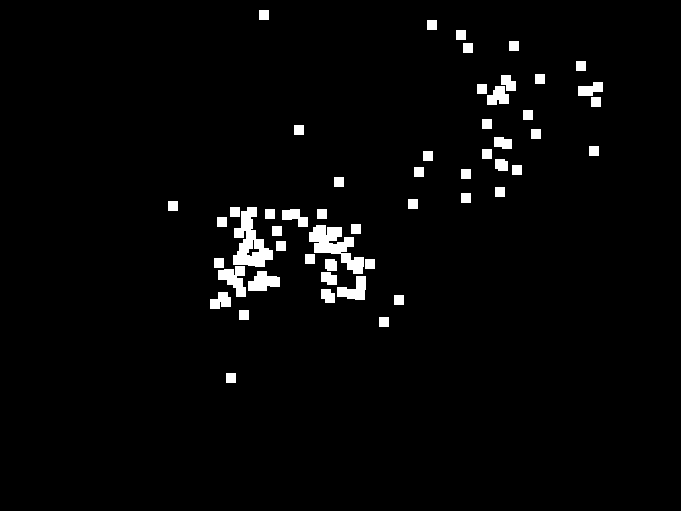} & 
		\includegraphics[width=0.18\textwidth]{figures/AIMfdm3.jpg} &
		\includegraphics[width=0.18\textwidth]{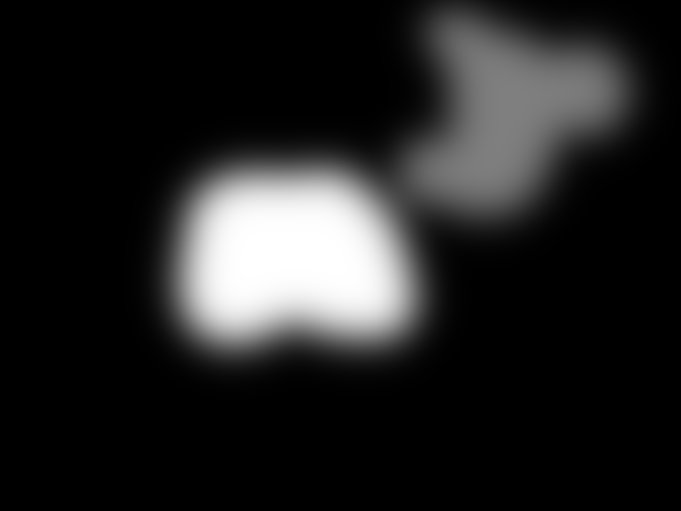} & 
		\includegraphics[width=0.18\textwidth]{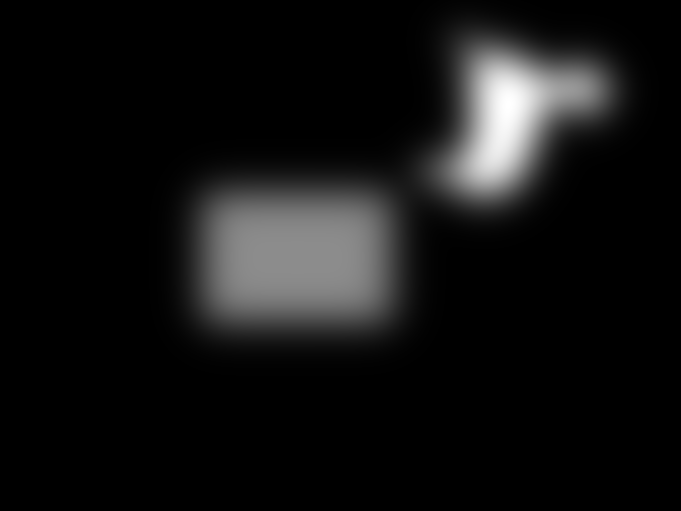} \\ 		
		\textbf{(a)} \added[id=MG]{Original Image} & 
		\textbf{(b)} \added[id=MG]{Fixation Map} & 
		\textbf{(c)} \added[id=MG]{Ground-Truth Saliency} &	\textbf{(d)}\added[id=MG]{ Saliency Map 1} &  
		\textbf{(e)}\added[id=MG]{ Saliency Map 2} \\	
	\end{tabular}}
	\footnotesize
	\begin{tabular}{|c|cc|}
		\hline
		Metric & (d) & (e) \\
		\hline
		$NSS$~\cite{NSS} & 2.4245 & 2.4605 \\ 
		$sNSS$~\cite{MilindSamTPAMI}   & 1.4715 &  1.8223  \\
		Proposed $WNSS$ &  3.1503
		&2.5531 \\
		Proposed $sWNSS$  & 2.2327 & 1.9541 \\
		\hline		
	\end{tabular}
	\caption{Problem with $NSS$ and $sNSS$:  The $NSS$ and $sNSS$ metrics give equal weight to every fixation point and ignore density. As a result, they incorrectly give higher scores to Saliency Map 2 (e) as compared to  Saliency Map 1 (d). The proposed $WNSS$ and its shuffled variant $sWNSS$  weight the fixations based on their local density and assign a higher score to map (d) as expected.}	
	\label{fig:NSSFlaw}	
\end{figure} 
For the $sAUC$, the locations used for determining false-positives are sampled from the distribution of fixations for all other images.  Because of the center-bias inherent in most eye-tracking datasets, these locations tend to be in the central portion of the image. As a result, if false-positives crop up in regions away from the center, $sAUC$ is not able to penalize them. In contrast, because of the zero-mean unit-standard deviation normalization in $sNSS$, blurrier maps are penalized as a result of which $sNSS$ is able to correctly assign a lower score to fuzzy maps such as map (c) in Figure~\ref{fig:snss}. However, a drawback of the  $NSS$ and $sNSS$ metrics is that in their computation all fixations are given equal weights and fixation density is ignored.  Figure~\ref{fig:NSSFlaw} illustrates this drawback through two created saliency maps. Though map (d) is much better than map (e) in Figure~\ref{fig:NSSFlaw}, it gets lower $NSS$ and $sNSS$ scores than map (e).  This happens because in the $NSS$ formulation,  when the normalized saliency values are averaged,  each fixation location contributes equally to the average. 
\section{Proposed Metric}
\label{sec:proposedmetrics}
\begin{figure}[t]
\begin{tabular}{cc}
\subfloat[Original Fixations]{\includegraphics[width=0.45\textwidth]{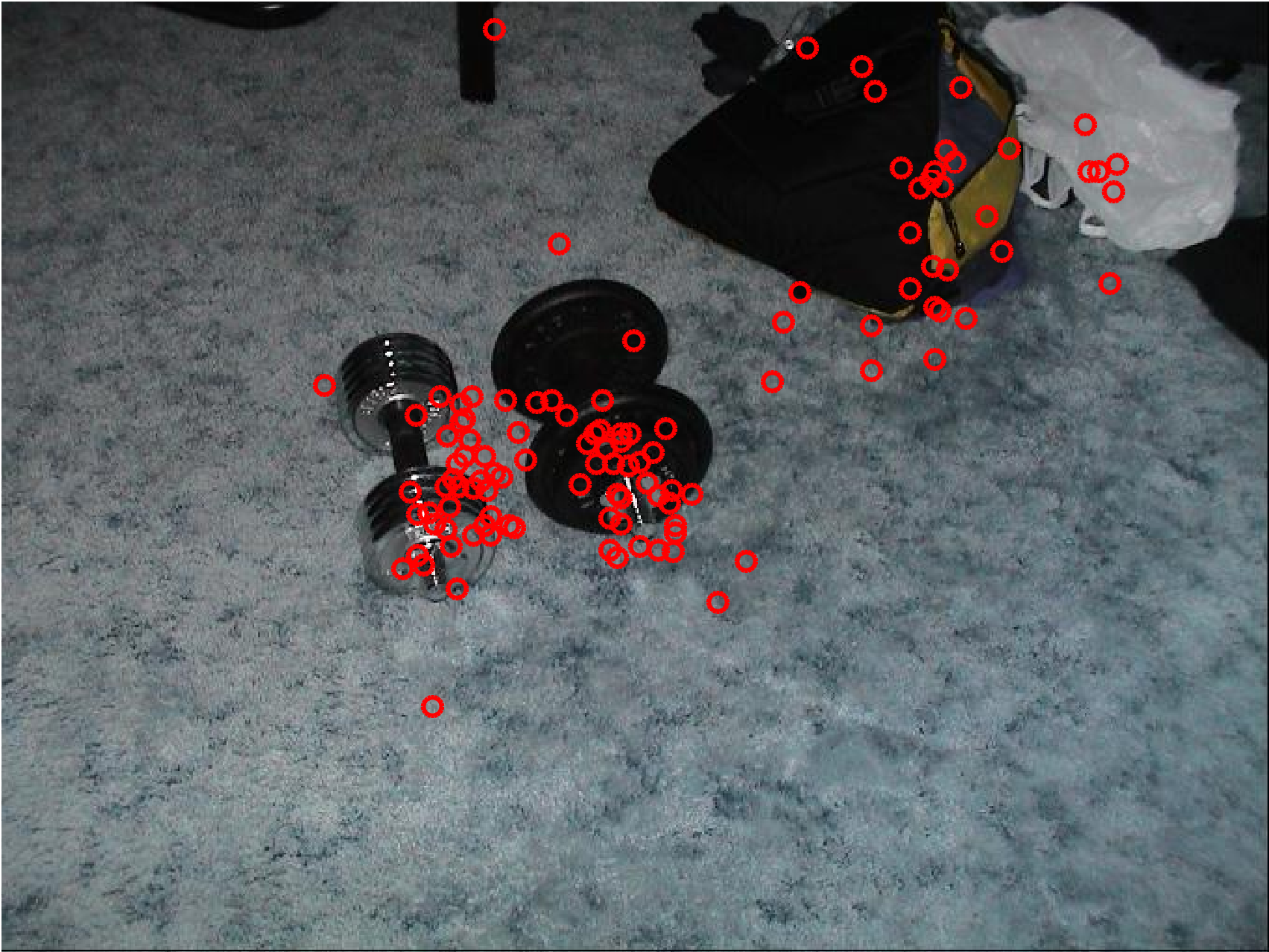}}
& \subfloat[Clustered Fixations using DBSCAN]{\includegraphics[width=0.45\textwidth]{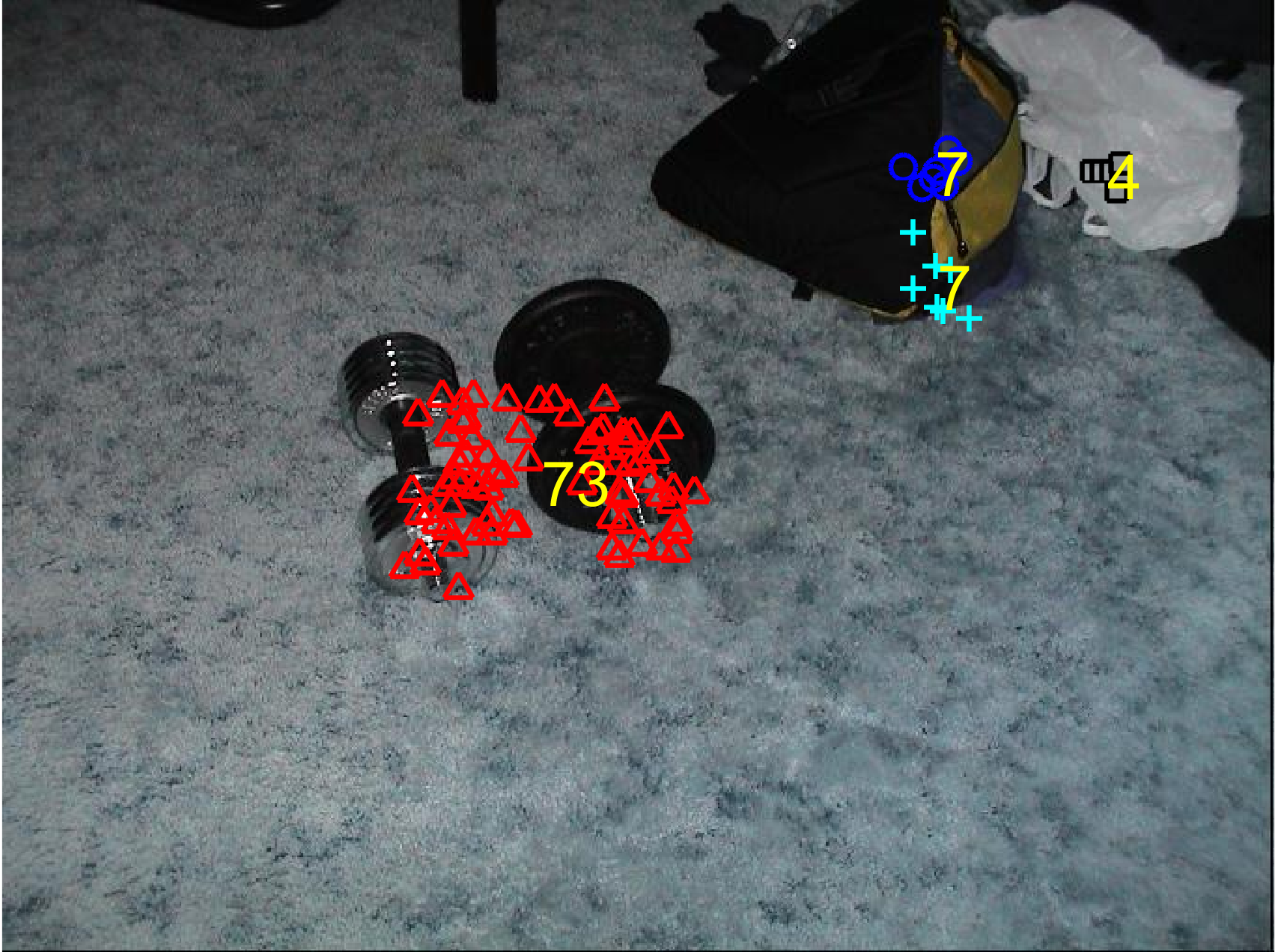}}
\end{tabular}
\caption{\added[id=MG]{(a) Importance of fixation density for identifying relevant fixations. The more important fixations are those that are clustered tightly together as they lie on the actual salient regions. The sparsely clustered fixations tend to lie on less salient background regions. (b) Fixation weights assigned based on the number of fixations in each cluster. Different symbols and colors represent different clusters with weights for each cluster shown.}}
\label{fig:DBSCANClusters}
\end{figure}

  Figure\added[id=MG]{~\ref{fig:DBSCANClusters}} illustrates that fixations that are closely clustered together lie on actual objects in the scene and are most important for identifying salient regions as compared to others that are scattered around and lie on background areas.  
  However, $sNSS$ and $NSS$ both do not discriminate between relevant fixations that belong to a dense cluster and represent objects, from fixations  that are sparse and usually fall on background regions and could be considered as outliers. One way to remedy this is to assign weights to each fixation point based on its importance. If $W(p)$ is the weight assigned to  the fixation point $p\in P$, where $P$ is the set of all fixation points, the proposed metric termed as weighted $NSS$ or $WNSS$ for short is defined as 
\begin{equation}
WNSS = \frac{1}{\sum\limits_{p \in P}W(p)}\sum\limits_{p \in P} \frac{W(p)(S(p) - \mu_s)}{\sigma_s} 
\label{eq:WNSS}
\end{equation}
where $\mu_s$ and $\sigma_s$ are the mean and standard deviation, respectively, of the predicted saliency map $S$,  and $P$ denotes the set of ground-truth fixation points for the image.
 To obtain appropriate weights for each fixation, we use the fact that fixations that are in higher density clusters are more important and should be weighted higher than those in low density clusters. For this purpose, we use the density-based spatial clustering of applications with noise (DBSCAN) algorithm~\cite{DBSCAN} for clustering the fixations based on their density to obtain fixation clusters. We then assign every fixation in a particular cluster a weight equal to the number of fixations in that cluster. Mathematically, if $P$ is the set of all fixation points, and if $C = \{c_1,c_2,...,c_N\}$ is the set of clusters obtained after applying the DBSCAN algorithm, such that $C$ is a partition of set $P$, then the weights are given by
 \begin{equation}
   W(p) = |c_i|, \forall p \in c_i 
 \end{equation} 
 where $|.|$ represents the $l_0$ norm corresponding to set cardinality (or the number of elements in the set).  
 The DBSCAN algorithm has two parameters, the minimum distance $\epsilon$ within which points are considered as belonging to the same cluster and the minimum number of points required to form a dense cluster $minPts$. We choose the $\epsilon$ parameter such that it is equal to the diameter of a circle that is subtended by one degree of visual angle for the eye-tracking setup for the Toronto dataset~\cite{AIM,VisualAngleDatabases}.  The $minPts$ parameter is chosen to be 3 so that isolated clusters with 2 or less points are rejected.  For rejected clusters, the weights are considered to be zero (as they represent clusters with zero points) and saliency values at such outlier fixations are ignored during the score computation. 
 An illustration of the weighting scheme is shown in Figure~\ref{fig:DBSCANClusters}(a)  where the different clusters of fixation points are shown using different colors, and the weight for each cluster is shown.  On comparison with the original fixations seen in Figure~\ref{fig:DBSCANClusters}(b) one can see that most of the outlier fixations on the carpet that are distant from the objects of interest are rejected and hence do not influence the score. 
 
 A shuffled version of the proposed metric that does not exhibit center-bias can be obtained in a manner similar to the $sNSS$ metric (\ref{eq:SNSS}) as follows: 
\begin{equation}
sWNSS= WNSS(p) - NSS(r) \\
\label{eq:sWNSS}
\end{equation}
where equal weights are assumed for the random set of fixation points $r$. \added[id=MG]{Equal weights are chosen for the random set of fixation points because, unlike the ``good'' fixation points, random fixations that are chosen from the set of fixations for other images cannot be weighted by a density based criteria and are treated equally in order to capture the centered distribution of fixations for the database to nullify center bias.}  
As shown in Figures~\ref{fig:snss} and~\ref{fig:NSSFlaw}, the proposed $WNSS$ and $sWNSS$ metrics give a higher score to the better map. The proposed $sWNSS$ metric is also able to correctly assign the lowest score to the centered Gaussian blob (e) in  Figure~\ref{fig:snss}.   

\section{Subjective Evaluation of VA Models}
\label{sec:subjectivestudy}
\begin{figure}[t]
	\centering
	\footnotesize
	\begin{tabular}{DCCC}
		& Original Image & Ground-Truth & Predicted Saliency \\
		\begin{turn}{90} Excellent \end{turn} & \includegraphics[width=0.12\textwidth]{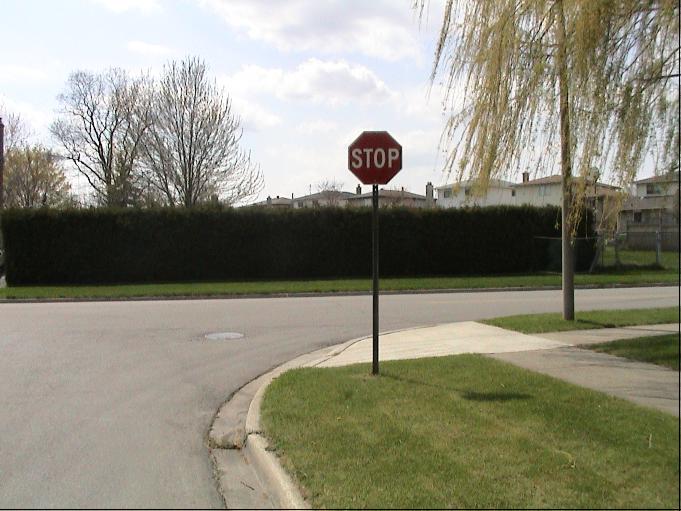} & 
		\includegraphics[width=0.12\textwidth]{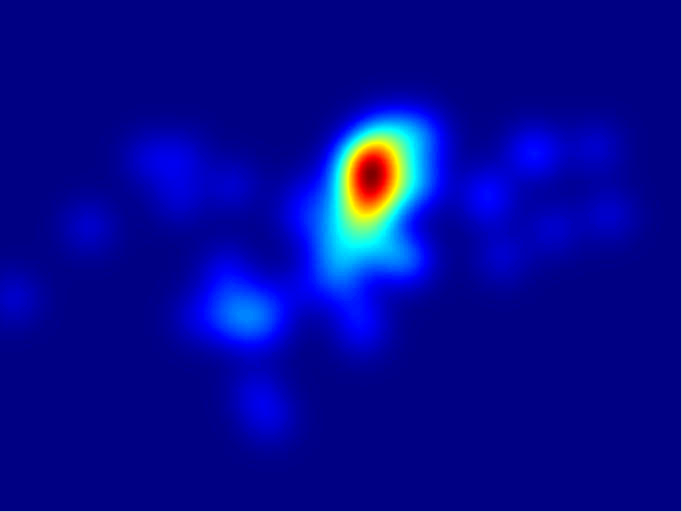} &
		\includegraphics[width=0.12\textwidth]{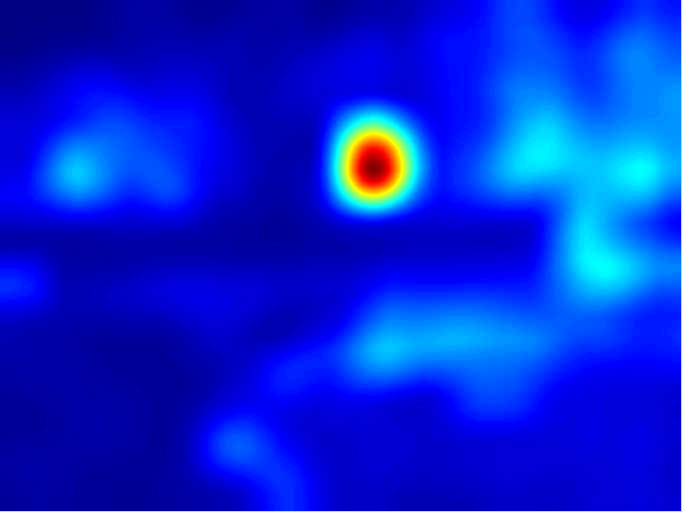} \\
		\begin{turn}{90} Good \end{turn} & \includegraphics[width=0.12\textwidth]{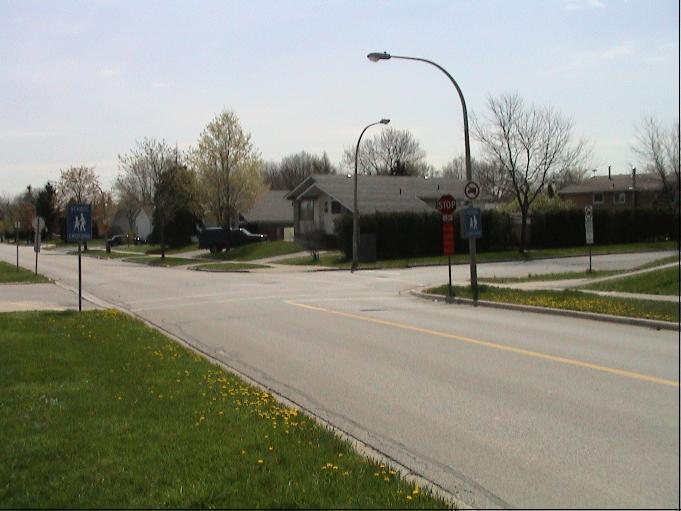} & 
		\includegraphics[width=0.12\textwidth]{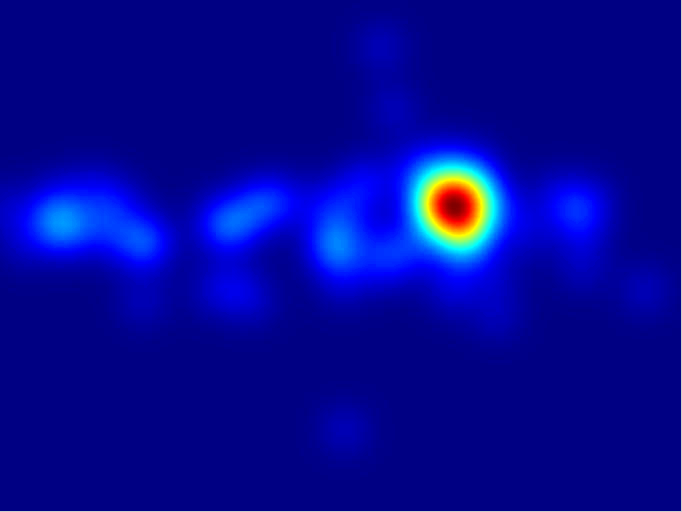} &
		\includegraphics[width=0.12\textwidth]{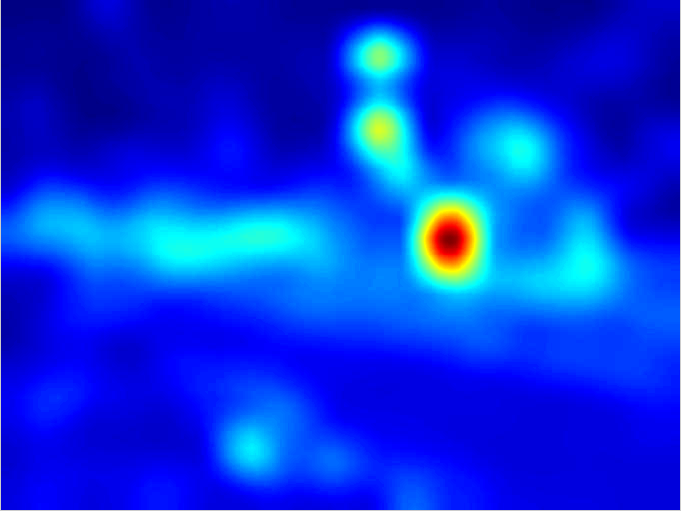} \\
		\begin{turn}{90} Fair \end{turn} & \includegraphics[width=0.12\textwidth]{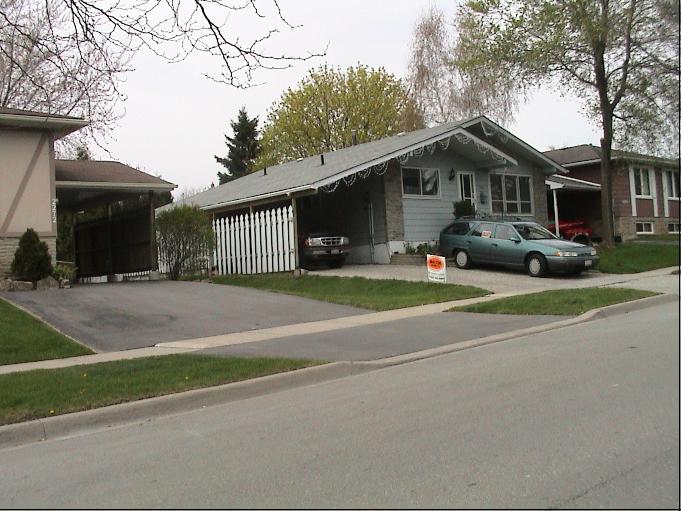} & 
		\includegraphics[width=0.12\textwidth]{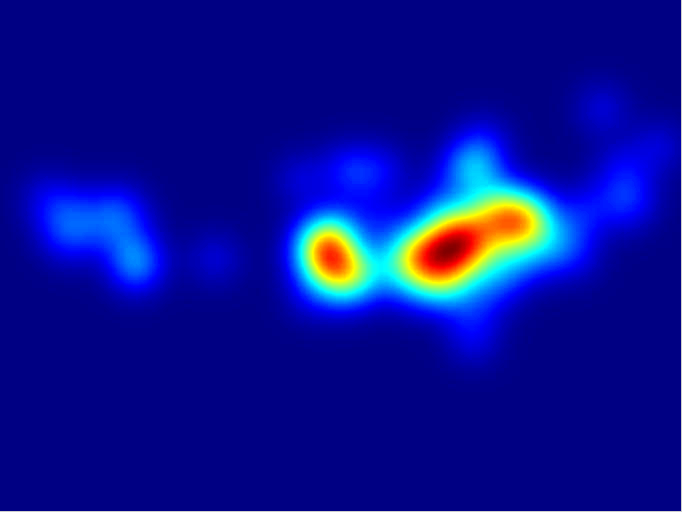} &
		\includegraphics[width=0.12\textwidth]{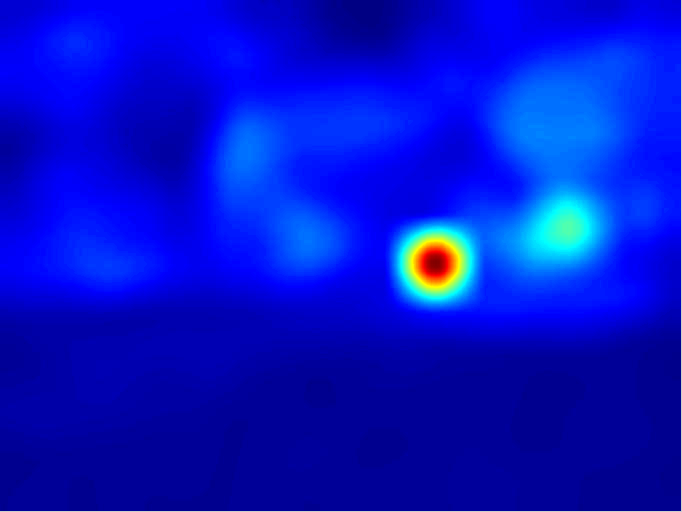} \\
		\begin{turn}{90} Poor \end{turn} & \includegraphics[width=0.12\textwidth]{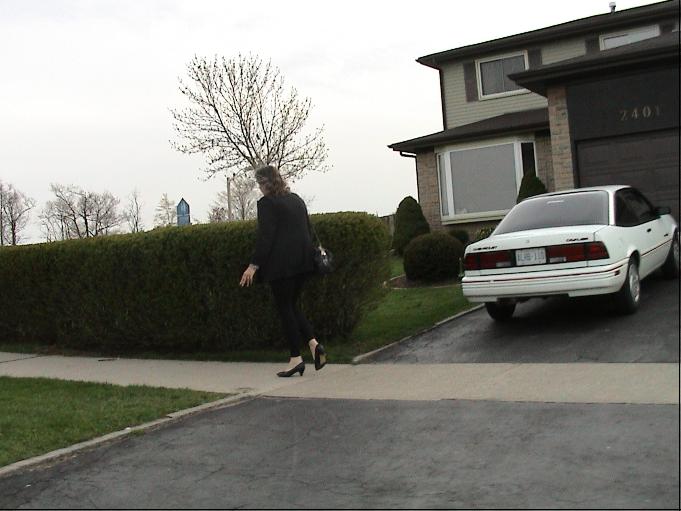} & 
		\includegraphics[width=0.12\textwidth]{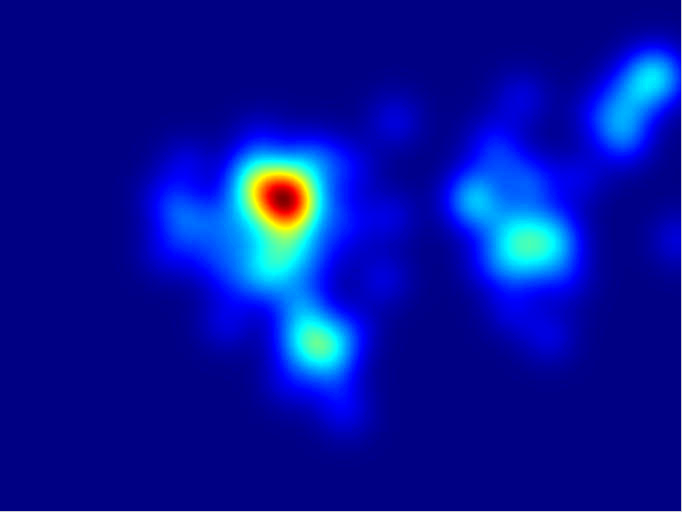} &
		\includegraphics[width=0.12\textwidth]{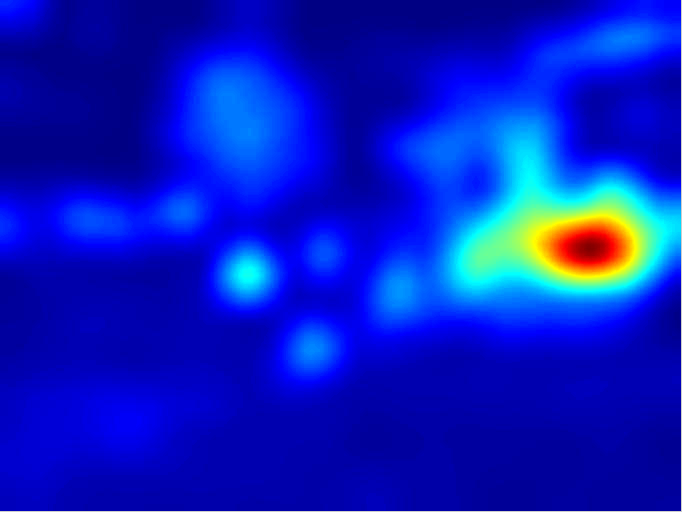} \\
		\begin{turn}{90} Bad \end{turn} & \includegraphics[width=0.12\textwidth]{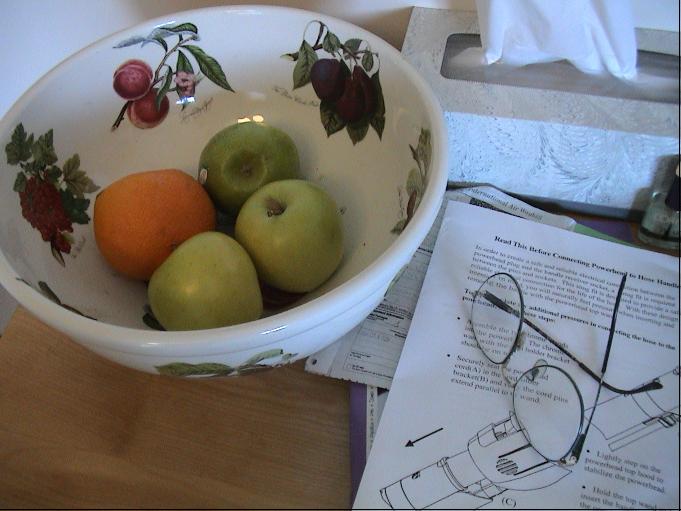} & 
		\includegraphics[width=0.12\textwidth]{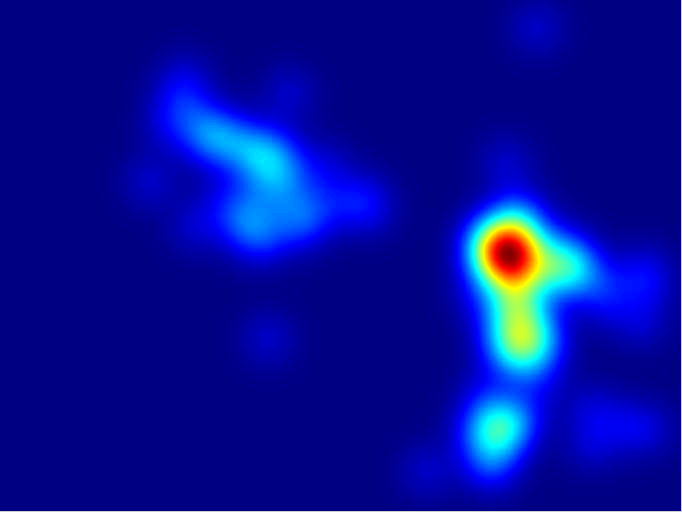} &
		\includegraphics[width=0.12\textwidth]{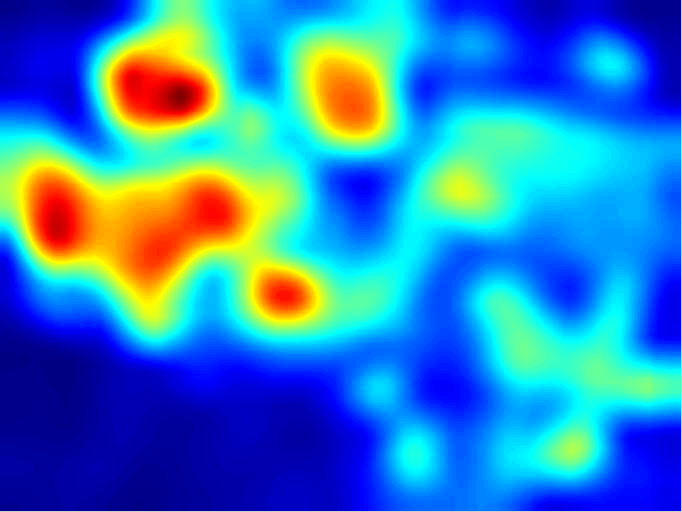} \\

	\end{tabular}
	\caption{Training samples from each category shown to subjects before taking the main test.}
	\label{fig:training}
\end{figure}

Even though a large number of metrics have been proposed in the literature for evaluating VA models, currently, there is no ground-truth subjective database that validates these metrics.  To address this need and evaluate the performance of our proposed metric,  a Visual Attention Quality (VAQ) database is constructed as part of this work.  The constructed database consists of saliency maps that are obtained from state-of-the-art VA models  and their corresponding ground-truth saliency maps. \added[id=MG] {A ground-truth saliency map is obtained by first aggregating the fixation locations obtained by eye-tracking for all subjects to get a fixation map. The obtained fixation map is then convolved with a 2D Gaussian kernel with a standard deviation $\sigma$ proportional to one degree of visual angle followed by normalization~\cite{LeMeur2013}. Thus, a ground-truth saliency map represents the likelihood that a pixel will be attended to by a human observer. As a result, ground-truth saliency maps are more suitable for at-a-glance visual comparisons as opposed to fixation points~\cite{AIM}.} Subjective ratings are obtained by asking human subjects to rate the similarity of the predicted saliency map to the corresponding ground-truth saliency map on a 5-point scale (5-Excellent, 4-Good, 3-Fair, 2-Poor, and 1-Bad). The two aspects the subjects are asked to focus on are the how well the locations of the highest intensity values in the ground-truth match those in the predicted saliency map and the amount of false-positive activity, i.e. high activity in the predicted saliency map that falls on regions of low activity in the ground-truth. The subjects are given a training session and are shown examples of each rating type from excellent to bad. Figure~\ref{fig:training} shows samples of the training images (one for each category) shown to the subjects. As seen in Figure~\ref{fig:training}, in the ``Excellent'' category the high saliency regions align very well with those in the ground-truth map and have minimal false-positive activity. The misalignment of high saliency regions and amount of false-positive activity increases for the ``Good'' and ``Fair'' categories. For the ``Poor'' to ``Bad'' categories the highest saliency regions in the ground truth and the predicted saliency maps are totally misaligned and the false positive activity increases from high to very-high, respectively.  

\begin{figure}
	\centering
	\includegraphics[width=0.5\textwidth]{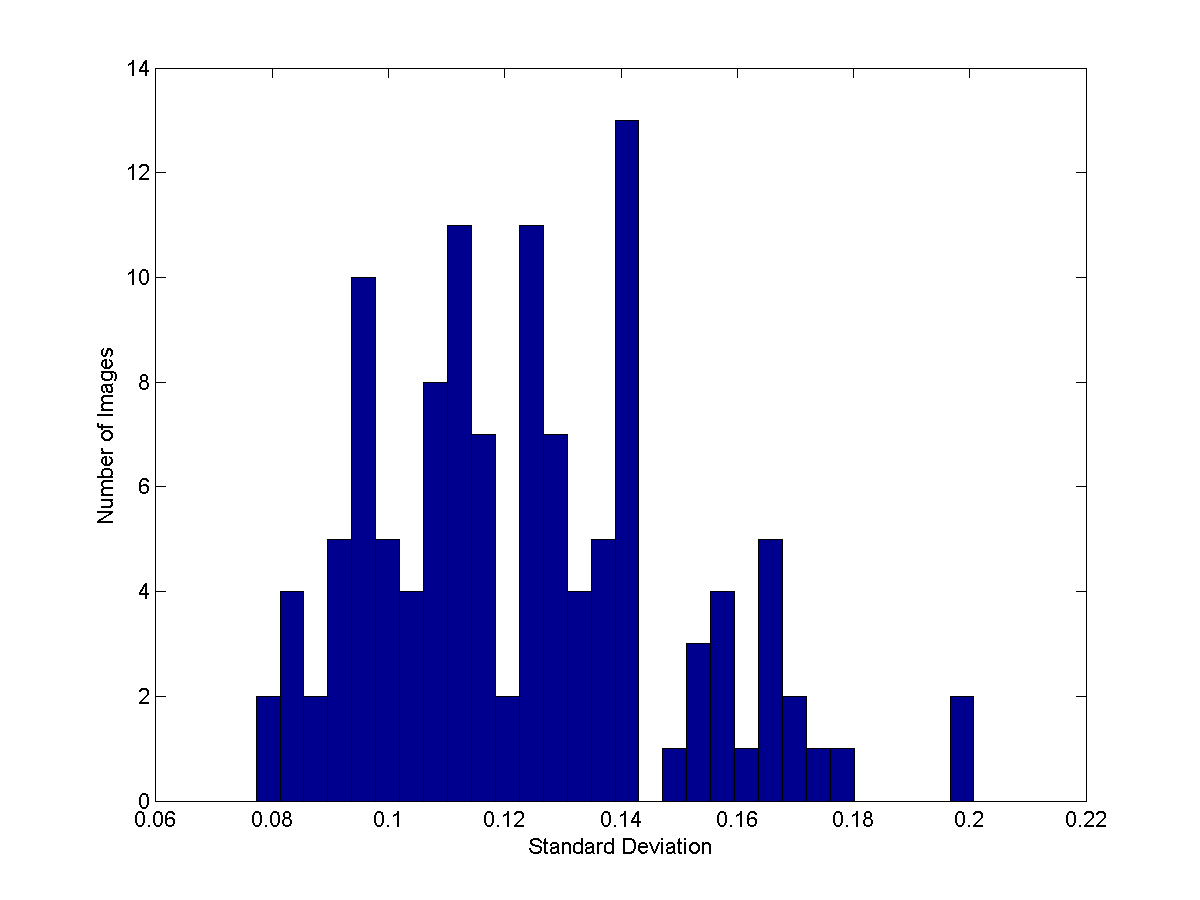}
	\caption{Histogram of standard deviations for all normalized ground-truth saliency maps in the Toronto dataset~\cite{AIM}.}
	\label{fig:stdhist}
	
\end{figure}
\begin{figure}[t]
	\centering
	\includegraphics[width=0.8\textwidth]{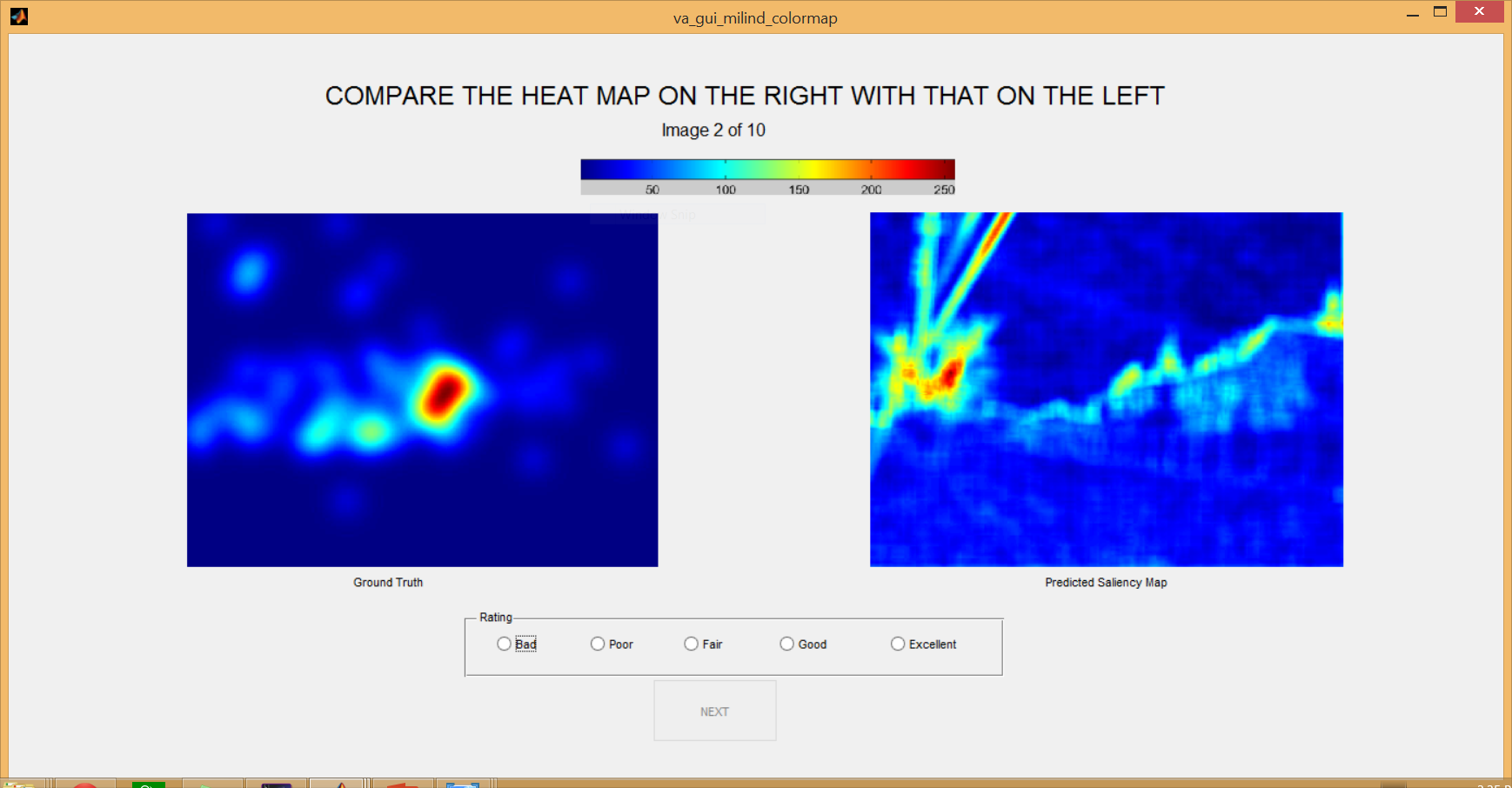}
	\caption{GUI used for obtaining subjective ratings.}
	\label{fig:GUI}	
\end{figure}
\begin{figure}[t]
	\centering
	\begin{tabular}{ccc}
		
		\includegraphics[width=0.12\textwidth]{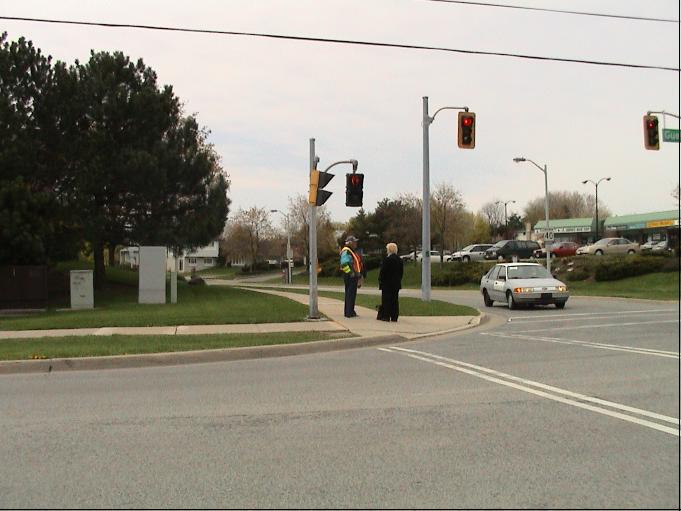} & 
		\includegraphics[width=0.12\textwidth]{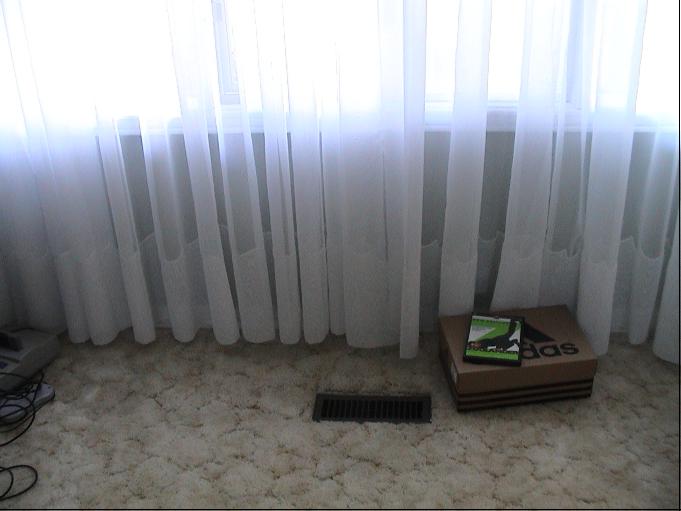} &
		\includegraphics[width=0.12\textwidth]{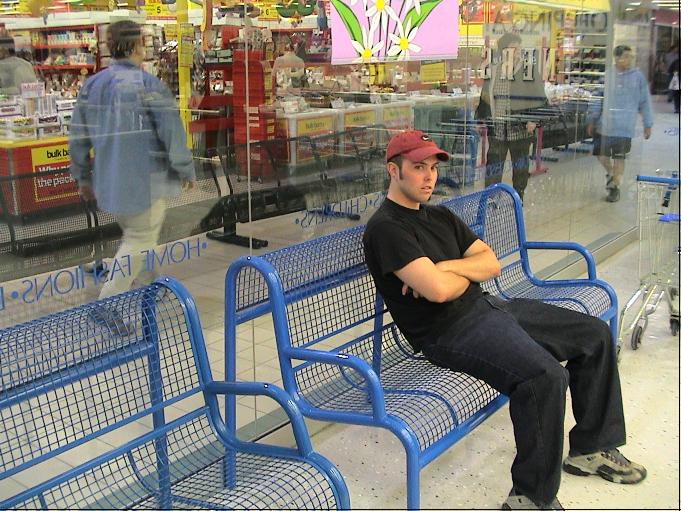} \\
		\includegraphics[width=0.12\textwidth]{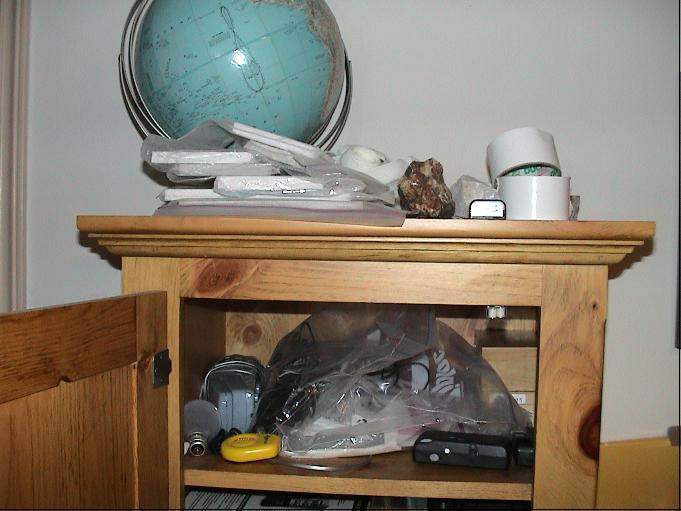} & 
		\includegraphics[width=0.12\textwidth]{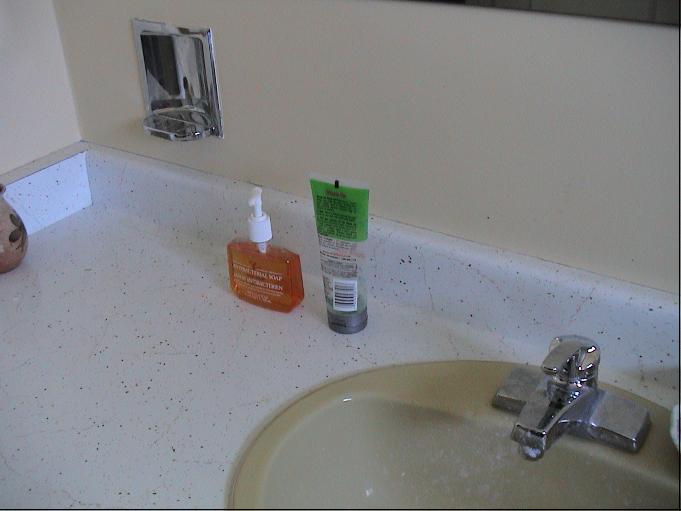} &
		\includegraphics[width=0.12\textwidth]{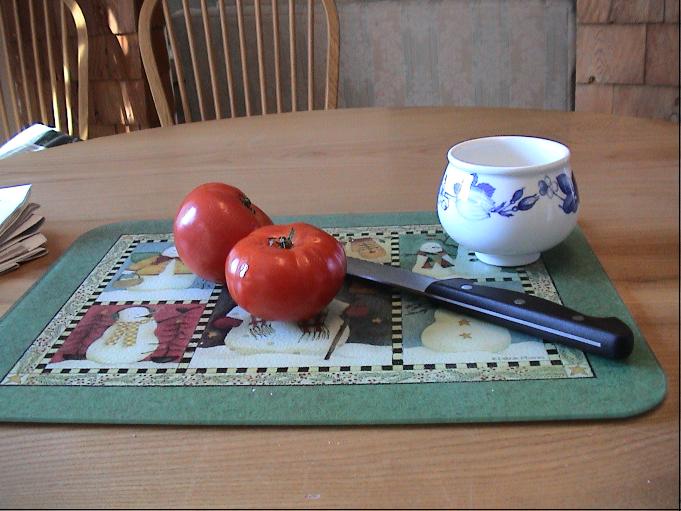} \\
		\includegraphics[width=0.12\textwidth]{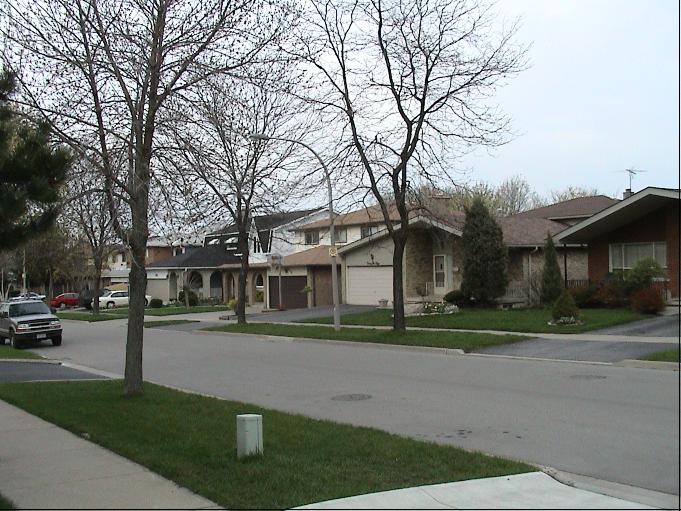} & 
		\includegraphics[width=0.12\textwidth]{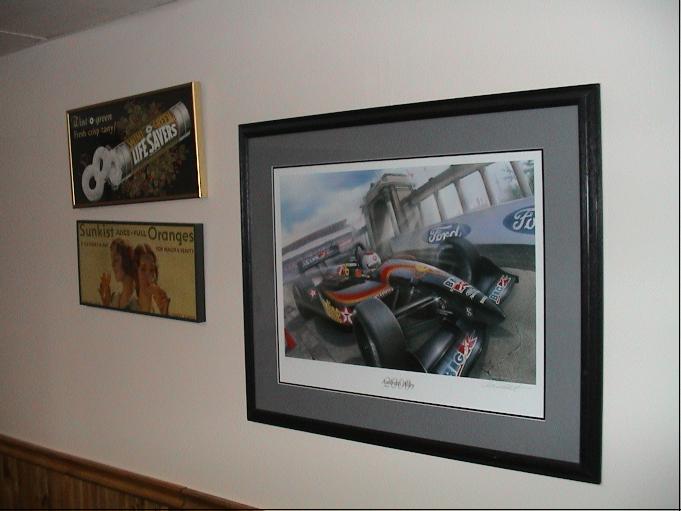} &
		\includegraphics[width=0.12\textwidth]{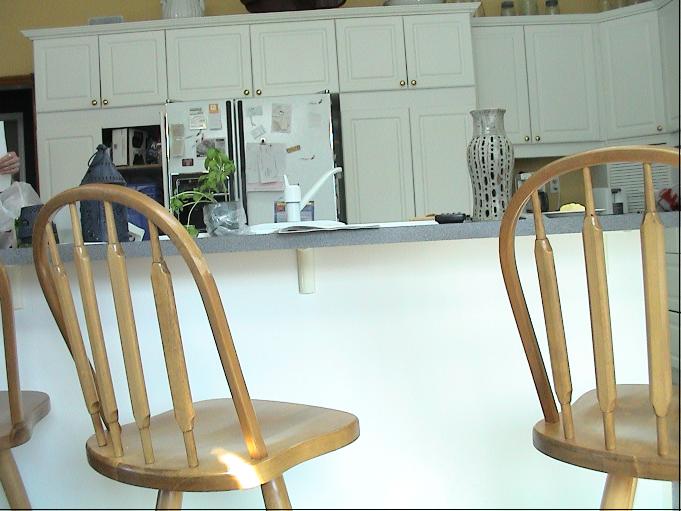} \\
		\includegraphics[width=0.12\textwidth]{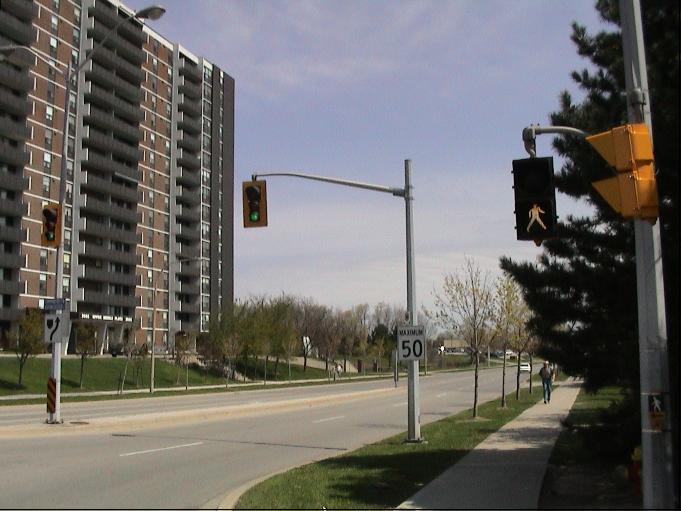} & 
		\includegraphics[width=0.12\textwidth]{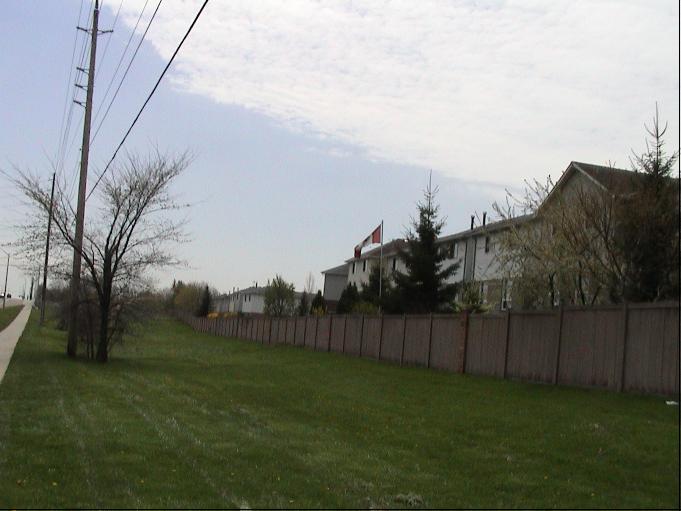} &
		\includegraphics[width=0.12\textwidth]{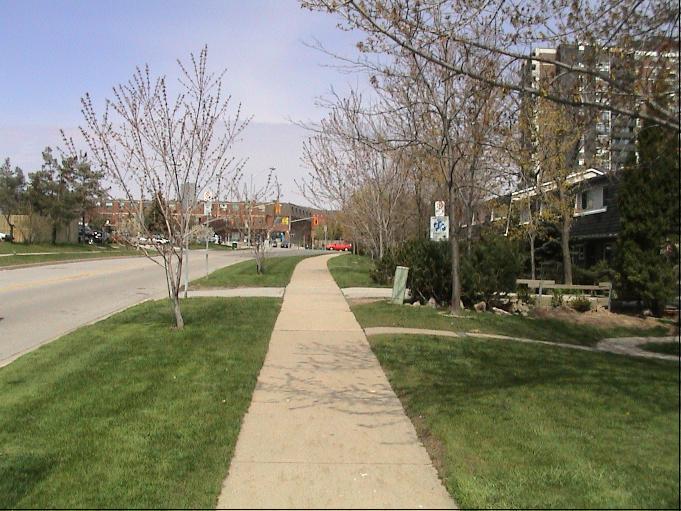} 
		
	\end{tabular}
	\caption{The 12 chosen images from the Toronto~\cite{AIM} dataset used in the main subjective test. }
	\label{fig:chosenOrigImgs}
\end{figure}

\begin{figure}[t]
	\centering
	\begin{tabular}{ccc}
		
		\includegraphics[width=0.12\textwidth]{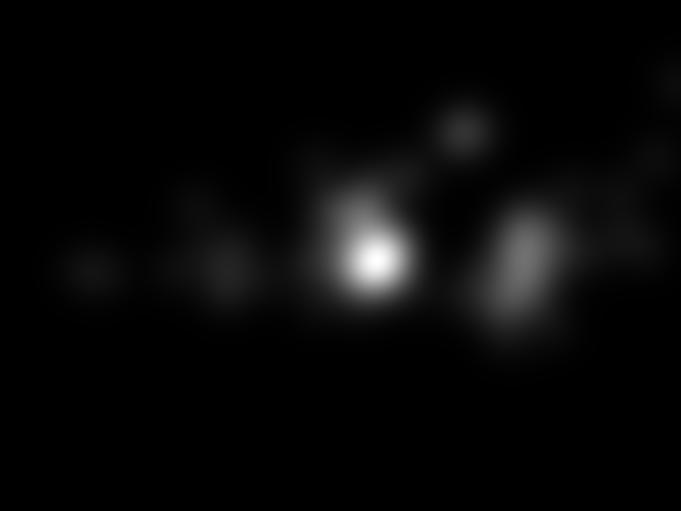} & 
		\includegraphics[width=0.12\textwidth]{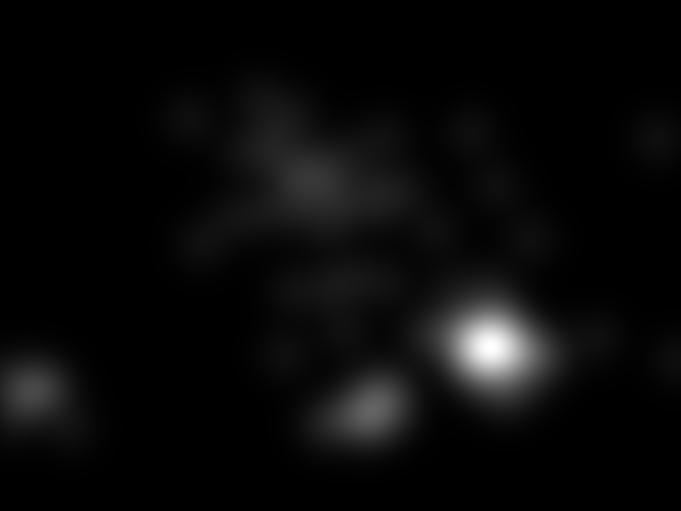} &
		\includegraphics[width=0.12\textwidth]{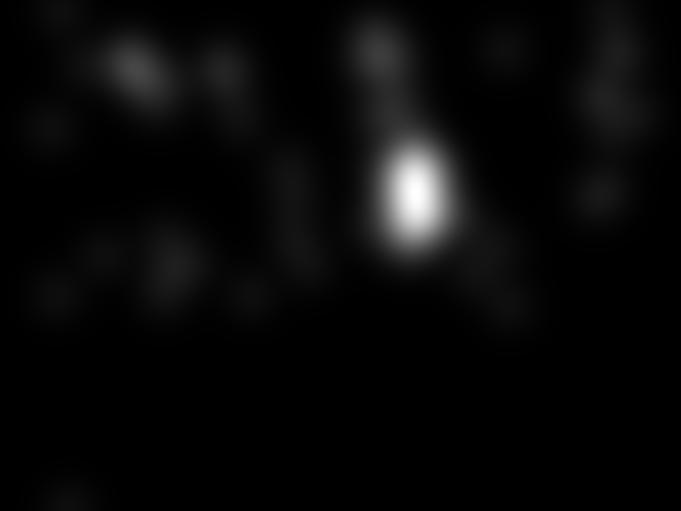} \\
		\includegraphics[width=0.12\textwidth]{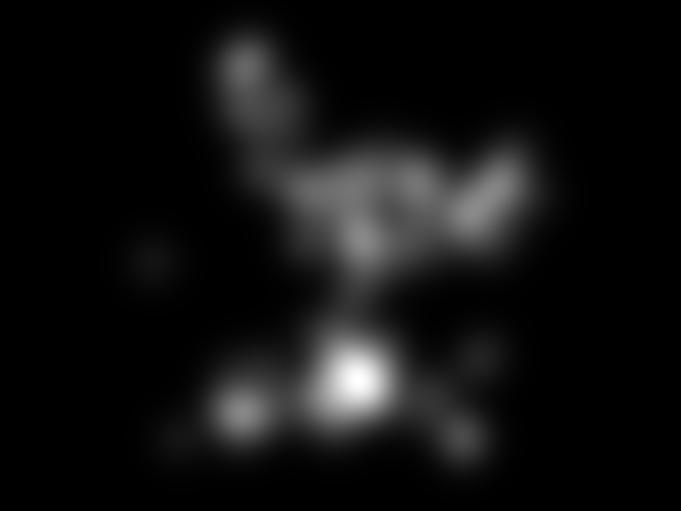} & 
		\includegraphics[width=0.12\textwidth]{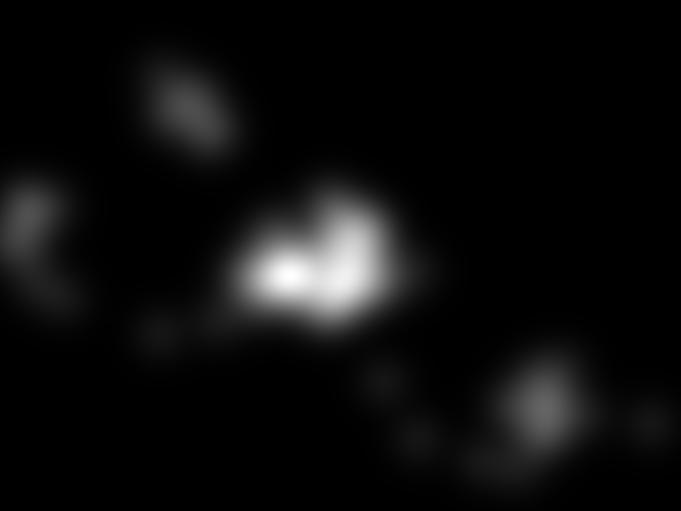} &
		\includegraphics[width=0.12\textwidth]{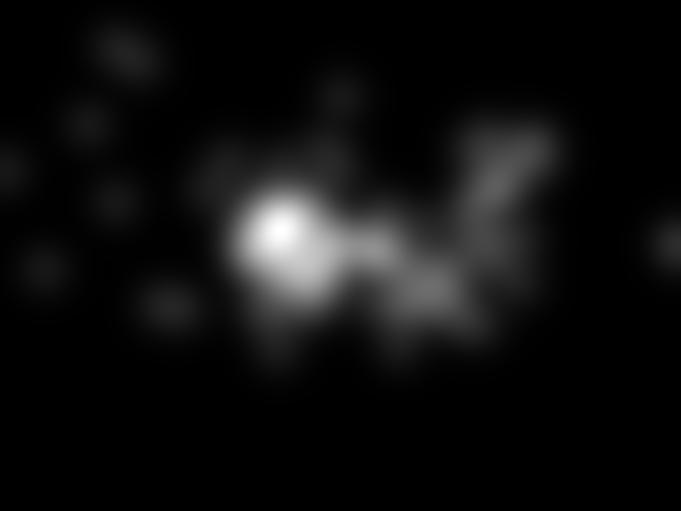} \\
		\includegraphics[width=0.12\textwidth]{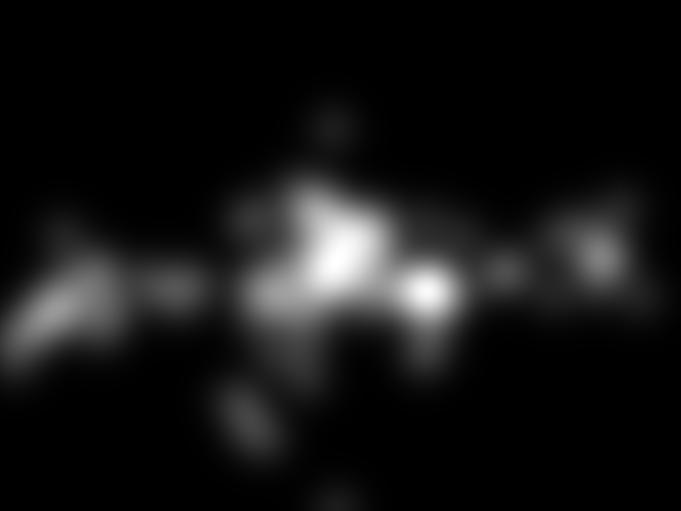} & 
		\includegraphics[width=0.12\textwidth]{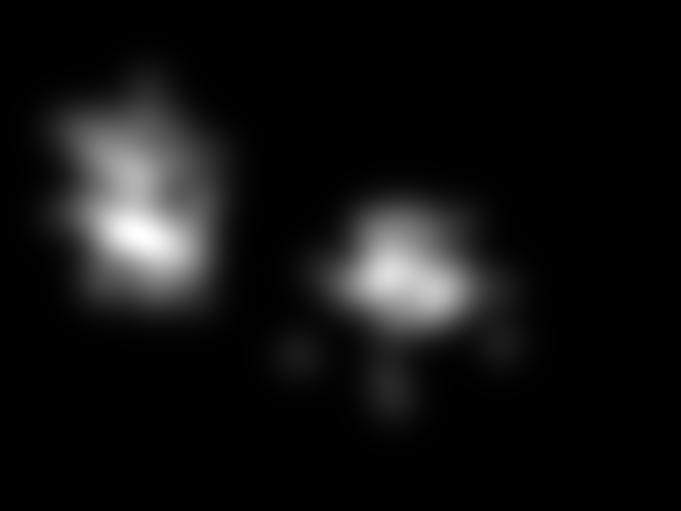} &
		\includegraphics[width=0.12\textwidth]{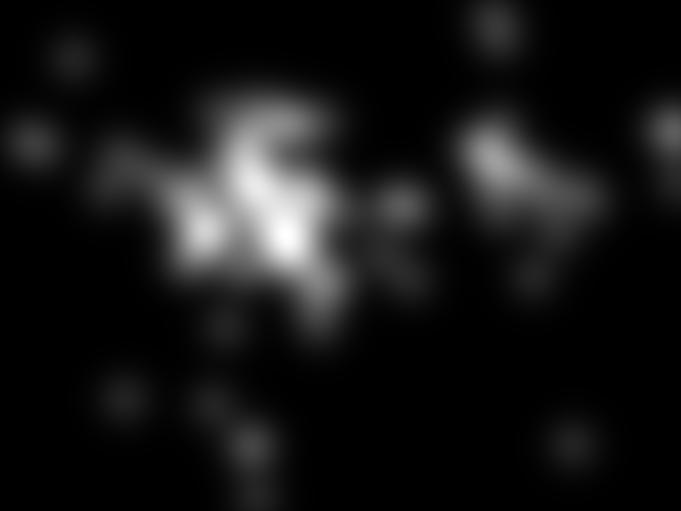} \\
		\includegraphics[width=0.12\textwidth]{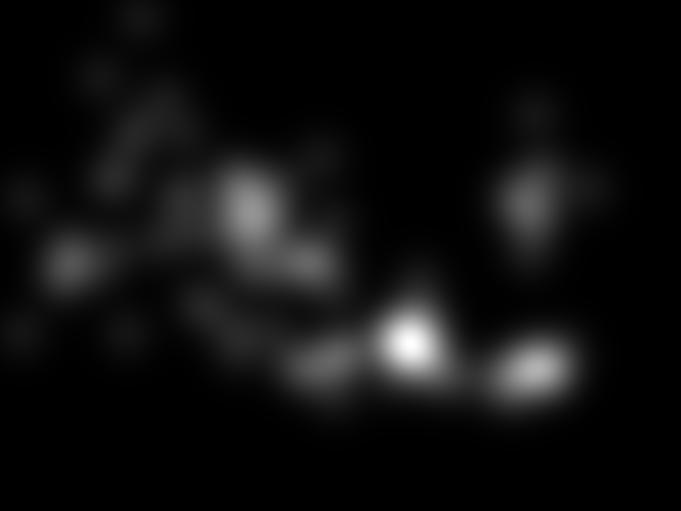} & 
		\includegraphics[width=0.12\textwidth]{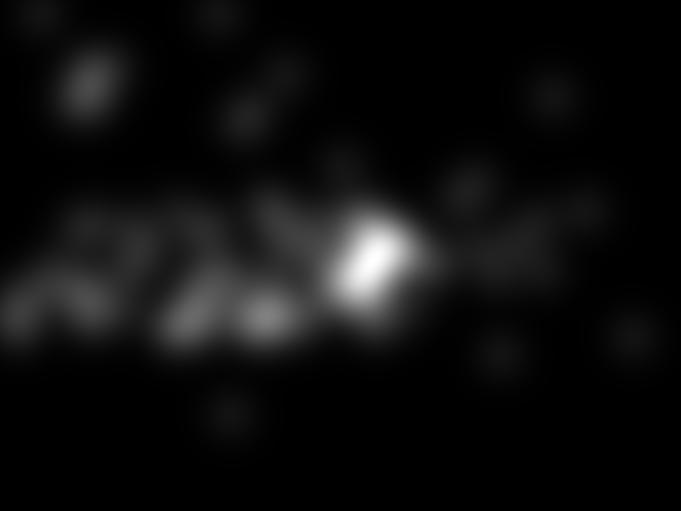} &
		\includegraphics[width=0.12\textwidth]{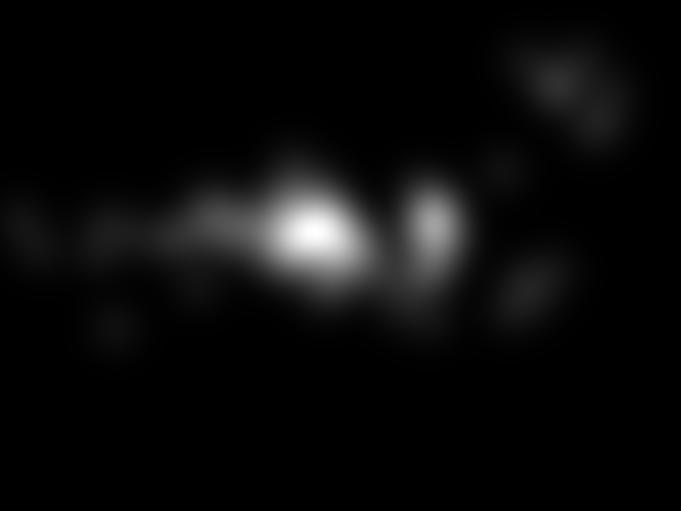} 
		
	\end{tabular}
	\caption{The ground-truth saliency maps for the 12 chosen images from the Toronto~\cite{AIM} dataset used in the main subjective test. }
	\label{fig:chosenFDM}
\end{figure}

The images shown in the training as well as main sessions were taken from the \added[id=MG]{popular} Toronto eye-tracking database~\cite{AIM}. \added[id=MG]{The images in that database have all the same size, which makes the computation of the shuffled metrics easier}. The images used in the training session were different from those in the main test. To ensure variety in the images shown in the main test, the ground truth maps for all the images were analyzed based on their standard deviation as it is a good measure of the spread of salient regions in an image. Figure~\ref{fig:stdhist} shows a histogram of the standard deviations of the ground-truth saliency maps of the images. Based on the 4 noticeable peaks in the histogram, we cluster the images in the dataset based on their standard deviations by using kmeans. Then, the 3 images with standard deviation nearest to the cluster centroids are chosen for each cluster. This gives us the 12 images that are used in the main test. The GUI used for the subjective testing and the colormap used is shown in Figure~\ref{fig:GUI}. Figure~\ref{fig:chosenOrigImgs} shows the images that are chosen and Figure~\ref{fig:chosenFDM} shows their ground-truth saliency maps. We then compute the predicted saliency maps for each of these 12 images using the following 17 state-of-the-art VA models:  GAFFE~\cite{GAFFE}, ITTI~\cite{Itti}, GBVS~\cite{GBVS}, AIM~\cite{AIM},  HouNIPS~\cite{HouNIPS}, GR~\cite{GR}, SDSR~\cite{SDSR}, SUN~\cite{SUN}, 
Torralba~\cite{AudeTorralba},      
FES~\cite{FES}, SigSal~\cite{SigSal}, SpectRes~\cite{SpectRes},
AWS~\cite{AWS}, BMS~\cite{BMS}, Context~\cite{Context}, CovSal~\cite{CovSal},
and RandomCS~\cite{RandomCS}. We also evaluate the ``center''  model which is an image independent model that consists of a centered Gaussian blob. In addition, the original ground truth saliency maps are also added to the list of images shown. We expect the ``center'' model to get lower scores in most cases and the original ground truth saliency maps to get the highest score. The total pairs of ground-truth saliency and test saliency maps shown are 228. These are presented to each subject in a randomized order.  Both the ground-truth and predicted saliency maps are shown with the `jet' colormap that indicates high intensity values in red and low intensity values in blue to make it easy for the subjects to assess the maps. 
 The maps were shown to 16 subjects \added[id=MG]{with age ranging between 22 to 33,} who were checked for both color blindness as well as visual acuity. \added[id=MG]{Out of the 16 participants, 1 subject was working in the area of visual attention, 7 subjects were working in the area of computer vision but not specifically in the area of visual attention, and 8 subjects were working in areas completely unrelated to computer vision. Out of the 16 subects,  6  were female and 10 were male.} The ratings for each predicted saliency map shown were averaged over 16 subjects to get a mean opinion score. These mean opinion scores (MOS) were then correlated with the scores obtained from popular VA performance metrics in addition to the proposed WNSS and sWNSS metrics.  Figure~\ref{fig:scoreDist} shows the distribution of the MOS scores obtained for the predicted saliency maps given by VA models. It illustrates that in only about 16\% of the cases, models received a subjective rating of good or excellent. It also shows that the saliency maps shown cover the entire range of ratings from Excellent to Poor.   The VAQ database will be provided online to download for free for the research community to benchmark metrics developed in the future. 

Figure~\ref{fig:OverallMOS}  shows the ranking of all models in terms of the mean subjective rating obtained for each VA model over all subjects for the VAQ database and shows which VA models are preferred by the human observers for the images in the VAQ database.  
\begin{figure}[t]
	\centering
	\includegraphics[width=0.5\linewidth]{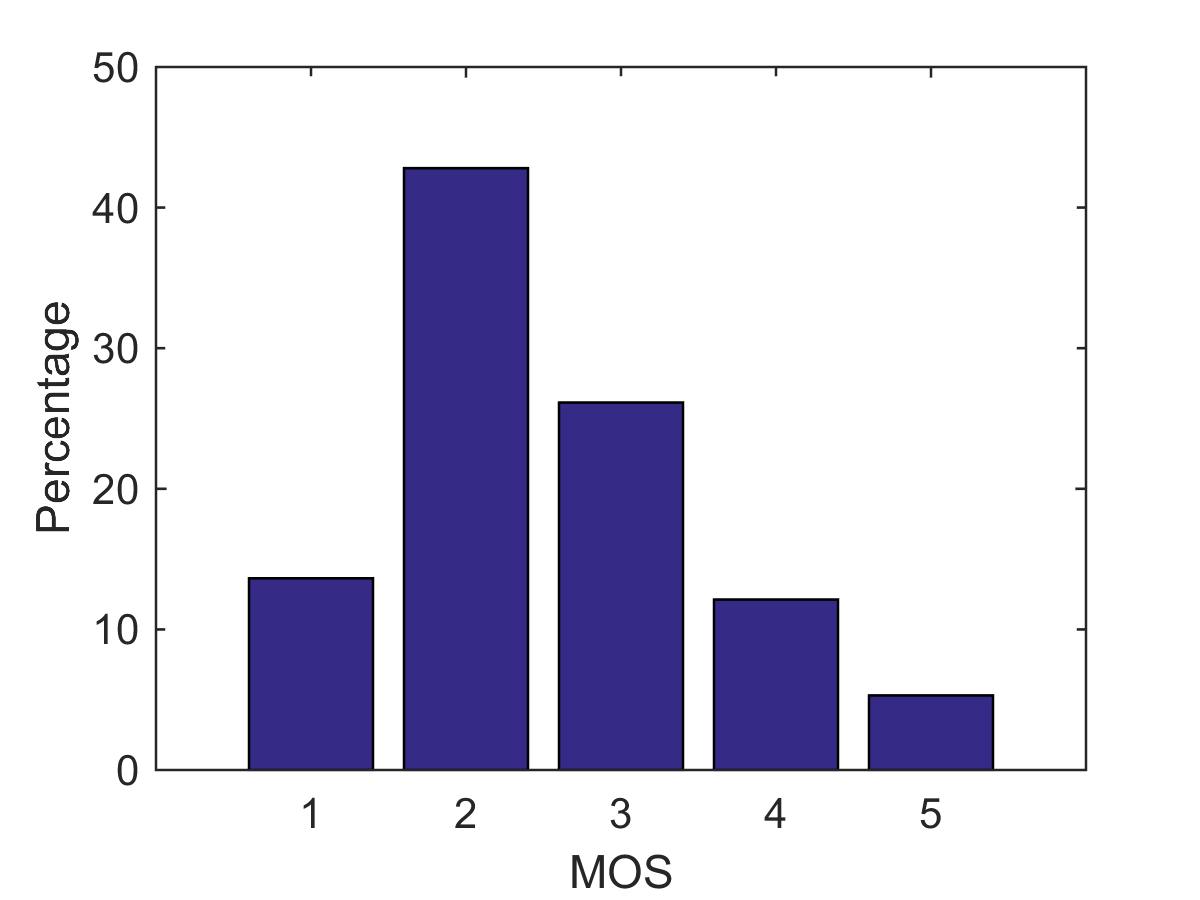}
	\caption{Distribution of MOS scores.}
	\label{fig:scoreDist}
\end{figure}
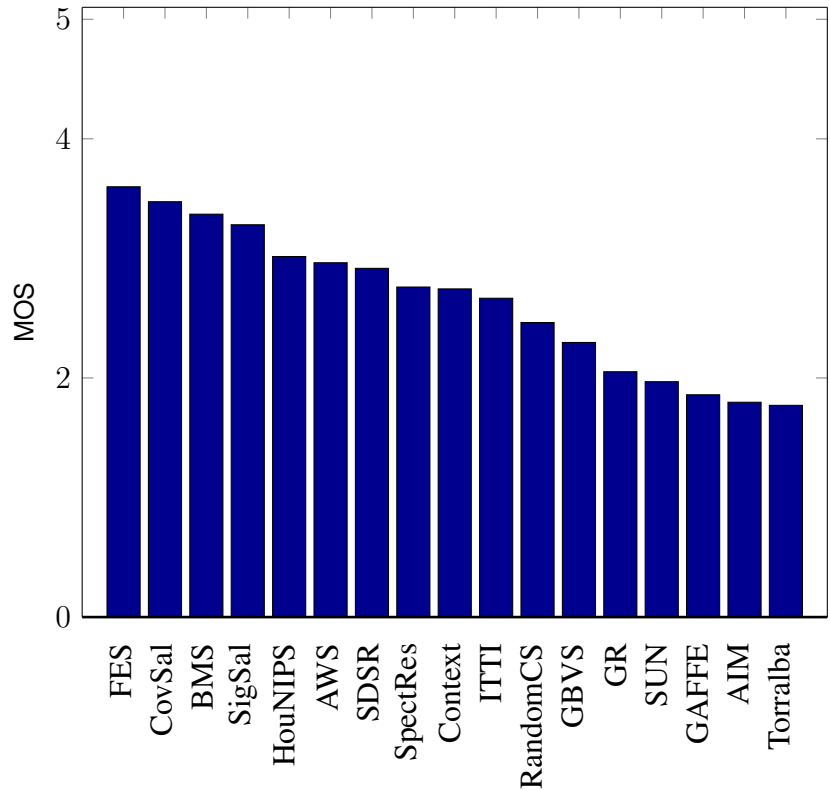
\begin{figure}[t]
	\centering
	\setlength{\figureheight}{0.5\linewidth}
	\setlength{\figurewidth}{0.6\linewidth}
%
%
\definecolor{mycolor1}{rgb}{0.00000,0.00000,0.56250}%
\begin{tikzpicture}

\begin{axis}[%
width=\figurewidth,
height=0.982818\figureheight,
at={(0\figurewidth,0\figureheight)},
scale only axis,
area legend,
clip=false,
separate axis lines,
every outer x axis line/.append style={black},
every x tick label/.append style={font=\color{black}},
xmin=0,
xmax=18,
xtick={1,2,3,4,5,6,7,8,9,10,11,12,13,14,15,16,17},
xticklabels={\empty},
every outer y axis line/.append style={black},
every y tick label/.append style={font=\color{black}},
ymin=0,
ymax=5.1,
ytick={0,2,4,5},
ylabel={MOS}
]
\addplot[ybar,bar width=0.044444\figurewidth,draw=black,fill=mycolor1] plot table[row sep=crcr] {%
1	3.59895833333333\\
2	3.47395833333333\\
3	3.36979166666667\\
4	3.28125\\
5	3.015625\\
6	2.96354166666667\\
7	2.91666666666667\\
8	2.76041666666667\\
9	2.74479166666667\\
10	2.66666666666667\\
11	2.46354166666667\\
12	2.296875\\
13	2.05208333333333\\
14	1.96875\\
15	1.859375\\
16	1.796875\\
17	1.77083333333333\\
};
\addplot [color=black,solid,line width=1.0pt,forget plot]
  table[row sep=crcr]{%
0	0\\
18	0\\
};
\node[left, align=right, inner sep=0mm, rotate=90, text=black]
at (rel axis cs:0.0513698630136986,-0.0418604651162791,0) {FES};
\node[left, align=right, inner sep=0mm, rotate=90, text=black]
at (rel axis cs:0.106164383561644,-0.0418604651162791,0) {CovSal};
\node[left, align=right, inner sep=0mm, rotate=90, text=black]
at (rel axis cs:0.160958904109589,-0.0418604651162791,0) {BMS};
\node[left, align=right, inner sep=0mm, rotate=90, text=black]
at (rel axis cs:0.215753424657534,-0.0418604651162791,0) {SigSal};
\node[left, align=right, inner sep=0mm, rotate=90, text=black]
at (rel axis cs:0.270547945205479,-0.0418604651162791,0) {HouNIPS};
\node[left, align=right, inner sep=0mm, rotate=90, text=black]
at (rel axis cs:0.328767123287671,-0.0418604651162791,0) {AWS};
\node[left, align=right, inner sep=0mm, rotate=90, text=black]
at (rel axis cs:0.383561643835616,-0.0418604651162791,0) {SDSR};
\node[left, align=right, inner sep=0mm, rotate=90, text=black]
at (rel axis cs:0.438356164383562,-0.0418604651162791,0) {SpectRes};
\node[left, align=right, inner sep=0mm, rotate=90, text=black]
at (rel axis cs:0.493150684931507,-0.0418604651162791,0) {Context};
\node[left, align=right, inner sep=0mm, rotate=90, text=black]
at (rel axis cs:0.547945205479452,-0.0418604651162791,0) {ITTI};
\node[left, align=right, inner sep=0mm, rotate=90, text=black]
at (rel axis cs:0.602739726027397,-0.0418604651162791,0) {RandomCS};
\node[left, align=right, inner sep=0mm, rotate=90, text=black]
at (rel axis cs:0.657534246575342,-0.0418604651162791,0) {GBVS};
\node[left, align=right, inner sep=0mm, rotate=90, text=black]
at (rel axis cs:0.715753424657534,-0.0418604651162791,0) {GR};
\node[left, align=right, inner sep=0mm, rotate=90, text=black]
at (rel axis cs:0.770547945205479,-0.0418604651162791,0) {SUN};
\node[left, align=right, inner sep=0mm, rotate=90, text=black]
at (rel axis cs:0.825342465753425,-0.0418604651162791,0) {GAFFE};
\node[left, align=right, inner sep=0mm, rotate=90, text=black]
at (rel axis cs:0.88013698630137,-0.0418604651162791,0) {AIM};
\node[left, align=right, inner sep=0mm, rotate=90, text=black]
at (rel axis cs:0.934931506849315,-0.0418604651162791,0) {Torralba};
\end{axis}
\end{tikzpicture}%
	\caption{MOS taken over all predicted saliency maps for each VA model and arranged in descending order.}
	\label{fig:OverallMOS}
\end{figure}

\section{Metrics Performance Evaluation Results}
\label{sec:results}
\begin{table}[t]
	\centering
	\caption{Correlation results using the VAQ database.}
	\label{tab:correlationresult}
	
	\begin{tabular}{|c|c|c|c|}
   
   \hline
      \textbf{Non-Shuffled Metrics} & SROCC & KROCC & PLCC \\
    \hline
	 $AUC_{Borji}$~\cite{borjieval} & 0.5617   & 0.3952  & 0.5493 \\
	 $AUC_{Judd}$~\cite{Judd_2012} &  0.5883 & 0.4222  &  0.5846 \\
	 $WF_{\beta}$~\cite{margolin2014evaluate} & 0.3126 & 0.2113 & 0.2958 \\   
	 $NSS$~\cite{NSS}   & 0.7563 & 0.5810 & 0.8297 \\
	 $EMD$~\cite{Judd_2012}   & 0.4470 & 0.3168 & 0.5683 \\ 
	 $CC$~\cite{Judd_2012}    & 0.7461 &  0.5726 & 0.8216 \\
	 $SIM$~\cite{Judd_2012}   & 0.5891  &  0.4246 & 0.6739 \\
	 \added[id=MG]{$MAE$~\cite{MAE}} &\added[id=MG]{0.5063} & \added[id=MG]{0.3599}& \added[id=MG]{0.4803}\\ 
     Proposed $WNSS$  & \bf{0.7858} & \bf{0.6178} & \bf{0.8687}\\
     \hline
        \textbf{Shuffled Metrics}  & SROCC & KROCC & PLCC \\
        \hline
     $sAUC$~\cite{borjieval}	 & 0.5455 & 0.3871 & 0.5631 \\    
	 $sNSS$~\cite{MilindSamTPAMI}	& 0.6526 & 0.4843 & 0.7533\\
	 Proposed $sWNSS$ & \bf{0.7624}& \bf{0.5891} &  \bf{0.8553} \\  
    \hline
	\end{tabular}
	
\end{table}

\begin{figure*}[t]
	\centering
	\setlength{\figureheight}{0.1\linewidth}
	\setlength{\figurewidth}{0.2\linewidth}
	\begin{tabular}{ccc}
		\hspace*{-0.3in}
%
%
\begin{tikzpicture}

\begin{axis}[%
width=\figurewidth,
height=0.982818\figureheight,
at={(0\figurewidth,0\figureheight)},
scale only axis,
every outer x axis line/.append style={black},
every x tick label/.append style={font=\color{black}},
xmin=0.4,
xmax=1.2,
xlabel={Metric Score},
every outer y axis line/.append style={black},
every y tick label/.append style={font=\color{black}},
ymin=1,
ymax=5,
ylabel={MOS},
title={SROCC 0.5617 KROCC 0.3952 PLCC 0.5493},
axis x line*=bottom,
axis y line*=left,
title style={font=\scriptsize},xlabel style={font=\scriptsize},ylabel style={font=\scriptsize}
]
\addplot[only marks,mark=o,mark options={},mark size=1.0000pt,color=blue] plot table[row sep=crcr,]{%
0.845533780568188	1.3125\\
0.928278939201761	2.25\\
0.826857614982099	1.6875\\
0.829020032535788	2.1875\\
0.837209507831806	1.5\\
0.807669857576465	2.1875\\
0.834013529453223	1.9375\\
0.85633502210366	1.9375\\
0.934529368670063	1.9375\\
0.825296519608695	1.625\\
0.826745924990904	1.5\\
0.929721406804486	2.3125\\
1.00081548539792	4.9375\\
1.00032628569112	5\\
1.00180618507987	5\\
0.997624592237222	4.9375\\
0.99647750115291	5\\
1.00302448624596	5\\
0.99618238147246	5\\
1.0032292787944	5\\
1.00034981533005	5\\
0.999427469100376	5\\
0.998543819869562	5\\
1.00217229082043	5\\
0.792709555260877	1.4375\\
0.731267853077357	1.5625\\
0.849810664661184	2.375\\
0.659835743660618	1.3125\\
0.771941349926513	1.5\\
0.949303495618344	2.5625\\
0.746373302911861	2.0625\\
0.803396614077331	1.5\\
0.947061961244193	1.6875\\
0.682741248986641	1.25\\
0.891642080557585	2.875\\
0.848821890108751	1.4375\\
0.746067618098742	2.5625\\
0.847338497821151	2\\
0.886841294264881	3\\
0.788370365739527	2.6875\\
0.786732092696008	1.6875\\
0.937109814451925	3.5625\\
0.704546767194022	3.9375\\
0.905805866062344	2.75\\
1.02777786273316	3.3125\\
0.746752730634561	2.0625\\
0.880654339998497	4.9375\\
0.982463375766402	3.0625\\
0.839727689511339	2.6875\\
0.878156315721373	3.1875\\
0.767473633496895	2.8125\\
0.816309382511458	1.9375\\
0.788862067426594	2.25\\
0.949241262130751	3.9375\\
0.737792251335319	4\\
0.858525870823942	4.625\\
0.991392674506403	2.75\\
0.778450046737408	2.8125\\
0.860903729803037	5\\
0.932371661229477	4.4375\\
0.719320474589176	2.6875\\
0.900428335951588	2.625\\
0.932939202490778	2.75\\
0.740725740034688	1.9375\\
0.681843261271171	1.1875\\
0.911948857222866	3.9375\\
0.776717932619809	3.25\\
0.862926104004664	2.25\\
0.999735405667295	2.5\\
0.87987959097741	2.625\\
0.899759659378226	4.3125\\
0.962681506975461	2.875\\
0.77811333880974	2.125\\
0.922001119197067	4.125\\
0.727894611608866	3.25\\
0.806727052954308	4.125\\
0.789089834781583	2.25\\
0.817903582285652	4.6875\\
0.817905058562802	3.5\\
0.829764657667175	4.5625\\
0.924587050950629	4.0625\\
0.824560866616822	2.9375\\
0.808864322402011	1.8125\\
0.945493588945599	4.25\\
0.810498956593394	1.375\\
0.816438937852072	1.4375\\
0.695986085463217	1.6875\\
0.696217418663393	1.5\\
0.83905853736677	1.4375\\
0.870331216362394	2.6875\\
0.811779044013854	2.125\\
0.870577483201979	2.4375\\
0.944700664804921	2\\
0.777173564726962	1.625\\
0.919058304690675	2.3125\\
0.893848803084554	1.6875\\
0.77776639621312	1.625\\
0.84132454058492	1.875\\
0.774284991727176	1.875\\
0.707801320434617	1.8125\\
0.797739214036137	1.6875\\
0.841267259799942	2.375\\
0.775937312771844	3.3125\\
0.853553040015022	2.4375\\
0.957552939212352	2.3125\\
0.790754278223332	2.1875\\
0.73396421600852	1.375\\
0.846565874012297	1.8125\\
0.787436308198502	1.5625\\
0.878288830077542	2.25\\
0.782196197544811	2\\
0.748251784172454	1.875\\
0.810216549065943	2\\
0.870290443570459	2.875\\
0.808466040019231	3.5\\
0.885382110738798	2.9375\\
0.976424631091675	2.6875\\
0.813223494824524	2.5\\
0.75836454412097	1.125\\
0.894859527011322	2.25\\
0.634792637004848	2.375\\
0.76393380779676	1.75\\
0.839489992039372	1.75\\
0.510528731110892	1.3125\\
0.678895938957246	1.3125\\
0.864107741078469	2.75\\
0.677261847139248	1.0625\\
0.704357963197439	2.625\\
0.825069081899272	2.375\\
0.630034719199334	2.125\\
0.809431963115699	2.9375\\
0.65542157680135	2.25\\
0.733213163810971	2.8125\\
0.841765200804115	2\\
0.906433404297071	3.125\\
0.734123397837831	1.5\\
0.810220006048125	1.5\\
0.944146536337942	4\\
0.667501130094346	2.8125\\
0.879362419223056	3.5\\
0.998438341164503	4.4375\\
0.750966352384181	1.5625\\
0.804498375926948	4.9375\\
0.945176914739697	4\\
0.574395761651226	2.1875\\
0.615489635218354	2.125\\
0.591800008039856	3\\
0.567091450589917	1.1875\\
0.6001244333837	1.8125\\
0.713285180718411	3.5625\\
0.568732812573495	2.6875\\
0.711058798377014	3.625\\
0.735901837822678	2.25\\
0.67882695887234	2.1875\\
0.760067416087641	3.0625\\
0.574374323100365	1.6875\\
0.801850152987456	2.625\\
0.830549143364254	2.875\\
0.767960285973615	2.1875\\
0.728001912113096	2.125\\
0.796897907768198	1.875\\
0.884767858706526	3.8125\\
0.746135948476668	2.875\\
0.891421402960726	2.875\\
0.956865379966564	2.9375\\
0.793760372599782	2.125\\
0.934884711747574	2.25\\
0.942563430816983	3.4375\\
0.61737217045783	1.4375\\
0.694046030579828	1.9375\\
0.526072905744636	1.625\\
0.588951435450594	1.3125\\
0.736281401353941	1.3125\\
0.800340799444308	2.3125\\
0.57611885461439	1.4375\\
0.813263541544057	1.75\\
0.874353850280202	2\\
0.6262341345438	1.3125\\
0.760095121026572	2.875\\
0.7310924587782	1.25\\
0.800275414756974	1.8125\\
0.894263275836727	1.875\\
0.703158987827224	2.1875\\
0.74332067166696	2.5625\\
0.828560901682327	1.625\\
0.895115850517517	3.75\\
0.736570334137318	2.625\\
0.848148256585353	3.4375\\
0.922397075235244	3.0625\\
0.794254993579659	2.25\\
0.887187439921465	2.5\\
0.818501973464918	1.875\\
0.786155601241534	2.25\\
0.884607722089871	4.625\\
0.853293454440202	4.5625\\
0.823080544677155	4\\
0.80710871746095	2.0625\\
0.727318440912463	3.625\\
0.718734093319194	3.4375\\
0.83115896236995	4.5\\
0.830866258579114	4.5625\\
0.837012369702731	3.125\\
0.910298925021593	2.375\\
0.942959100667822	4.0625\\
0.513458828986096	1.75\\
0.887222260846279	2.9375\\
0.929695359640569	3.125\\
0.755638839978077	1.4375\\
0.627912742133773	1.6875\\
0.916474195074499	4.125\\
0.684105631610979	2.3125\\
0.881029459371965	3.3125\\
1.01019507806632	2.75\\
0.869116310277673	2.5625\\
0.810645229154846	4.625\\
0.93625932872635	4.375\\
0.569681540823714	1.625\\
0.880856061678658	3.0625\\
0.923446100221709	3\\
0.791955044794424	1.6875\\
0.718511807630633	1.625\\
0.952127793946952	4\\
0.639812599707328	2.3125\\
0.880346136040359	3.5625\\
1.00637352127649	2.5\\
0.870330863491301	2.625\\
0.902849650807187	4.5625\\
0.959594468384664	3.375\\
0.782292317450628	1.5625\\
0.705829386145144	1.75\\
0.858976234387135	2.0625\\
0.559456498511115	1.25\\
0.773086116314493	1.8125\\
0.947474671057216	1.8125\\
0.726958414679308	2.25\\
0.735936881576772	1.8125\\
0.908800396314358	2.5625\\
0.510812474439626	1.25\\
0.845932805223584	3.75\\
0.795616782440143	1.75\\
0.639764732490575	2.5625\\
0.896276571738051	3.375\\
0.941797191860202	3.1875\\
0.853611442537905	1.875\\
0.722801433957927	1.5\\
0.95393664479402	4.6875\\
0.736744835054706	3.5625\\
0.912774965985385	3.625\\
1.02693848426047	4.5\\
0.828759291023442	1.625\\
0.895357083568935	4.75\\
0.974960597262546	4.125\\
0.697933899780835	2.1875\\
0.771898044987417	1.6875\\
0.861007183313221	2.875\\
0.669432490717849	2\\
0.732033869193772	2.0625\\
0.912844826955452	2.1875\\
0.749110993969413	3.8125\\
0.844900869266337	3.5\\
0.960905186734006	2.6875\\
0.726770400957119	2.125\\
0.889516681217817	5\\
0.93986090057217	3\\
0.739581625809704	1.4375\\
0.687573023413816	1.25\\
0.781010510618825	1.875\\
0.660025522120964	1.625\\
0.778266560680428	1.625\\
0.92222134037801	2.375\\
0.721435536678933	2.0625\\
0.849600363956746	1.625\\
0.956346048500777	2.0625\\
0.570122182564463	1.125\\
0.880643102952002	2.75\\
0.838993165198604	1.4375\\
};
\end{axis}
\end{tikzpicture}%
 & \hspace*{-0.2in}
%
%
\begin{tikzpicture}

\begin{axis}[%
width=0.970772\figurewidth,
height=\figureheight,
at={(0\figurewidth,0\figureheight)},
scale only axis,
every outer x axis line/.append style={black},
every x tick label/.append style={font=\color{black}},
xmin=0.4,
xmax=1.1,
xlabel={Metric Score},
every outer y axis line/.append style={black},
every y tick label/.append style={font=\color{black}},
ymin=1,
ymax=5,
ylabel={MOS},
title={SROCC 0.5883 KROCC 0.4222 PLCC 0.5846},
axis x line*=bottom,
axis y line*=left,
title style={font=\scriptsize},xlabel style={font=\scriptsize},ylabel style={font=\scriptsize}
]
\addplot[only marks,mark=o,mark options={},mark size=1.0000pt,color=blue] plot table[row sep=crcr,]{%
0.849623584986059	1.3125\\
0.928994678649666	2.25\\
0.823961820976932	1.6875\\
0.85387119365117	2.1875\\
0.81399952558891	1.5\\
0.8152954013012	2.1875\\
0.84884816934079	1.9375\\
0.844378844194786	1.9375\\
0.893708868454173	1.9375\\
0.742557756682233	1.625\\
0.755487460032335	1.5\\
0.907706062573236	2.3125\\
1.00004871890266	4.9375\\
1.00006880932562	5\\
1.00001333540341	5\\
0.999982447947032	4.9375\\
1.00020911876446	5\\
0.999989009594075	5\\
0.999991960702137	5\\
1.00006457685043	5\\
1.00003701930736	5\\
0.999853007349338	5\\
1.00018989143436	5\\
0.999958248815572	5\\
0.795579442060354	1.4375\\
0.72218491174853	1.5625\\
0.852634518766198	2.375\\
0.653282441744835	1.3125\\
0.754280686619873	1.5\\
0.922822511579195	2.5625\\
0.748866677050518	2.0625\\
0.779643113076876	1.5\\
0.884339151220993	1.6875\\
0.619885836092414	1.25\\
0.806204321902804	2.875\\
0.809490286103582	1.4375\\
0.749762025612225	2.5625\\
0.834108923915053	2\\
0.887603981187621	3\\
0.797565773339512	2.6875\\
0.778947984434895	1.6875\\
0.920456200889777	3.5625\\
0.705570141618288	3.9375\\
0.877515694888533	2.75\\
0.956456688752588	3.3125\\
0.680026507846849	2.0625\\
0.791624004304731	4.9375\\
0.937930938708625	3.0625\\
0.833524581629979	2.6875\\
0.859885952989345	3.1875\\
0.772793079298506	2.8125\\
0.85572457007435	1.9375\\
0.772529563591279	2.25\\
0.932627601488854	3.9375\\
0.733380359665417	4\\
0.816507483639032	4.625\\
0.959415824356203	2.75\\
0.697585031961682	2.8125\\
0.76979691474184	5\\
0.92791308930554	4.4375\\
0.720318516537964	2.6875\\
0.892508928999597	2.625\\
0.939356262206444	2.75\\
0.745504455691747	1.9375\\
0.688582968770693	1.1875\\
0.919747598400529	3.9375\\
0.792639499143253	3.25\\
0.841852204665605	2.25\\
0.932285834789382	2.5\\
0.788335080704829	2.625\\
0.840278552732686	4.3125\\
0.922464136851967	2.875\\
0.858832177422644	2.125\\
0.933142209638089	4.125\\
0.82019609538501	3.25\\
0.855243121418185	4.125\\
0.838586232103933	2.25\\
0.8273332169588	4.6875\\
0.873124580857934	3.5\\
0.853469074699983	4.5625\\
0.916982681159051	4.0625\\
0.737703639670354	2.9375\\
0.781194876421607	1.8125\\
0.915215805000512	4.25\\
0.812667770660676	1.375\\
0.810460954555214	1.4375\\
0.700282026703648	1.6875\\
0.718354319689321	1.5\\
0.824581997897218	1.4375\\
0.863650567330705	2.6875\\
0.817239357635768	2.125\\
0.838380653520883	2.4375\\
0.889060462306538	2\\
0.699609836137209	1.625\\
0.845081858592845	2.3125\\
0.862347089400351	1.6875\\
0.765137300425663	1.625\\
0.824620073803964	1.875\\
0.777403452541061	1.875\\
0.721010939835201	1.8125\\
0.776929287527994	1.6875\\
0.836956663281311	2.375\\
0.780673387009997	3.3125\\
0.830278646528657	2.4375\\
0.903593193189132	2.3125\\
0.722816299729178	2.1875\\
0.670341460028126	1.375\\
0.81812904190029	1.8125\\
0.786257750612642	1.5625\\
0.867609240984917	2.25\\
0.795332049881751	2\\
0.756606355219991	1.875\\
0.79154641679573	2\\
0.860063114530499	2.875\\
0.8084010228146	3.5\\
0.845706042191092	2.9375\\
0.918715334077279	2.6875\\
0.744586720113439	2.5\\
0.686598773099531	1.125\\
0.86548090884373	2.25\\
0.658355869344995	2.375\\
0.759254235980736	1.75\\
0.848175077872648	1.75\\
0.539435235191918	1.3125\\
0.689121502464625	1.3125\\
0.88093058631671	2.75\\
0.709899050511578	1.0625\\
0.721901786913332	2.625\\
0.800902009053964	2.375\\
0.542716997758915	2.125\\
0.773797934595656	2.9375\\
0.651721649165493	2.25\\
0.74333270538356	2.8125\\
0.84064479040489	2\\
0.92791628957799	3.125\\
0.75229410981478	1.5\\
0.792327259666688	1.5\\
0.946155729897783	4\\
0.726250051783881	2.8125\\
0.860420580777894	3.5\\
0.959048067321564	4.4375\\
0.735234196666117	1.5625\\
0.809373175953444	4.9375\\
0.914481809756232	4\\
0.570835713866978	2.1875\\
0.610881963184446	2.125\\
0.625814640953749	3\\
0.573949326673283	1.1875\\
0.584784826031972	1.8125\\
0.724249493520783	3.5625\\
0.543017666834683	2.6875\\
0.706123295845213	3.625\\
0.709728142954999	2.25\\
0.585005132791084	2.1875\\
0.683386143780543	3.0625\\
0.504097090940821	1.6875\\
0.796548453928874	2.625\\
0.812088865624104	2.875\\
0.771992482418588	2.1875\\
0.736050183183947	2.125\\
0.779785063190212	1.875\\
0.878749080956693	3.8125\\
0.753202421915951	2.875\\
0.85452122650883	2.875\\
0.893063283628284	2.9375\\
0.728238828785267	2.125\\
0.853162629534563	2.25\\
0.904706399048554	3.4375\\
0.635926724545272	1.4375\\
0.695018417832505	1.9375\\
0.553793371079407	1.625\\
0.616000409040751	1.3125\\
0.734961191943842	1.3125\\
0.835421642858412	2.3125\\
0.575445778787772	1.4375\\
0.847621778896486	1.75\\
0.841639358436607	2\\
0.571347431434264	1.3125\\
0.723492590029981	2.875\\
0.702881032161498	1.25\\
0.810539230343784	1.8125\\
0.887536737986523	1.875\\
0.711565861697532	2.1875\\
0.749896404031935	2.5625\\
0.811351363150078	1.625\\
0.913730833618174	3.75\\
0.766967738979904	2.625\\
0.829439848447277	3.4375\\
0.871690528787608	3.0625\\
0.731606570468243	2.25\\
0.818282321105095	2.5\\
0.791424992865325	1.875\\
0.822248336475017	2.25\\
0.921540017243171	4.625\\
0.918427529644402	4.5625\\
0.89091083124271	4\\
0.867018910378635	2.0625\\
0.861257802287444	3.625\\
0.881642484579235	3.4375\\
0.891870398065572	4.5\\
0.8971644896085	4.5625\\
0.794277443807092	3.125\\
0.85892457924293	2.375\\
0.934767472212125	4.0625\\
0.58793056301047	1.75\\
0.87105689313392	2.9375\\
0.927661084821826	3.125\\
0.772061938730736	1.4375\\
0.652493137249828	1.6875\\
0.925372994895003	4.125\\
0.702840442734739	2.3125\\
0.86467652166302	3.3125\\
0.94156709471087	2.75\\
0.787982163591144	2.5625\\
0.805440420238082	4.625\\
0.909277824906775	4.375\\
0.564142260267922	1.625\\
0.866182522871744	3.0625\\
0.915993716002415	3\\
0.799027673638712	1.6875\\
0.70216565146799	1.625\\
0.926664576938439	4\\
0.653510180319136	2.3125\\
0.84565639712151	3.5625\\
0.939398701921866	2.5\\
0.777496749325135	2.625\\
0.832448204499111	4.5625\\
0.919182079189002	3.375\\
0.776685006083768	1.5625\\
0.699226443592398	1.75\\
0.86287538400524	2.0625\\
0.583195039609494	1.25\\
0.748273001699682	1.8125\\
0.930261416076737	1.8125\\
0.726163443068919	2.25\\
0.748245957956132	1.8125\\
0.865020499315911	2.5625\\
0.446288396198119	1.25\\
0.772318625322059	3.75\\
0.767187423692711	1.75\\
0.639546785970293	2.5625\\
0.881389879870266	3.375\\
0.945767997046388	3.1875\\
0.864932952072058	1.875\\
0.737677887754759	1.5\\
0.932264377989998	4.6875\\
0.736642603619116	3.5625\\
0.88227418020887	3.625\\
0.962577753359021	4.5\\
0.758845802420353	1.625\\
0.815771332222776	4.75\\
0.938931234588401	4.125\\
0.708228738726997	2.1875\\
0.767440394309109	1.6875\\
0.869946838512267	2.875\\
0.680792924230454	2\\
0.731018685381831	2.0625\\
0.898612571397586	2.1875\\
0.758010963583348	3.8125\\
0.834618959719955	3.5\\
0.903863347717759	2.6875\\
0.656202527927819	2.125\\
0.780742283075471	5\\
0.897190923219038	3\\
0.735788026429647	1.4375\\
0.667143004056926	1.25\\
0.794416190811195	1.875\\
0.668618859453916	1.625\\
0.761250348281487	1.625\\
0.901118104642267	2.375\\
0.724710260160232	2.0625\\
0.80753998715244	1.625\\
0.889146755555671	2.0625\\
0.512349461986674	1.125\\
0.795539480431386	2.75\\
0.806565606930311	1.4375\\
};
\end{axis}
\end{tikzpicture}%
 &  \hspace*{-0.2in}
%
%
\begin{tikzpicture}

\begin{axis}[%
width=0.970772\figurewidth,
height=\figureheight,
at={(0\figurewidth,0\figureheight)},
scale only axis,
every outer x axis line/.append style={black},
every x tick label/.append style={font=\color{black}},
xmin=0.4,
xmax=1.1,
xlabel={Metric Score},
every outer y axis line/.append style={black},
every y tick label/.append style={font=\color{black}},
ymin=1,
ymax=5,
ylabel={MOS},
title={SROCC 0.5455 KROCC 0.3871 PLCC 0.5631},
axis x line*=bottom,
axis y line*=left,
title style={font=\scriptsize},xlabel style={font=\scriptsize},ylabel style={font=\scriptsize}
]
\addplot[only marks,mark=o,mark options={},mark size=1.0000pt,color=blue] plot table[row sep=crcr,]{%
0.549927149017163	1.3125\\
0.802815544016483	2.25\\
0.53267807383128	1.6875\\
0.649714239627108	2.1875\\
0.522542842259011	1.5\\
0.604459318517163	2.1875\\
0.578447935415623	1.9375\\
0.66199187619014	1.9375\\
0.689830025286192	1.9375\\
0.539782725810065	1.625\\
0.447388179631726	1.5\\
0.784783735745018	2.3125\\
1.00272066254786	4.9375\\
1.00829298153366	5\\
0.995048555495928	5\\
0.998144184313965	4.9375\\
1.00163067042565	5\\
1.00154833312147	5\\
0.992636505878073	5\\
0.997941416652653	5\\
1.00046351150035	5\\
0.999078388538029	5\\
0.999220185716369	5\\
1.00438394967576	5\\
0.740731694860667	1.4375\\
0.723687576007218	1.5625\\
0.757800405105422	2.375\\
0.62650396882215	1.3125\\
0.836124731341546	1.5\\
0.979890944946979	2.5625\\
0.806804867831228	2.0625\\
0.833508629699773	1.5\\
0.873804216327627	1.6875\\
0.722161490430751	1.25\\
0.923256127592417	2.875\\
0.865517551531476	1.4375\\
0.694866932073279	2.5625\\
0.802694175192299	2\\
0.797193980527951	3\\
0.763056950103266	2.6875\\
0.861386711446331	1.6875\\
0.944249515576091	3.5625\\
0.832307085977001	3.9375\\
0.903440656825254	2.75\\
0.979978210843316	3.3125\\
0.777955895274639	2.0625\\
0.929378998919604	4.9375\\
0.938111365808982	3.0625\\
0.804598693167107	2.6875\\
0.842054992764109	3.1875\\
0.577429445762616	2.8125\\
0.832030533149192	1.9375\\
0.851740851250964	2.25\\
0.928855661223334	3.9375\\
0.772477433377137	4\\
0.868832259778389	4.625\\
0.963952751743529	2.75\\
0.78686773073952	2.8125\\
0.921934966947594	5\\
0.905174979300672	4.4375\\
0.713056633334934	2.6875\\
0.835570675517224	2.625\\
0.837806332238091	2.75\\
0.702788465330394	1.9375\\
0.707689867295647	1.1875\\
0.850634802642409	3.9375\\
0.839953344752138	3.25\\
0.843458774030476	2.25\\
0.910494367692988	2.5\\
0.866592672382423	2.625\\
0.893345002068137	4.3125\\
0.925190453052991	2.875\\
0.540991343378399	2.125\\
0.836806865740723	4.125\\
0.502980332050004	3.25\\
0.703045143971852	4.125\\
0.534802826994088	2.25\\
0.736608362637963	4.6875\\
0.655029669303192	3.5\\
0.738432176058094	4.5625\\
0.784859732964689	4.0625\\
0.523358621124943	2.9375\\
0.461574804369322	1.8125\\
0.863555200276235	4.25\\
0.559914030690126	1.375\\
0.709525584790761	1.4375\\
0.491170364844036	1.6875\\
0.495209921353086	1.5\\
0.687314660250215	1.4375\\
0.809640735091806	2.6875\\
0.692838395649707	2.125\\
0.717980872600043	2.4375\\
0.807837182851702	2\\
0.533424815243367	1.625\\
0.845830616908239	2.3125\\
0.765572790139318	1.6875\\
0.460673627577409	1.625\\
0.741749105067311	1.875\\
0.533723127124735	1.875\\
0.512043488690636	1.8125\\
0.603876187335373	1.6875\\
0.819623438605937	2.375\\
0.706315676819717	3.3125\\
0.708233126083414	2.4375\\
0.833862070279789	2.3125\\
0.598046436245212	2.1875\\
0.567234775972766	1.375\\
0.736744448166047	1.8125\\
0.464700283128659	1.5625\\
0.784808514537489	2.25\\
0.514227564667856	2\\
0.548656420953925	1.875\\
0.559469462596018	2\\
0.819301062372243	2.875\\
0.709960760842134	3.5\\
0.723075758680963	2.9375\\
0.839621668176884	2.6875\\
0.604674105812544	2.5\\
0.534156150164395	1.125\\
0.784261298595065	2.25\\
0.59337545654448	2.375\\
0.796863905924972	1.75\\
0.812898092073692	1.75\\
0.575253120165379	1.3125\\
0.707315839984577	1.3125\\
0.898387321253542	2.75\\
0.781659538194096	1.0625\\
0.76812807895308	2.625\\
0.755855928909569	2.375\\
0.668877273277514	2.125\\
0.854175666669473	2.9375\\
0.721579264617239	2.25\\
0.71141585639751	2.8125\\
0.800884592692733	2\\
0.76169797529376	3.125\\
0.660572940385575	1.5\\
0.868316582748331	1.5\\
0.904018862269702	4\\
0.74240883876001	2.8125\\
0.857172330276881	3.5\\
0.923955713034316	4.4375\\
0.761211285317324	1.5625\\
0.813608046415846	4.9375\\
0.944462448208082	4\\
0.617339559988152	2.1875\\
0.704142841147634	2.125\\
0.643534977823306	3\\
0.675750836435224	1.1875\\
0.622068001647689	1.8125\\
0.760451265605314	3.5625\\
0.631078670637142	2.6875\\
0.779828376695967	3.625\\
0.774452621425934	2.25\\
0.672949875705952	2.1875\\
0.776772068008843	3.0625\\
0.609596804631732	1.6875\\
0.627759496562811	2.625\\
0.709898204043443	2.875\\
0.564945260940565	2.1875\\
0.606207172951502	2.125\\
0.658118015762441	1.875\\
0.804893044363533	3.8125\\
0.649022126196833	2.875\\
0.811595698716938	2.875\\
0.757690888267448	2.9375\\
0.524210195200724	2.125\\
0.810538851117777	2.25\\
0.834976107343527	3.4375\\
0.560890254883296	1.4375\\
0.726187838035718	1.9375\\
0.495635847103994	1.625\\
0.625396878446576	1.3125\\
0.774518760642941	1.3125\\
0.864065337767108	2.3125\\
0.676230294167883	1.4375\\
0.821482924738038	1.75\\
0.817606092236552	2\\
0.646210338965234	1.3125\\
0.776474058811303	2.875\\
0.723575110467393	1.25\\
0.572086875609479	1.8125\\
0.826947621744277	1.875\\
0.545856103499761	2.1875\\
0.576553060469918	2.5625\\
0.584487000670154	1.625\\
0.88851945715062	3.75\\
0.592984152915223	2.625\\
0.679604258434548	3.4375\\
0.779800668325615	3.0625\\
0.606608860648738	2.25\\
0.777322145598725	2.5\\
0.661912772014675	1.875\\
0.565920034982995	2.25\\
0.856892072732594	4.625\\
0.715418299231321	4.5625\\
0.715914734776692	4\\
0.709553114694503	2.0625\\
0.747233995829015	3.625\\
0.734910831889146	3.4375\\
0.811832582983931	4.5\\
0.769291148920545	4.5625\\
0.632788370434943	3.125\\
0.708848481713311	2.375\\
0.872147171922553	4.0625\\
0.579193529321309	1.75\\
0.837440221104279	2.9375\\
0.855132641855088	3.125\\
0.89764401196852	1.4375\\
0.698016215724821	1.6875\\
0.881625352605649	4.125\\
0.838900794908262	2.3125\\
0.878216647468836	3.3125\\
0.952358124598299	2.75\\
0.840874594577322	2.5625\\
0.824576643107043	4.625\\
0.860586061790977	4.375\\
0.57706872043031	1.625\\
0.848508962746866	3.0625\\
0.836814197401313	3\\
0.896933461801938	1.6875\\
0.820265121605482	1.625\\
0.900152555470187	4\\
0.816012940956142	2.3125\\
0.868738562357942	3.5625\\
0.935277119354212	2.5\\
0.823634322864661	2.625\\
0.951443499880221	4.5625\\
0.882134917812235	3.375\\
0.741051974394297	1.5625\\
0.737466443914156	1.75\\
0.76981933764418	2.0625\\
0.543478115185118	1.25\\
0.837814253319977	1.8125\\
0.966770336428669	1.8125\\
0.804230640516363	2.25\\
0.817249934456541	1.8125\\
0.855147712190088	2.5625\\
0.574114891658354	1.25\\
0.888959611718701	3.75\\
0.814649301148118	1.75\\
0.650107012551717	2.5625\\
0.859057157901995	3.375\\
0.808602216998769	3.1875\\
0.870315974873615	1.875\\
0.803496023247542	1.5\\
0.892828561085196	4.6875\\
0.816366367152134	3.5625\\
0.875637721326812	3.625\\
0.944729268080789	4.5\\
0.827090452594	1.625\\
0.914838837355043	4.75\\
0.917175175794398	4.125\\
0.691354006709892	2.1875\\
0.738872852179839	1.6875\\
0.720339839459742	2.875\\
0.655724153399117	2\\
0.783616071195812	2.0625\\
0.996317019461465	2.1875\\
0.836918234897334	3.8125\\
0.792067075987194	3.5\\
0.913611514052075	2.6875\\
0.673911477797428	2.125\\
0.927238071091271	5\\
0.867916042949787	3\\
0.663967825323162	1.4375\\
0.702263790686347	1.25\\
0.697272463322574	1.875\\
0.647867446077988	1.625\\
0.832135326906633	1.625\\
0.974039026627566	2.375\\
0.789863636126398	2.0625\\
0.874997862650326	1.625\\
0.872854199083225	2.0625\\
0.619185236970812	1.125\\
0.90892287825934	2.75\\
0.818255896235448	1.4375\\
};
\end{axis}
\end{tikzpicture}%
 \\
		(a) $AUC_{Borji}$~\cite{borjieval}&  (b) $AUC_{Judd}$~\cite{Judd_2012}& (c) $sAUC$~\cite{borjieval} \\
		\hspace*{-0.3in}
%
%
\begin{tikzpicture}

\begin{axis}[%
width=0.970772\figurewidth,
height=\figureheight,
at={(0\figurewidth,0\figureheight)},
scale only axis,
every outer x axis line/.append style={black},
every x tick label/.append style={font=\color{black}},
xmin=-0.2,
xmax=1.2,
xlabel={Metric Score},
every outer y axis line/.append style={black},
every y tick label/.append style={font=\color{black}},
ymin=1,
ymax=5,
ylabel={MOS},
title={SROCC 0.7461 KROCC 0.5726 PLCC 0.8216},
axis x line*=bottom,
axis y line*=left,
title style={font=\scriptsize},xlabel style={font=\scriptsize},ylabel style={font=\scriptsize}
]
\addplot[only marks,mark=o,mark options={},mark size=1.0000pt,color=blue] plot table[row sep=crcr,]{%
0.477184645065083	1.3125\\
0.513043425248779	2.25\\
0.373685076029503	1.6875\\
0.473124443083953	2.1875\\
0.439456119450898	1.5\\
0.338441372161096	2.1875\\
0.451687351026964	1.9375\\
0.452419821444659	1.9375\\
0.384054539350634	1.9375\\
0.233189325535716	1.625\\
0.281294255292245	1.5\\
0.423092403803064	2.3125\\
1	4.9375\\
1	5\\
1	5\\
1	4.9375\\
1	5\\
1	5\\
1	5\\
1	5\\
1	5\\
1	5\\
1	5\\
1	5\\
0.371722669489973	1.4375\\
0.23241468386076	1.5625\\
0.403603819440032	2.375\\
0.206620321403324	1.3125\\
0.285491825026076	1.5\\
0.517358926734937	2.5625\\
0.318463373543193	2.0625\\
0.412770508565865	1.5\\
0.406790892988039	1.6875\\
0.111247428088483	1.25\\
0.369609061897137	2.875\\
0.284810621199707	1.4375\\
0.346819576334025	2.5625\\
0.398319791048849	2\\
0.504952131390251	3\\
0.389196376883159	2.6875\\
0.300635838267343	1.6875\\
0.657931283815065	3.5625\\
0.303171626069589	3.9375\\
0.58025737341348	2.75\\
0.639109626442081	3.3125\\
0.160640043553877	2.0625\\
0.711151347821099	4.9375\\
0.525968105609998	3.0625\\
0.500073286481905	2.6875\\
0.51579958936047	3.1875\\
0.384254702183216	2.8125\\
0.478093524919854	1.9375\\
0.335483942934017	2.25\\
0.711911312459213	3.9375\\
0.328551114179563	4\\
0.619320723313175	4.625\\
0.638143329871382	2.75\\
0.236585735246421	2.8125\\
0.793990354734697	5\\
0.678903049032467	4.4375\\
0.237725087236687	2.6875\\
0.544834830442662	2.625\\
0.638329511122393	2.75\\
0.293972173013962	1.9375\\
0.109838566819475	1.1875\\
0.734992966834211	3.9375\\
0.388593625867534	3.25\\
0.450563471167336	2.25\\
0.562942638271292	2.5\\
0.320622311653007	2.625\\
0.754906177804972	4.3125\\
0.508737585157781	2.875\\
0.506650997344124	2.125\\
0.812067186315629	4.125\\
0.381059487565377	3.25\\
0.720715437420347	4.125\\
0.461121470034951	2.25\\
0.749291259282055	4.6875\\
0.656372205385747	3.5\\
0.801776084013981	4.5625\\
0.666834082746157	4.0625\\
0.260512438395029	2.9375\\
0.237363196270351	1.8125\\
0.804583570585577	4.25\\
0.442316175250604	1.375\\
0.381964797800762	1.4375\\
0.239231151588297	1.6875\\
0.321264106492475	1.5\\
0.495992993024947	1.4375\\
0.490957823428224	2.6875\\
0.43287536503176	2.125\\
0.555739440685688	2.4375\\
0.447091898877252	2\\
0.187355545241877	1.625\\
0.464246221186771	2.3125\\
0.427070709850074	1.6875\\
0.351421709757169	1.625\\
0.414409038762406	1.875\\
0.33120763743686	1.875\\
0.223838973295822	1.8125\\
0.380564508832354	1.6875\\
0.380428915838135	2.375\\
0.392004656888817	3.3125\\
0.477985792842731	2.4375\\
0.436851991690543	2.3125\\
0.205371668157338	2.1875\\
0.0994801341923554	1.375\\
0.301900782331713	1.8125\\
0.387379682521488	1.5625\\
0.494764938268	2.25\\
0.355767778472331	2\\
0.309714374919788	1.875\\
0.41055165514347	2\\
0.467278378948831	2.875\\
0.459454132348809	3.5\\
0.554439419897038	2.9375\\
0.503414135116063	2.6875\\
0.238518262725313	2.5\\
0.139129946835155	1.125\\
0.400791367902602	2.25\\
0.186589634890806	2.375\\
0.324464027368034	1.75\\
0.38855015206287	1.75\\
0.0107047919875673	1.3125\\
0.162822693559277	1.3125\\
0.469351918515041	2.75\\
0.208371729949148	1.0625\\
0.273077252910011	2.625\\
0.317324027180318	2.375\\
0.0542222531808925	2.125\\
0.319418481798783	2.9375\\
0.193746807898188	2.25\\
0.291955342568026	2.8125\\
0.395023068080753	2\\
0.61156224022933	3.125\\
0.208284345995753	1.5\\
0.329435031480498	1.5\\
0.685247508573507	4\\
0.207140826531529	2.8125\\
0.582034888285648	3.5\\
0.741605230923656	4.4375\\
0.216350990589281	1.5625\\
0.757405284420746	4.9375\\
0.524012250183784	4\\
0.0797186321427834	2.1875\\
0.209462325636085	2.125\\
0.208421754590563	3\\
0.0390001422093684	1.1875\\
0.100809153441038	1.8125\\
0.491886620761257	3.5625\\
0.145035168744387	2.6875\\
0.394154271120199	3.625\\
0.237311126174447	2.25\\
0.119656584419799	2.1875\\
0.387602376115185	3.0625\\
0.0573133831507201	1.6875\\
0.42266784004409	2.625\\
0.370911040556051	2.875\\
0.320730007494379	2.1875\\
0.298157955939058	2.125\\
0.381534132822064	1.875\\
0.539128466286209	3.8125\\
0.308745360192062	2.875\\
0.538466593499636	2.875\\
0.484326794952063	2.9375\\
0.190026066302078	2.125\\
0.427520495540029	2.25\\
0.523522071048746	3.4375\\
0.148327609339964	1.4375\\
0.164250517566642	1.9375\\
0.0272143366505244	1.625\\
0.0688072955449886	1.3125\\
0.268519641550315	1.3125\\
0.345546771778683	2.3125\\
0.0755297203744188	1.4375\\
0.417016021371656	1.75\\
0.330321689204014	2\\
0.0077014206337979	1.3125\\
0.283023422724002	2.875\\
0.236882219479857	1.25\\
0.421175416569085	1.8125\\
0.549760913855331	1.875\\
0.259971804321008	2.1875\\
0.361184969848175	2.5625\\
0.457160812207626	1.625\\
0.615020283419865	3.75\\
0.359956642088235	2.625\\
0.555600422798387	3.4375\\
0.525277748994063	3.0625\\
0.2558211902846	2.25\\
0.42511885660282	2.5\\
0.37690903143933	1.875\\
0.363288376323606	2.25\\
0.757789640389955	4.625\\
0.632003091963068	4.5625\\
0.681330885864079	4\\
0.432216075497369	2.0625\\
0.647144107797493	3.625\\
0.53204136514389	3.4375\\
0.762035914291884	4.5\\
0.766644233801025	4.5625\\
0.375454684948995	3.125\\
0.541876147158181	2.375\\
0.812097324155752	4.0625\\
-0.0167947064075803	1.75\\
0.511896265526255	2.9375\\
0.634189410746681	3.125\\
0.326894146515928	1.4375\\
0.10917399312034	1.6875\\
0.743058050512394	4.125\\
0.238559954261739	2.3125\\
0.684013849300077	3.3125\\
0.545675105638071	2.75\\
0.347289158072662	2.5625\\
0.761552324068714	4.625\\
0.657863925404952	4.375\\
0.0276872068032686	1.625\\
0.491697843809088	3.0625\\
0.558262056426087	3\\
0.392528211036065	1.6875\\
0.204511935940306	1.625\\
0.719064565202507	4\\
0.178125178863938	2.3125\\
0.612538161898505	3.5625\\
0.517962419331586	2.5\\
0.2975572434246	2.625\\
0.70153395280628	4.5625\\
0.546241510039014	3.375\\
0.326945113035797	1.5625\\
0.173302938446884	1.75\\
0.387523452538818	2.0625\\
-0.0899908041318704	1.25\\
0.290624258293477	1.8125\\
0.530379919823133	1.8125\\
0.294358372770391	2.25\\
0.26681868693775	1.8125\\
0.42152921363263	2.5625\\
-0.0810962874584914	1.25\\
0.429264822593893	3.75\\
0.23363595839837	1.75\\
0.17762039540768	2.5625\\
0.517210381129082	3.375\\
0.677761709364079	3.1875\\
0.516958777545498	1.875\\
0.158901876178473	1.5\\
0.700261896242344	4.6875\\
0.342532062286413	3.5625\\
0.674321404594885	3.625\\
0.754917746735121	4.5\\
0.230910668841279	1.625\\
0.790223088270152	4.75\\
0.694462771767374	4.125\\
0.223902328934329	2.1875\\
0.255448228735278	1.6875\\
0.467008638495232	2.875\\
0.142981210835201	2\\
0.26251994730583	2.0625\\
0.497409083398878	2.1875\\
0.337515776509988	3.8125\\
0.505446815846499	3.5\\
0.453863568615555	2.6875\\
0.111959550348115	2.125\\
0.655069438710719	5\\
0.456775012111405	3\\
0.297095936516126	1.4375\\
0.153272659658608	1.25\\
0.320411956524675	1.875\\
0.189216633787618	1.625\\
0.292016813189036	1.625\\
0.49400206222007	2.375\\
0.298160329947552	2.0625\\
0.443311840392522	1.625\\
0.437853296372253	2.0625\\
-0.0118996439490865	1.125\\
0.42592838392298	2.75\\
0.301400014995272	1.4375\\
};
\end{axis}
\end{tikzpicture}%
 & \hspace*{-0.2in}
%
%
\begin{tikzpicture}

\begin{axis}[%
width=0.970772\figurewidth,
height=\figureheight,
at={(0\figurewidth,0\figureheight)},
scale only axis,
every outer x axis line/.append style={black},
every x tick label/.append style={font=\color{black}},
xmin=0,
xmax=7,
xlabel={Metric Score},
every outer y axis line/.append style={black},
every y tick label/.append style={font=\color{black}},
ymin=0,
ymax=4,
ylabel={MOS},
title={SROCC 0.4470 KROCC 0.3168 PLCC 0.5683},
axis x line*=bottom,
axis y line*=left,
title style={font=\scriptsize},xlabel style={font=\scriptsize},ylabel style={font=\scriptsize}
]
\addplot[only marks,mark=o,mark options={},mark size=1.0000pt,color=blue] plot table[row sep=crcr,]{%
3.07058912735101	3.6875\\
3.06423644714193	2.75\\
4.19313855073823	3.3125\\
2.8386152921454	2.8125\\
2.35309920531649	3.5\\
3.13854919807047	2.8125\\
2.82938024012542	3.0625\\
3.08224955926729	3.0625\\
3.70613871978459	3.0625\\
3.53561122817948	3.375\\
3.40967886684988	3.5\\
3.38507023172479	2.6875\\
0.380749715213921	0.0625\\
0.320097909222938	0\\
0.26649257632037	0\\
0.336286480081975	0.0625\\
0.35489627475288	0\\
0.319893854420137	0\\
0.252286954254883	0\\
0.297050529087293	0\\
0.395270604792762	0\\
0.386293667254274	0\\
0.372404795416128	0\\
0.366069221077083	0\\
2.8933222440949	3.5625\\
3.30600939000522	3.4375\\
4.20676409898262	2.625\\
3.16971945527445	3.6875\\
2.78436654803201	3.5\\
2.99140803477601	2.4375\\
3.21073199118072	2.9375\\
2.87576279702949	3.5\\
3.70645902851002	3.3125\\
3.63891883137705	3.75\\
3.22765310109965	2.125\\
3.78347670470326	3.5625\\
2.93692692320731	2.4375\\
3.10240039468479	3\\
4.72087891875721	2\\
3.21891845988911	2.3125\\
3.15961078452898	3.3125\\
2.86779415713272	1.4375\\
3.41028306227062	1.0625\\
2.69633412834087	2.25\\
3.46615883210293	1.6875\\
3.62853677683348	2.9375\\
3.31502810640521	0.0625\\
3.41623495258406	1.9375\\
2.34032044348483	2.3125\\
2.67988251324859	1.8125\\
4.91690292339284	2.1875\\
3.01534261409363	3.0625\\
2.95828543843248	2.75\\
2.1940232579267	1.0625\\
3.25416111098526	1\\
2.72668497450106	0.375\\
2.59918858186756	2.25\\
3.26680116948346	2.1875\\
2.26271412630293	0\\
3.64929230590118	0.5625\\
3.36383819274601	2.3125\\
2.55956550765484	2.375\\
2.9489485258919	2.25\\
3.28693517052032	3.0625\\
6.02823316328444	3.8125\\
1.88720347956887	1.0625\\
3.01077469390413	1.75\\
2.5338111142895	2.75\\
2.81977514843439	2.5\\
2.72994410715179	2.375\\
2.19654272982449	0.6875\\
3.58095361172343	2.125\\
1.50266240270009	2.875\\
1.06652058812327	0.875\\
3.34919287759469	1.75\\
1.96785349694824	0.875\\
1.91684605539437	2.75\\
2.04403440917748	0.3125\\
2.0002038797744	1.5\\
2.32372724680709	0.4375\\
2.05165057561809	0.9375\\
3.31391378364219	2.0625\\
2.6384871038364	3.1875\\
1.31826048721482	0.75\\
2.12983912749078	3.625\\
2.23475101622924	3.5625\\
4.05434585466846	3.3125\\
2.32672308543885	3.5\\
1.73462797074424	3.5625\\
2.34191178022078	2.3125\\
2.36824428096728	2.875\\
2.45778517382738	2.5625\\
2.64676646403159	3\\
3.40376931244152	3.375\\
2.80214978240424	2.6875\\
2.39175671571035	3.3125\\
2.45970967415631	3.375\\
2.34952361160559	3.125\\
4.4622587545984	3.125\\
2.97326558616552	3.1875\\
2.1194045144603	3.3125\\
2.51139017887813	2.625\\
2.84051801419972	1.6875\\
2.36904038730544	2.5625\\
2.83298714809818	2.6875\\
3.25032735352734	2.8125\\
4.15619706118709	3.625\\
2.74218055247121	3.1875\\
2.2737336582852	3.4375\\
2.13277150067861	2.75\\
4.31815526203701	3\\
2.80369092937235	3.125\\
2.00398078807761	3\\
2.30218223246474	2.125\\
2.67494122325589	1.5\\
2.24052499877421	2.0625\\
2.66067077355104	2.3125\\
3.18436488807977	2.5\\
3.99736558513715	3.875\\
2.50970930011696	2.75\\
3.15226578344141	2.625\\
3.32402500796211	3.25\\
3.82145570802503	3.25\\
3.79826382007674	3.6875\\
2.9513203477971	3.6875\\
3.1974552031558	2.25\\
3.60253336398428	3.9375\\
3.02011399028668	2.375\\
3.76197353657862	2.625\\
3.90810782050534	2.875\\
3.50123311180495	2.0625\\
4.03686715440447	2.75\\
2.70557410655124	2.1875\\
2.56253113387938	3\\
3.0246774211979	1.875\\
3.29615892791089	3.5\\
3.0554484476863	3.5\\
1.61237988161687	1\\
3.56795380039539	2.1875\\
2.17266852984942	1.5\\
2.12239712128448	0.5625\\
3.12986143915413	3.4375\\
2.69187970526961	0.0625\\
3.04269830942959	1\\
3.70797109233256	2.8125\\
3.36872245409649	2.875\\
4.37410213109413	2\\
3.65339843660765	3.8125\\
3.2402048175712	3.1875\\
2.81808328674046	1.4375\\
3.41616531506636	2.3125\\
3.22903951331191	1.375\\
3.92590853714235	2.75\\
3.43597809835288	2.8125\\
3.35508146368037	1.9375\\
3.94945311775127	3.3125\\
2.20117851485052	2.375\\
2.47708656252005	2.125\\
4.93267981365761	2.8125\\
2.43335573013555	2.875\\
2.15398421841333	3.125\\
1.9579337683115	1.1875\\
3.04605902976956	2.125\\
2.75340041727099	2.125\\
2.85341422128143	2.0625\\
3.40013205185224	2.875\\
2.79529443332695	2.75\\
2.54204491417863	1.5625\\
3.01707503599206	3.5625\\
3.33046037506361	3.0625\\
5.33151532513131	3.375\\
3.31528657098415	3.6875\\
2.82104760277065	3.6875\\
3.63815766004526	2.6875\\
3.71720190331157	3.5625\\
2.75880975377705	3.25\\
3.71292623741959	3\\
4.12660612865571	3.6875\\
3.09820104283322	2.125\\
3.66332676338032	3.75\\
2.02683829800825	3.1875\\
1.87113770070074	3.125\\
4.16195403832858	2.8125\\
2.63714308226583	2.4375\\
1.95315332437298	3.375\\
1.84408074673601	1.25\\
2.74861913060568	2.375\\
3.14771565713592	1.5625\\
3.13030026489331	1.9375\\
3.28732914949052	2.75\\
3.16387461265793	2.5\\
2.64199594217848	3.125\\
1.93678994307289	2.75\\
1.205731806965	0.375\\
2.2305245049873	0.4375\\
2.45431152332261	1\\
3.11460677387093	2.9375\\
2.78198650771472	1.375\\
2.34178234203036	1.5625\\
2.94139974492325	0.5\\
2.58789309279622	0.4375\\
2.82981027046014	1.875\\
2.39822877844753	2.625\\
1.37631206890649	0.9375\\
3.84098655864303	3.25\\
2.63136253889083	2.0625\\
2.71012178563284	1.875\\
3.14780258579192	3.5625\\
4.33575467597873	3.3125\\
1.33302935006273	0.875\\
3.86453488867004	2.6875\\
1.98720420798467	1.6875\\
2.7294848779014	2.25\\
2.65886330099272	2.4375\\
2.90690164407289	0.375\\
2.36944489730244	0.625\\
3.55298715009779	3.375\\
2.97154913339146	1.9375\\
3.63576102726191	2\\
2.96938284802239	3.3125\\
3.25343161805176	3.375\\
2.05160878777575	1\\
3.80996105952775	2.6875\\
2.65539962596799	1.4375\\
3.22752830217765	2.5\\
3.04366102991328	2.375\\
2.76122933520718	0.4375\\
3.40603365701526	1.625\\
3.09719001909742	3.4375\\
3.5226610007231	3.25\\
4.29008264896491	2.9375\\
4.28820084769682	3.75\\
3.00345290533967	3.1875\\
3.11479483658116	3.1875\\
3.33289462195291	2.75\\
3.13533244444828	3.1875\\
3.88816854830121	2.4375\\
4.31797367987131	3.75\\
3.33541837324085	1.25\\
3.87353447732668	3.25\\
3.2133623370467	2.4375\\
2.9290929179678	1.625\\
3.26084549677061	1.8125\\
2.8418729600438	3.125\\
3.63654424993193	3.5\\
2.06623829254141	0.3125\\
3.10636748401223	1.4375\\
2.49129164537458	1.375\\
2.76921024038534	0.5\\
3.20158637953227	3.375\\
2.39691275922639	0.25\\
3.32481128166184	0.875\\
2.95845544498241	2.8125\\
3.36686294138434	3.3125\\
4.07885174120331	2.125\\
3.6293501952136	3\\
3.59297504059804	2.9375\\
3.39230578749254	2.8125\\
3.20498078731779	1.1875\\
2.81659370972306	1.5\\
3.33946042024565	2.3125\\
3.90702785841444	2.875\\
2.75868161739372	0\\
3.5495315293786	2\\
3.04171677452503	3.5625\\
3.58788835332946	3.75\\
4.759287728564	3.125\\
3.300629756967	3.375\\
2.89789873635529	3.375\\
3.3980664892	2.625\\
3.34512228335198	2.9375\\
2.97448533026986	3.375\\
3.63866207093595	2.9375\\
3.94320630198927	3.875\\
3.04598909513615	2.25\\
3.84453870565833	3.5625\\
};
\end{axis}
\end{tikzpicture}%
 & \hspace*{-0.2in}
%
%
\begin{tikzpicture}

\begin{axis}[%
width=0.970772\figurewidth,
height=\figureheight,
at={(0\figurewidth,0\figureheight)},
scale only axis,
every outer x axis line/.append style={black},
every x tick label/.append style={font=\color{black}},
xmin=-0.2,
xmax=1.2,
xlabel={Metric Score},
every outer y axis line/.append style={black},
every y tick label/.append style={font=\color{black}},
ymin=1,
ymax=5,
ylabel={MOS},
title={SROCC 0.7563 KROCC 0.5810 PLCC 0.8297},
axis x line*=bottom,
axis y line*=left,
title style={font=\scriptsize},xlabel style={font=\scriptsize},ylabel style={font=\scriptsize}
]
\addplot[only marks,mark=o,mark options={},mark size=1.0000pt,color=blue] plot table[row sep=crcr,]{%
0.369185719055705	1.3125\\
0.432375397940944	2.25\\
0.322024230989887	1.6875\\
0.37595957486903	2.1875\\
0.302145552187649	1.5\\
0.27048282663757	2.1875\\
0.376778982924986	1.9375\\
0.345318447909135	1.9375\\
0.287143129036215	1.9375\\
0.171391274887165	1.625\\
0.209731264430382	1.5\\
0.334894214363062	2.3125\\
1	4.9375\\
1	5\\
1	5\\
1	4.9375\\
1	5\\
1	5\\
1	5\\
1	5\\
1	5\\
1	5\\
1	5\\
1	5\\
0.322401789533399	1.4375\\
0.211837699985427	1.5625\\
0.387561764474563	2.375\\
0.16111168978574	1.3125\\
0.236137096027207	1.5\\
0.477305254266113	2.5625\\
0.270321492087833	2.0625\\
0.36056877964089	1.5\\
0.331343772425768	1.6875\\
0.0788765973377849	1.25\\
0.349764228202509	2.875\\
0.283168956573787	1.4375\\
0.29296256569017	2.5625\\
0.362376004069282	2\\
0.46373680020983	3\\
0.341108696456561	2.6875\\
0.267752328599459	1.6875\\
0.589035894314998	3.5625\\
0.261140985321697	3.9375\\
0.496334245954834	2.75\\
0.530630174707381	3.3125\\
0.125489156213508	2.0625\\
0.663309006662986	4.9375\\
0.488135599382706	3.0625\\
0.429473473527577	2.6875\\
0.463445439055745	3.1875\\
0.358364288923188	2.8125\\
0.458832723635482	1.9375\\
0.288500255208369	2.25\\
0.636974494927993	3.9375\\
0.29230215982862	4\\
0.574784112080062	4.625\\
0.560573879207255	2.75\\
0.19233788673849	2.8125\\
0.711179676036818	5\\
0.678324118251586	4.4375\\
0.198594498975889	2.6875\\
0.495583514010859	2.625\\
0.593527424556916	2.75\\
0.257322792805476	1.9375\\
0.084775096465713	1.1875\\
0.649087425262973	3.9375\\
0.327398716427425	3.25\\
0.382324950775551	2.25\\
0.442076800755908	2.5\\
0.250464735979824	2.625\\
0.678008509386529	4.3125\\
0.457001765088592	2.875\\
0.405892421186462	2.125\\
0.710256356965039	4.125\\
0.336596973513309	3.25\\
0.594218697562977	4.125\\
0.339170543899857	2.25\\
0.642973395993513	4.6875\\
0.556700061577475	3.5\\
0.646433037217916	4.5625\\
0.520543398297229	4.0625\\
0.188603204612355	2.9375\\
0.18248749228394	1.8125\\
0.692115231118519	4.25\\
0.360186085130601	1.375\\
0.32215251399993	1.4375\\
0.179504166713608	1.6875\\
0.21334284628638	1.5\\
0.342596777844222	1.4375\\
0.400621213939806	2.6875\\
0.362842003955565	2.125\\
0.430368759510423	2.4375\\
0.351714813899068	2\\
0.135191899638091	1.625\\
0.396895174885363	2.3125\\
0.368647887985575	1.6875\\
0.273817070907922	1.625\\
0.36169756295057	1.875\\
0.291310600659618	1.875\\
0.178661018240596	1.8125\\
0.264611671437992	1.6875\\
0.35169175263208	2.375\\
0.343196002001771	3.3125\\
0.378806678181465	2.4375\\
0.33971168071865	2.3125\\
0.151167642747194	2.1875\\
0.0688815919291322	1.375\\
0.244990167974905	1.8125\\
0.299394512330602	1.5625\\
0.432087860498362	2.25\\
0.313429509654358	2\\
0.249312190707293	1.875\\
0.284521227307571	2\\
0.43131886007803	2.875\\
0.397667105971218	3.5\\
0.438610471267884	2.9375\\
0.393221214622962	2.6875\\
0.174738540767301	2.5\\
0.0989157756526795	1.125\\
0.3279193403623	2.25\\
0.151026351800616	2.375\\
0.335926993080716	1.75\\
0.436857042471104	1.75\\
0.0443084720709015	1.3125\\
0.13844221654863	1.3125\\
0.438985525548572	2.75\\
0.236376206459521	1.0625\\
0.20917925880283	2.625\\
0.254032577283772	2.375\\
0.0384012559828813	2.125\\
0.368126395554999	2.9375\\
0.167407223340905	2.25\\
0.23756818791796	2.8125\\
0.371264252203935	2\\
0.571705008608676	3.125\\
0.184760103768517	1.5\\
0.284494366131155	1.5\\
0.601154303840274	4\\
0.186160917880969	2.8125\\
0.504548061374713	3.5\\
0.605505969705256	4.4375\\
0.185642839906613	1.5625\\
0.745780501376119	4.9375\\
0.490290791612634	4\\
0.157977515774649	2.1875\\
0.184770581591059	2.125\\
0.171348531850809	3\\
0.0749563917476765	1.1875\\
0.126510149178767	1.8125\\
0.547931761107408	3.5625\\
0.151465677976244	2.6875\\
0.348367594097078	3.625\\
0.287547267401336	2.25\\
0.125719714367253	2.1875\\
0.395068543944434	3.0625\\
0.0407906698698232	1.6875\\
0.334819375326438	2.625\\
0.31568727326634	2.875\\
0.282195485727164	2.1875\\
0.234142666452579	2.125\\
0.274414322606294	1.875\\
0.460595066620983	3.8125\\
0.280483911430385	2.875\\
0.452654631457588	2.875\\
0.358146009151445	2.9375\\
0.139992898875873	2.125\\
0.349978661844543	2.25\\
0.443159864983486	3.4375\\
0.133708342181663	1.4375\\
0.108562856029337	1.9375\\
-0.0272644118312804	1.625\\
0.0507197473337344	1.3125\\
0.240659297510615	1.3125\\
0.353179952288135	2.3125\\
0.0475472637867307	1.4375\\
0.398724979637011	1.75\\
0.29978169289442	2\\
0.0284356056801166	1.3125\\
0.256845371855049	2.875\\
0.23920454771297	1.25\\
0.307571834142132	1.8125\\
0.459860467210289	1.875\\
0.180818582786836	2.1875\\
0.259995797221478	2.5625\\
0.345437700412384	1.625\\
0.626923383215133	3.75\\
0.288451886312208	2.625\\
0.417853605993784	3.4375\\
0.412512159966307	3.0625\\
0.219313321712912	2.25\\
0.312396855953934	2.5\\
0.297250980936516	1.875\\
0.287614244504304	2.25\\
0.67274706350301	4.625\\
0.570774682649722	4.5625\\
0.553102787274846	4\\
0.33321139293223	2.0625\\
0.649904395793082	3.625\\
0.465905264384765	3.4375\\
0.63759743553938	4.5\\
0.683787033473845	4.5625\\
0.293355650658384	3.125\\
0.474444970577824	2.375\\
0.722854987806673	4.0625\\
-0.00691366765959954	1.75\\
0.463426464962745	2.9375\\
0.567355308900197	3.125\\
0.282136828677181	1.4375\\
0.106793963583574	1.6875\\
0.629091081458545	4.125\\
0.20176653406993	2.3125\\
0.58432967462862	3.3125\\
0.432120047649694	2.75\\
0.265066520086578	2.5625\\
0.669256117899772	4.625\\
0.58054295148907	4.375\\
0.019739995723819	1.625\\
0.44657563530581	3.0625\\
0.496057902932503	3\\
0.336412085789198	1.6875\\
0.172147964806168	1.625\\
0.60813000481606	4\\
0.147968354351291	2.3125\\
0.510507169405908	3.5625\\
0.4065065046087	2.5\\
0.230450565051565	2.625\\
0.61395533957363	4.5625\\
0.474582265707931	3.375\\
0.310558992132037	1.5625\\
0.180076514671901	1.75\\
0.398384465979164	2.0625\\
-0.0769833113996024	1.25\\
0.253081072482837	1.8125\\
0.490548291846794	1.8125\\
0.280090320819111	2.25\\
0.207608937005972	1.8125\\
0.340199985390926	2.5625\\
-0.0718030289245185	1.25\\
0.441990537650733	3.75\\
0.222274764075026	1.75\\
0.146373366567537	2.5625\\
0.465738793095704	3.375\\
0.622066820053903	3.1875\\
0.450285002072159	1.875\\
0.136977983559794	1.5\\
0.605167148966614	4.6875\\
0.293472841699006	3.5625\\
0.563038742904704	3.625\\
0.624533386623731	4.5\\
0.177779322484489	1.625\\
0.680362362147514	4.75\\
0.620736485998028	4.125\\
0.179021415783935	2.1875\\
0.219143083929128	1.6875\\
0.403027255112502	2.875\\
0.129009756177676	2\\
0.228969289921254	2.0625\\
0.467259600952842	2.1875\\
0.297955780623685	3.8125\\
0.407661568046157	3.5\\
0.360783489068962	2.6875\\
0.079787148947799	2.125\\
0.550564374459848	5\\
0.399475998208915	3\\
0.244515742971011	1.4375\\
0.134409305177284	1.25\\
0.298454449269444	1.875\\
0.162159200452384	1.625\\
0.252762846540563	1.625\\
0.437222462224081	2.375\\
0.249601098674129	2.0625\\
0.388563993702505	1.625\\
0.338271776280762	2.0625\\
-0.0231285037286008	1.125\\
0.407469333971377	2.75\\
0.288250294534519	1.4375\\
};
\end{axis}
\end{tikzpicture}%
 \\
		(d) $CC$~\cite{Judd_2012} & (e) $EMD$~\cite{Judd_2012}  & (f) $NSS$~\cite{NSS}  \\
		\hspace*{-0.3in}
%
%
\begin{tikzpicture}

\begin{axis}[%
width=0.970772\figurewidth,
height=\figureheight,
at={(0\figurewidth,0\figureheight)},
scale only axis,
every outer x axis line/.append style={black},
every x tick label/.append style={font=\color{black}},
xmin=0.2,
xmax=1.2,
xlabel={Metric Score},
every outer y axis line/.append style={black},
every y tick label/.append style={font=\color{black}},
ymin=1,
ymax=5,
ylabel={MOS},
title={SROCC 0.5891 KROCC 0.4246 PLCC 0.6739},
axis x line*=bottom,
axis y line*=left,
title style={font=\scriptsize},xlabel style={font=\scriptsize},ylabel style={font=\scriptsize}
]
\addplot[only marks,mark=o,mark options={},mark size=1.0000pt,color=blue] plot table[row sep=crcr,]{%
0.381705546925715	1.3125\\
0.387783555680796	2.25\\
0.305746265326488	1.6875\\
0.392849874495774	2.1875\\
0.452355132053737	1.5\\
0.310315725090753	2.1875\\
0.429539614532745	1.9375\\
0.431668342581271	1.9375\\
0.326910517444355	1.9375\\
0.373874340946954	1.625\\
0.410104521692823	1.5\\
0.32557235249815	2.3125\\
0.999999999999751	4.9375\\
1.00000000000017	5\\
0.999999999999602	5\\
0.999999999999962	4.9375\\
1.00000000000039	5\\
1.00000000000025	5\\
0.999999999999955	5\\
1.00000000000045	5\\
1.00000000000028	5\\
1.00000000000019	5\\
1.00000000000058	5\\
1.00000000000017	5\\
0.409768316372741	1.4375\\
0.362030734307281	1.5625\\
0.374775826372379	2.375\\
0.363476700570483	1.3125\\
0.443363922034016	1.5\\
0.436819799997661	2.5625\\
0.429999695268481	2.0625\\
0.453569608132255	1.5\\
0.386370907985337	1.6875\\
0.362909859027438	1.25\\
0.402394905095376	2.875\\
0.34194133532032	1.4375\\
0.39684365962573	2.5625\\
0.435631362550129	2\\
0.364219896179797	3\\
0.409832247302241	2.6875\\
0.451017015906703	1.6875\\
0.497811811301119	3.5625\\
0.419764663587075	3.9375\\
0.532766121884402	2.75\\
0.48348613011539	3.3125\\
0.334125652895165	2.0625\\
0.451533822836957	4.9375\\
0.407754864457916	3.0625\\
0.481771012916573	2.6875\\
0.472114327097993	3.1875\\
0.363434172802412	2.8125\\
0.445278560892594	1.9375\\
0.464927106304679	2.25\\
0.553594253500933	3.9375\\
0.431750123605718	4\\
0.492283235804112	4.625\\
0.527190402339252	2.75\\
0.375053675514913	2.8125\\
0.495763737399577	5\\
0.392617185310948	4.4375\\
0.37090604057826	2.6875\\
0.49412788618247	2.625\\
0.484537407045507	2.75\\
0.3928015393164	1.9375\\
0.339264260195696	1.1875\\
0.543276880144195	3.9375\\
0.462445458638793	3.25\\
0.487726415473806	2.25\\
0.498693439278949	2.5\\
0.413140804241334	2.625\\
0.505616606656873	4.3125\\
0.410866491087988	2.875\\
0.527005893391949	2.125\\
0.675605323678753	4.125\\
0.415783088615156	3.25\\
0.598958257408765	4.125\\
0.532849079163234	2.25\\
0.514568707698161	4.6875\\
0.559770323212368	3.5\\
0.639469912687614	4.5625\\
0.540162842115451	4.0625\\
0.330091789453558	2.9375\\
0.426912415500473	1.8125\\
0.603943026158373	4.25\\
0.472494155844964	1.375\\
0.455203446109409	1.4375\\
0.329899343578824	1.6875\\
0.415053734976177	1.5\\
0.535329084362146	1.4375\\
0.400962809480698	2.6875\\
0.474881468228296	2.125\\
0.535885168480376	2.4375\\
0.41601949514185	2\\
0.347486230701879	1.625\\
0.473622625921785	2.3125\\
0.395028130489144	1.6875\\
0.426642590804472	1.625\\
0.451357402127853	1.875\\
0.353348967330769	1.875\\
0.383951022474203	1.8125\\
0.496266543307046	1.6875\\
0.377757829334972	2.375\\
0.45603754040506	3.3125\\
0.521363943439719	2.4375\\
0.416979569001861	2.3125\\
0.361692330592058	2.1875\\
0.375713004361318	1.375\\
0.357302834815551	1.8125\\
0.444832134797977	1.5625\\
0.480389401807956	2.25\\
0.365665725118549	2\\
0.406950336057327	1.875\\
0.510791223682209	2\\
0.396112743383527	2.875\\
0.478595888860312	3.5\\
0.544795626598466	2.9375\\
0.439227820639136	2.6875\\
0.363716538492676	2.5\\
0.397272402811635	1.125\\
0.383576379887145	2.25\\
0.364369300674237	2.375\\
0.37156155600155	1.75\\
0.371495827533443	1.75\\
0.294486504723959	1.3125\\
0.401302152686659	1.3125\\
0.427778854943032	2.75\\
0.392075148494263	1.0625\\
0.426590502500484	2.625\\
0.364807301587583	2.375\\
0.333142287351844	2.125\\
0.376390236745117	2.9375\\
0.296385648657807	2.25\\
0.414881320869004	2.8125\\
0.444908478973252	2\\
0.509737601083504	3.125\\
0.380799846100892	1.5\\
0.46661036869222	1.5\\
0.565606716826398	4\\
0.387834649148085	2.8125\\
0.531079291577782	3.5\\
0.590119232290551	4.4375\\
0.391470070044854	1.5625\\
0.496375004490282	4.9375\\
0.412948288369175	4\\
0.313029321399012	2.1875\\
0.355521788667742	2.125\\
0.290675520356017	3\\
0.315349377068797	1.1875\\
0.389350467613277	1.8125\\
0.41152954838562	3.5625\\
0.379419343141788	2.6875\\
0.445828944576328	3.625\\
0.297524893177053	2.25\\
0.350070981149791	2.1875\\
0.393614023323293	3.0625\\
0.275391729752052	1.6875\\
0.457890308976093	2.625\\
0.433839918939295	2.875\\
0.344614032115549	2.1875\\
0.415449707578722	2.125\\
0.496881427446547	1.875\\
0.435880540026034	3.8125\\
0.428828790780262	2.875\\
0.528426433608769	2.875\\
0.440603807628858	2.9375\\
0.354943722684111	2.125\\
0.481771188301188	2.25\\
0.433472206584143	3.4375\\
0.353489994754853	1.4375\\
0.362308892244665	1.9375\\
0.263584597711605	1.625\\
0.328202415049791	1.3125\\
0.443293221940771	1.3125\\
0.400742072721339	2.3125\\
0.352302044329742	1.4375\\
0.490084519600808	1.75\\
0.395428861214522	2\\
0.311998901096837	1.3125\\
0.387552551267995	2.875\\
0.326120949742367	1.25\\
0.468493006350365	1.8125\\
0.515443082809101	1.875\\
0.331248085222091	2.1875\\
0.431861258406389	2.5625\\
0.520066917608431	1.625\\
0.494955875699204	3.75\\
0.449519869039737	2.625\\
0.519157823703765	3.4375\\
0.421125477321593	3.0625\\
0.364429824800019	2.25\\
0.466491214439068	2.5\\
0.370686010524999	1.875\\
0.471598437739036	2.25\\
0.630787884647931	4.625\\
0.560542282461981	4.5625\\
0.598256351548003	4\\
0.542446634634765	2.0625\\
0.460417482972073	3.625\\
0.517335252536444	3.4375\\
0.612014426946315	4.5\\
0.581406935021389	4.5625\\
0.36851872001278	3.125\\
0.510805672979354	2.375\\
0.6518490141879	4.0625\\
0.278681273385258	1.75\\
0.481642103468285	2.9375\\
0.507852192501214	3.125\\
0.44061944491866	1.4375\\
0.349677718311399	1.6875\\
0.608897270494467	4.125\\
0.392586195323487	2.3125\\
0.594750773327102	3.3125\\
0.519044098926262	2.75\\
0.398270555627702	2.5625\\
0.487679163126258	4.625\\
0.500014212696637	4.375\\
0.321094072037044	1.625\\
0.452407870761561	3.0625\\
0.406257189220591	3\\
0.427065266717977	1.6875\\
0.419793974788999	1.625\\
0.532785606441527	4\\
0.387710680724926	2.3125\\
0.511627321320578	3.5625\\
0.464359669810815	2.5\\
0.386980482264824	2.625\\
0.474307021582779	4.5625\\
0.400811771211334	3.375\\
0.385577962758818	1.5625\\
0.335810283499577	1.75\\
0.350077539099646	2.0625\\
0.258949880422508	1.25\\
0.436040010480987	1.8125\\
0.409227995763184	1.8125\\
0.424272923991614	2.25\\
0.416860512715732	1.8125\\
0.373173424217964	2.5625\\
0.322721491829231	1.25\\
0.396390212583916	3.75\\
0.327142569983538	1.75\\
0.359134081815281	2.5625\\
0.443980521052069	3.375\\
0.47717518947517	3.1875\\
0.446420760607475	1.875\\
0.409546489367265	1.5\\
0.495788430958643	4.6875\\
0.44526466679216	3.5625\\
0.520567728059255	3.625\\
0.511029548685914	4.5\\
0.396462058587495	1.625\\
0.509613507056346	4.75\\
0.4045307223712	4.125\\
0.38480326273524	2.1875\\
0.41953061101636	1.6875\\
0.416076255342122	2.875\\
0.331648555912254	2\\
0.423767250391345	2.0625\\
0.471882804582864	2.1875\\
0.434153955816696	3.8125\\
0.502316642780568	3.5\\
0.479125465736897	2.6875\\
0.336222576260917	2.125\\
0.483180956526291	5\\
0.399822434647396	3\\
0.383315479741152	1.4375\\
0.344305685815736	1.25\\
0.332178361482638	1.875\\
0.355732822380918	1.625\\
0.442881311709099	1.625\\
0.426688949920983	2.375\\
0.41903874164506	2.0625\\
0.466648813672296	1.625\\
0.409034012209561	2.0625\\
0.342032834048251	1.125\\
0.408618435086458	2.75\\
0.34040723614105	1.4375\\
};
\end{axis}
\end{tikzpicture}%
 & \hspace*{-0.2in}
%
%
\begin{tikzpicture}

\begin{axis}[%
width=0.95092\figurewidth,
height=\figureheight,
at={(0\figurewidth,0\figureheight)},
scale only axis,
every outer x axis line/.append style={black},
every x tick label/.append style={font=\color{black}},
xmin=-0.4,
xmax=1.2,
xlabel={Metric Score},
every outer y axis line/.append style={black},
every y tick label/.append style={font=\color{black}},
ymin=1,
ymax=5,
ylabel={MOS},
title={SROCC 0.6526 KROCC 0.4843 PLCC 0.7533},
axis x line*=bottom,
axis y line*=left,
title style={font=\scriptsize},xlabel style={font=\scriptsize},ylabel style={font=\scriptsize}
]
\addplot[only marks,mark=o,mark options={},mark size=1.5000pt,color=blue] plot table[row sep=crcr,]{%
0.00606015166919935	1.3125\\
0.141880775096355	2.25\\
-0.0155874294820666	1.6875\\
0.0160143155185078	2.1875\\
-0.0466175582933712	1.5\\
-0.0160289395329212	2.1875\\
-0.00504902304377901	1.9375\\
-0.0290471937057927	1.9375\\
0.0634511881499858	1.9375\\
-0.0820051704952594	1.625\\
-0.0846571503798711	1.5\\
0.094860406178948	2.3125\\
1.01361236432147	4.9375\\
1.00681030003958	5\\
0.995240522438698	5\\
0.972612215239434	4.9375\\
1.02162884054077	5\\
0.986526966448972	5\\
0.984778905848888	5\\
0.993481022149703	5\\
0.994829260273411	5\\
1.00405807389827	5\\
0.999576097676323	5\\
1.02402760584632	5\\
0.187951808388558	1.4375\\
0.0980903711399859	1.5625\\
0.205226147832694	2.375\\
0.00523067443438156	1.3125\\
0.290389475173929	1.5\\
0.493466552934419	2.5625\\
0.313494088130093	2.0625\\
0.324984037374564	1.5\\
0.218565404502529	1.6875\\
0.0793221680664024	1.25\\
0.332337476725905	2.875\\
0.226026714842792	1.4375\\
0.187898487387481	2.5625\\
0.229636565127978	2\\
0.26935981139453	3\\
0.218621695663486	2.6875\\
0.333358474921867	1.6875\\
0.582784558332313	3.5625\\
0.420743554115131	3.9375\\
0.441711216595541	2.75\\
0.415435512058109	3.3125\\
0.141734296212685	2.0625\\
0.760030126488556	4.9375\\
0.39618565867598	3.0625\\
0.343132173214072	2.6875\\
0.337183959981067	3.1875\\
0.0978685167309732	2.8125\\
0.471178430358557	1.9375\\
0.313593614516618	2.25\\
0.608339858878376	3.9375\\
0.305529013533423	4\\
0.555676083551078	4.625\\
0.507428995165214	2.75\\
0.193926498318483	2.8125\\
0.802751045602151	5\\
0.640819931159375	4.4375\\
0.161593137908642	2.6875\\
0.333168598920755	2.625\\
0.384820391988853	2.75\\
0.0991250981262089	1.9375\\
0.085865798629591	1.1875\\
0.524649889554519	3.9375\\
0.387829203073066	3.25\\
0.286397553463672	2.25\\
0.290389385615136	2.5\\
0.222717570850661	2.625\\
0.690411274028198	4.3125\\
0.33731844356765	2.875\\
-0.103109291184963	2.125\\
0.435841719235983	4.125\\
-0.11989420610348	3.25\\
0.212343972193083	4.125\\
-0.108653641225082	2.25\\
0.370358690919254	4.6875\\
0.10119451295989	3.5\\
0.295767668312874	4.5625\\
0.2924760060773	4.0625\\
-0.14212389077196	2.9375\\
-0.216124165003583	1.8125\\
0.486242717750334	4.25\\
-0.00101550802574731	1.375\\
0.0780519966722856	1.4375\\
-0.0571687328726051	1.6875\\
-0.190512571844346	1.5\\
0.0929660767837028	1.4375\\
0.191447540654373	2.6875\\
0.087042843780333	2.125\\
0.0822246977664858	2.4375\\
0.161415200215843	2\\
-0.0879604844439014	1.625\\
0.263966227992824	2.3125\\
0.158304409355127	1.6875\\
-0.0941292164622346	1.625\\
0.124465293961786	1.875\\
-0.016629395779747	1.875\\
-0.156693120058381	1.8125\\
0.00672405998336814	1.6875\\
0.230258835773921	2.375\\
0.173478940650191	3.3125\\
0.0941138531265768	2.4375\\
0.150971407948881	2.3125\\
-0.0362437370946625	2.1875\\
-0.0511281338705084	1.375\\
0.0607228073571402	1.8125\\
-0.111322286512691	1.5625\\
0.184849372211415	2.25\\
-0.030698486315573	2\\
-0.132359511800156	1.875\\
-0.0235049394101964	2\\
0.288608374272897	2.875\\
0.168568077170788	3.5\\
0.104288179432259	2.9375\\
0.176078998525917	2.6875\\
-0.0535021323419474	2.5\\
-0.0636713161391309	1.125\\
0.118366892760352	2.25\\
0.0402143432586021	2.375\\
0.305664179935675	1.75\\
0.379692502894988	1.75\\
0.0107592330811858	1.3125\\
0.1665318318811	1.3125\\
0.467788817971594	2.75\\
0.331011378418701	1.0625\\
0.167037358592993	2.625\\
0.175176446682669	2.375\\
0.0693331212079386	2.125\\
0.425824367345581	2.9375\\
0.183942875312651	2.25\\
0.175560192617816	2.8125\\
0.215527088707599	2\\
0.342991272445686	3.125\\
0.01567078020402	1.5\\
0.327420286707656	1.5\\
0.503164351599039	4\\
0.210804419700974	2.8125\\
0.410102953214806	3.5\\
0.451057426035368	4.4375\\
0.183312974382796	1.5625\\
0.795056342571422	4.9375\\
0.420966325065664	4\\
0.170092025714817	2.1875\\
0.10944819614022	2.125\\
0.191520874309497	3\\
0.138004126975392	1.1875\\
0.124838670875703	1.8125\\
0.570927859148162	3.5625\\
0.191277739459556	2.6875\\
0.326158915523674	3.625\\
0.311422868358086	2.25\\
0.0520635423426456	2.1875\\
0.436103482841508	3.0625\\
-0.0240071530846156	1.6875\\
0.0652924614816264	2.625\\
0.0681694067165633	2.875\\
0.00588684057123404	2.1875\\
-0.0553396046397972	2.125\\
0.0741914945782933	1.875\\
0.274760769560593	3.8125\\
0.0657746286059974	2.875\\
0.210257965736842	2.875\\
0.135935609370351	2.9375\\
-0.108021976801673	2.125\\
0.185933976586317	2.25\\
0.243499187226832	3.4375\\
-0.0228046031221739	1.4375\\
0.019230989563843	1.9375\\
-0.0947661936827436	1.625\\
-0.0332173741630467	1.3125\\
0.25347102035472	1.3125\\
0.402286478927024	2.3125\\
0.0973790957872162	1.4375\\
0.357011721974119	1.75\\
0.229568897072063	2\\
0.0208177933231411	1.3125\\
0.283833418226846	2.875\\
0.184639480265254	1.25\\
-0.0493794651826784	1.8125\\
0.249530857976496	1.875\\
-0.0471326453340074	2.1875\\
-0.110905883172785	2.5625\\
-0.00956142843975342	1.625\\
0.565037277214233	3.75\\
0.0290862889389936	2.625\\
0.00513151661492627	3.4375\\
0.219667942144205	3.0625\\
-0.00495531771640337	2.25\\
0.158370019458953	2.5\\
0.0725174252491588	1.875\\
-0.134567600866627	2.25\\
0.52298471035678	4.625\\
0.299037945991435	4.5625\\
0.16707433263673	4\\
0.140888145128994	2.0625\\
0.602717453642324	3.625\\
0.388425929217222	3.4375\\
0.447444119106347	4.5\\
0.512672977432708	4.5625\\
0.00743724228869491	3.125\\
0.176884403845151	2.375\\
0.549603107184005	4.0625\\
0.0259033891200295	1.75\\
0.33494696437262	2.9375\\
0.38972844401803	3.125\\
0.410162654249563	1.4375\\
0.165465062118579	1.6875\\
0.520759244461879	4.125\\
0.364152784554233	2.3125\\
0.493769782953464	3.3125\\
0.297401696575974	2.75\\
0.231828715436817	2.5625\\
0.714282706783572	4.625\\
0.431685613177078	4.375\\
-0.00112648971130267	1.625\\
0.284725058880521	3.0625\\
0.301758065128414	3\\
0.384590486606207	1.6875\\
0.25733376220692	1.625\\
0.526717714257102	4\\
0.322843520592893	2.3125\\
0.426547859539272	3.5625\\
0.261014434153786	2.5\\
0.174396415433424	2.625\\
0.6860796720421	4.5625\\
0.33404570490774	3.375\\
0.198641074048764	1.5625\\
0.109877121368985	1.75\\
0.255382359138867	2.0625\\
-0.186624956757704	1.25\\
0.323052038138069	1.8125\\
0.495216727955521	1.8125\\
0.363817553960858	2.25\\
0.226798745901909	1.8125\\
0.266536655371132	2.5625\\
-0.0454501565906551	1.25\\
0.481431750154335	3.75\\
0.193569697876836	1.75\\
0.11930174432167	2.5625\\
0.30487613537156	3.375\\
0.347924377010533	3.1875\\
0.386840485061934	1.875\\
0.200407095479206	1.5\\
0.479420545074448	4.6875\\
0.369978660376098	3.5625\\
0.41760031976981	3.625\\
0.47970843051922	4.5\\
0.167677795655507	1.625\\
0.717325166685665	4.75\\
0.51898744375532	4.125\\
0.105341471488356	2.1875\\
0.114651583810506	1.6875\\
0.193439136079334	2.875\\
-0.00989567781038743	2\\
0.26440958824889	2.0625\\
0.561350805599968	2.1875\\
0.376916758269781	3.8125\\
0.228154284153436	3.5\\
0.270568746728498	2.6875\\
0.0210093521841109	2.125\\
0.625816233781232	5\\
0.271695848239175	3\\
0.115068175794871	1.4375\\
0.0500644133057158	1.25\\
0.139340976253497	1.875\\
0.0299829201089126	1.625\\
0.303636348540585	1.625\\
0.476523612770787	2.375\\
0.335903738422515	2.0625\\
0.383305758054938	1.625\\
0.229272044396009	2.0625\\
0.0014452177223567	1.125\\
0.448890799194521	2.75\\
0.218681520793693	1.4375\\
};
\end{axis}
\end{tikzpicture}%
 & \hspace*{-0.2in} 
%
%
\begin{tikzpicture}

\begin{axis}[%
width=0.970772\figurewidth,
height=\figureheight,
at={(0\figurewidth,0\figureheight)},
scale only axis,
every outer x axis line/.append style={black},
every x tick label/.append style={font=\color{black}},
xmin=0.2,
xmax=1.6,
xlabel={Metric Score},
every outer y axis line/.append style={black},
every y tick label/.append style={font=\color{black}},
ymin=1,
ymax=5,
ylabel={MOS},
title={SROCC 0.3126 KROCC 0.2113 PLCC 0.2958},
axis x line*=bottom,
axis y line*=left,
title style={font=\scriptsize},xlabel style={font=\scriptsize},ylabel style={font=\scriptsize}
]
\addplot[only marks,mark=o,mark options={},mark size=1.0000pt,color=blue] plot table[row sep=crcr,]{%
0.782326933720877	1.3125\\
0.733350216641925	2.25\\
0.458144407273194	1.6875\\
0.70387560710658	2.1875\\
1.16929021410271	1.5\\
0.484466413860036	2.1875\\
0.754663307853934	1.9375\\
1.03302104316154	1.9375\\
0.782915715277029	1.9375\\
0.75524463005125	1.625\\
1.0823522843932	1.5\\
0.614313021233164	2.3125\\
1	4.9375\\
1	5\\
1	5\\
1	4.9375\\
1	5\\
1	5\\
1	5\\
1	5\\
1	5\\
1	5\\
1	5\\
1	5\\
0.760738132706546	1.4375\\
0.636851692662146	1.5625\\
0.582012998853306	2.375\\
0.590583870264729	1.3125\\
1.09583227141099	1.5\\
0.745246880422734	2.5625\\
0.734305596666581	2.0625\\
0.990967871954661	1.5\\
0.979976358286893	1.6875\\
0.698696388543855	1.25\\
0.925123883986018	2.875\\
0.614274125683711	1.4375\\
0.644398360502065	2.5625\\
0.744595742178512	2\\
0.519207048451346	3\\
0.633999613730937	2.6875\\
0.923361940182	1.6875\\
0.851532664898711	3.5625\\
0.631188135700489	3.9375\\
1.17073972529594	2.75\\
1.14866914189099	3.3125\\
0.656089222417674	2.0625\\
0.72758968130201	4.9375\\
0.805102655529759	3.0625\\
0.853331645331436	2.6875\\
0.830991342498329	3.1875\\
0.477301578023937	2.8125\\
0.509201990836455	1.9375\\
0.860234767395531	2.25\\
0.840383969366559	3.9375\\
0.649594644666926	4\\
0.74566460753402	4.625\\
0.900839648088446	2.75\\
0.75063685726274	2.8125\\
0.826232591511142	5\\
0.541052216515835	4.4375\\
0.580503245315086	2.6875\\
0.907663449169768	2.625\\
0.74652962355759	2.75\\
0.628090904490005	1.9375\\
0.521535236238742	1.1875\\
0.851296092377569	3.9375\\
0.731065167266428	3.25\\
1.12517950097289	2.25\\
1.30637723783822	2.5\\
0.901916069447076	2.625\\
0.929825372261255	4.3125\\
0.7528031651527	2.875\\
0.828990003657779	2.125\\
1.04762502701001	4.125\\
0.542674363729277	3.25\\
0.801793467400744	4.125\\
1.14642067988427	2.25\\
0.822789515631432	4.6875\\
0.707182334186394	3.5\\
0.941893591291827	4.5625\\
1.22001228625335	4.0625\\
0.731105050744438	2.9375\\
1.30182606145739	1.8125\\
1.04346149556643	4.25\\
0.839439267304957	1.375\\
0.840176780230016	1.4375\\
0.456659168179454	1.6875\\
0.632664391469497	1.5\\
1.245301613244	1.4375\\
0.646432349311121	2.6875\\
0.725059950191677	2.125\\
1.08998392689725	2.4375\\
1.0060847557146	2\\
0.701480487110614	1.625\\
1.04912292900505	2.3125\\
0.75027833186192	1.6875\\
0.779940384968042	1.625\\
0.812031213419143	1.875\\
0.492068769481569	1.875\\
0.564404343689355	1.8125\\
1.17183019921346	1.6875\\
0.605655003945163	2.375\\
0.6703821309071	3.3125\\
1.24421633574512	2.4375\\
1.03235212567798	2.3125\\
0.686921750810812	2.1875\\
0.757954397828847	1.375\\
0.643069798706135	1.8125\\
0.815955091634059	1.5625\\
0.879800144785887	2.25\\
0.515700245248519	2\\
0.618779971293846	1.875\\
1.18775729932852	2\\
0.655315394941989	2.875\\
0.690296039483531	3.5\\
1.27813446534034	2.9375\\
1.13274430745378	2.6875\\
0.691876744606172	2.5\\
0.855699867012428	1.125\\
0.719030720247624	2.25\\
0.495357546263888	2.375\\
0.489653144694687	1.75\\
0.441733360574288	1.75\\
0.273232712971622	1.3125\\
0.62347558147598	1.3125\\
0.598077397543019	2.75\\
0.473678365816306	1.0625\\
0.618155552109676	2.625\\
0.672569310538125	2.375\\
0.502847283547023	2.125\\
0.63568970464594	2.9375\\
0.365112976590646	2.25\\
0.563978774059774	2.8125\\
0.680020398925194	2\\
0.567304846018974	3.125\\
0.504999665316887	1.5\\
0.919777556843811	1.5\\
0.840664024424833	4\\
0.539446966453023	2.8125\\
1.06500777536131	3.5\\
1.13958578929336	4.4375\\
0.587417158893762	1.5625\\
0.599763077754122	4.9375\\
0.721263843248779	4\\
0.291994293049609	2.1875\\
0.318552756957254	2.125\\
0.227366767787677	3\\
0.286954065954417	1.1875\\
0.404857562961867	1.8125\\
0.359903185546919	3.5625\\
0.33725852327666	2.6875\\
0.501410974191058	3.625\\
0.387329512788992	2.25\\
0.382931771030079	2.1875\\
0.480016048760645	3.0625\\
0.24947015912353	1.6875\\
0.795536441765531	2.625\\
0.82291914697468	2.875\\
0.487599460620328	2.1875\\
0.635078090977866	2.125\\
1.07252772823716	1.875\\
0.739465730516943	3.8125\\
0.666875857574624	2.875\\
1.2188252464151	2.875\\
1.13621023246578	2.9375\\
0.707604589262684	2.125\\
1.13735091793652	2.25\\
0.892639274552181	3.4375\\
0.483995159640267	1.4375\\
0.479456292954087	1.9375\\
0.327077163520383	1.625\\
0.390549257213132	1.3125\\
0.801688067157781	1.3125\\
0.501672746795076	2.3125\\
0.451514850727498	1.4375\\
0.683624090475257	1.75\\
0.781734831656975	2\\
0.511149485108637	1.3125\\
0.663826481435677	2.875\\
0.493334072252033	1.25\\
0.793337292928174	1.8125\\
0.867023735648072	1.875\\
0.476147597715249	2.1875\\
0.669539841595504	2.5625\\
1.09203935214422	1.625\\
0.657354188731791	3.75\\
0.547319168288377	2.625\\
1.05344636121146	3.4375\\
0.973521366302783	3.0625\\
0.708739284827704	2.25\\
1.17958312437961	2.5\\
0.649395354145942	1.875\\
0.703427423012265	2.25\\
0.889277829399416	4.625\\
0.60894042611169	4.5625\\
0.824218702977617	4\\
0.574907508748941	2.0625\\
0.332388473574146	3.625\\
0.378201536512173	3.4375\\
0.714539306590621	4.5\\
0.98084819156503	4.5625\\
0.81918275233382	3.125\\
1.35059116398461	2.375\\
1.03578104742151	4.0625\\
0.421081564427694	1.75\\
1.0124022876507	2.9375\\
0.851045668410014	3.125\\
0.772159643816028	1.4375\\
0.668272482122955	1.6875\\
1.17718721926597	4.125\\
0.674386694449931	2.3125\\
1.17276644177647	3.3125\\
1.57142916023401	2.75\\
0.960624861777003	2.5625\\
0.714486840046429	4.625\\
0.922723316093435	4.375\\
0.542730362993949	1.625\\
0.884729111362708	3.0625\\
0.644451607835535	3\\
0.75011210829569	1.6875\\
0.922317290489003	1.625\\
0.967076074931011	4\\
0.650048904256244	2.3125\\
1.08562044411258	3.5625\\
1.27011336703201	2.5\\
0.822992428030791	2.625\\
0.959291176592643	4.5625\\
0.777857640640587	3.375\\
0.616765198077479	1.5625\\
0.526396982885918	1.75\\
0.4622116406124	2.0625\\
0.338943787078632	1.25\\
0.96422090055895	1.8125\\
0.571057051523333	1.8125\\
0.608276002365271	2.25\\
0.711857722909901	1.8125\\
0.828023009190709	2.5625\\
0.513147807108465	1.25\\
0.827272099573671	3.75\\
0.574336664814431	1.75\\
0.587714375178751	2.5625\\
0.773316062767752	3.375\\
0.596150383112651	3.1875\\
0.702297945904784	1.875\\
0.677489362986281	1.5\\
0.805727622933647	4.6875\\
0.619486064583133	3.5625\\
1.09931238460561	3.625\\
1.10790409595356	4.5\\
0.715961537554945	1.625\\
0.855007138084967	4.75\\
0.714971227668179	4.125\\
0.557735121097021	2.1875\\
0.677456042412645	1.6875\\
0.585965851379258	2.875\\
0.41864585043863	2\\
0.806900562185436	2.0625\\
0.799977575020287	2.1875\\
0.57180864387924	3.8125\\
1.07047153924748	3.5\\
1.0404551120475	2.6875\\
0.613926614929984	2.125\\
0.899458896895984	5\\
0.712674991911568	3\\
0.690275604761118	1.4375\\
0.55259558804299	1.25\\
0.474906031126996	1.875\\
0.540706856355007	1.625\\
1.01778760595719	1.625\\
0.701922761016335	2.375\\
0.660431330244124	2.0625\\
1.0256605702369	1.625\\
0.983263909103136	2.0625\\
0.585782081093038	1.125\\
0.866343139309419	2.75\\
0.594739210974292	1.4375\\
};
\end{axis}
\end{tikzpicture}%
 \\
		(j) $SIM$~\cite{Judd_2012}  & (k)  $sNSS$~\cite{MilindSamTPAMI} & (l)  $WF_{\beta}~\cite{margolin2014evaluate}$ \\
		\hspace*{-0.3in} 
%
%
\definecolor{mycolor1}{rgb}{0.00000,0.44700,0.74100}%
\begin{tikzpicture}

\begin{axis}[%
width=0.95092\figurewidth,
height=\figureheight,
at={(0\figurewidth,0\figureheight)},
scale only axis,
xmin=0.3,
xmax=1,
xlabel={Metric Score},
ymin=1,
ymax=5,
ylabel={Mean Opinion Score},
title style={font=\bfseries},
title={SROCC 0.5063 KROCC 0.3599 PLCC 0.4803},
axis x line*=bottom,
axis y line*=left,
title style={font=\footnotesize},xlabel style={font=\footnotesize},ylabel style={font=\footnotesize}
]
\addplot[only marks,mark=o,mark options={},mark size=1.5000pt,color=blue] plot table[row sep=crcr,]{%
0.402146641047416	1.3125\\
0.394214657681388	2.25\\
0.403347201733556	1.6875\\
0.416984561410571	2.1875\\
0.395439232962002	1.5\\
0.385127359333992	2.1875\\
0.421080689431916	1.9375\\
0.391206714214748	1.9375\\
0.366787184771218	1.9375\\
0.368434319999691	1.625\\
0.371692596737859	1.5\\
0.377000013692118	2.3125\\
1	4.9375\\
1	5\\
1	5\\
1	4.9375\\
1	5\\
1	5\\
1	5\\
1	5\\
1	5\\
1	5\\
1	5\\
1	5\\
0.738276654777043	1.4375\\
0.739998887733132	1.5625\\
0.742818500884054	2.375\\
0.664446809842486	1.3125\\
0.677674610995793	1.5\\
0.801951921113939	2.5625\\
0.695858656139427	2.0625\\
0.802260237334391	1.5\\
0.73294250494434	1.6875\\
0.68860990939525	1.25\\
0.703827229020197	2.875\\
0.731385976231772	1.4375\\
0.815535797325308	2.5625\\
0.817776851452156	2\\
0.766377944978338	3\\
0.751092278079441	2.6875\\
0.802989450763831	1.6875\\
0.889543265048162	3.5625\\
0.774360786094179	3.9375\\
0.851743134443231	2.75\\
0.832894292228925	3.3125\\
0.771621003720913	2.0625\\
0.922681378789363	4.9375\\
0.753999238542276	3.0625\\
0.825483834633677	2.6875\\
0.830569023618422	3.1875\\
0.824925300919116	2.8125\\
0.879174236025111	1.9375\\
0.842150244926838	2.25\\
0.922745939846008	3.9375\\
0.795606670242322	4\\
0.911765004515312	4.625\\
0.950169536162812	2.75\\
0.715084224907457	2.8125\\
0.930100220644696	5\\
0.902425686736141	4.4375\\
0.836817641384725	2.6875\\
0.827025479191622	2.625\\
0.83940953848191	2.75\\
0.728422145164443	1.9375\\
0.885191170990957	1.1875\\
0.91936319059057	3.9375\\
0.824127196874652	3.25\\
0.874561957625552	2.25\\
0.815608787718733	2.5\\
0.698262367727336	2.625\\
0.917304273307509	4.3125\\
0.792744842792517	2.875\\
0.916341706155187	2.125\\
0.944692450633037	4.125\\
0.896079913267954	3.25\\
0.924882911948192	4.125\\
0.896709769539292	2.25\\
0.922781674373916	4.6875\\
0.924343355510484	3.5\\
0.952230723118231	4.5625\\
0.941975262939259	4.0625\\
0.897543000464058	2.9375\\
0.895748182804532	1.8125\\
0.945602672505461	4.25\\
0.832501426535542	1.375\\
0.809162283383251	1.4375\\
0.788137094597969	1.6875\\
0.812076723889884	1.5\\
0.824012070179075	1.4375\\
0.829785207990142	2.6875\\
0.833174905751641	2.125\\
0.872848165275385	2.4375\\
0.823532307940484	2\\
0.791135977654493	1.625\\
0.862655406740553	2.3125\\
0.816344033238075	1.6875\\
0.787369731953192	1.625\\
0.801879336410498	1.875\\
0.799278480325748	1.875\\
0.80022213781606	1.8125\\
0.771319553508754	1.6875\\
0.806647951961234	2.375\\
0.81809108089952	3.3125\\
0.815299133553098	2.4375\\
0.776493104030486	2.3125\\
0.803636391092054	2.1875\\
0.829812377951288	1.375\\
0.780698204894874	1.8125\\
0.80700740457498	1.5625\\
0.82135737001528	2.25\\
0.813578250856261	2\\
0.806720198589648	1.875\\
0.79440244707695	2\\
0.808273427851111	2.875\\
0.840770729871744	3.5\\
0.837271123926312	2.9375\\
0.783026696487052	2.6875\\
0.830366742074214	2.5\\
0.823714383868615	1.125\\
0.788440900065998	2.25\\
0.878297787845636	2.375\\
0.893189484672849	1.75\\
0.890422277755639	1.75\\
0.876633557291084	1.3125\\
0.886080894248951	1.3125\\
0.909642941520788	2.75\\
0.875796618810441	1.0625\\
0.913024345175517	2.625\\
0.92151540317618	2.375\\
0.854749443880637	2.125\\
0.918071748643983	2.9375\\
0.904461547658946	2.25\\
0.890373239258778	2.8125\\
0.883859200550638	2\\
0.924162424911216	3.125\\
0.830186672057749	1.5\\
0.85704024010984	1.5\\
0.940730177775074	4\\
0.853885583180943	2.8125\\
0.899292989378081	3.5\\
0.945333260534544	4.4375\\
0.917256382942253	1.5625\\
0.954884431595367	4.9375\\
0.849574326944698	4\\
0.893696337988446	2.1875\\
0.92805185800108	2.125\\
0.903695740159164	3\\
0.886188638752832	1.1875\\
0.918463724073131	1.8125\\
0.941320896230101	3.5625\\
0.89800285008498	2.6875\\
0.934178453229129	3.625\\
0.92991090991145	2.25\\
0.943454972156426	2.1875\\
0.937435411474788	3.0625\\
0.924040733304991	1.6875\\
0.819917711416938	2.625\\
0.791368505642452	2.875\\
0.794787232777731	2.1875\\
0.808415171431254	2.125\\
0.827065856616322	1.875\\
0.840619520191498	3.8125\\
0.790142904867681	2.875\\
0.818582315149931	2.875\\
0.804278609639421	2.9375\\
0.799083298356577	2.125\\
0.835152092337674	2.25\\
0.764086810674129	3.4375\\
0.874708443350946	1.4375\\
0.882619355549026	1.9375\\
0.801949836318083	1.625\\
0.849260610555095	1.3125\\
0.87571700214692	1.3125\\
0.925113715753668	2.3125\\
0.840551679807299	1.4375\\
0.933544581028189	1.75\\
0.90326764705041	2\\
0.832336141665955	1.3125\\
0.907416728886602	2.875\\
0.885881267513877	1.25\\
0.856256007522359	1.8125\\
0.888277547858581	1.875\\
0.780256724015956	2.1875\\
0.822374851817419	2.5625\\
0.844812844776556	1.625\\
0.927690095208151	3.75\\
0.881655785440923	2.625\\
0.843748167703856	3.4375\\
0.849622344864517	3.0625\\
0.831221406954277	2.25\\
0.785720444313858	2.5\\
0.831796427459965	1.875\\
0.908830648707874	2.25\\
0.955983907855195	4.625\\
0.939733408701538	4.5625\\
0.936185491838012	4\\
0.947761675828798	2.0625\\
0.959624795344896	3.625\\
0.92948574678598	3.4375\\
0.96295818107986	4.5\\
0.96863673677385	4.5625\\
0.924330215661967	3.125\\
0.941726349582732	2.375\\
0.963923971213829	4.0625\\
0.794743947907962	1.75\\
0.819804918326324	2.9375\\
0.830427854765847	3.125\\
0.773384707210862	1.4375\\
0.835384011790689	1.6875\\
0.933927049386735	4.125\\
0.725720943537968	2.3125\\
0.922185805909683	3.3125\\
0.790112094965668	2.75\\
0.76124043325202	2.5625\\
0.962317607830868	4.625\\
0.914780926552011	4.375\\
0.696545172088816	1.625\\
0.740707729594643	3.0625\\
0.708105770821897	3\\
0.646509891144822	1.6875\\
0.712535387296663	1.625\\
0.87339457336651	4\\
0.631215208912552	2.3125\\
0.827875512444135	3.5625\\
0.71110249019841	2.5\\
0.626587198756088	2.625\\
0.881687226416317	4.5625\\
0.730395191085788	3.375\\
0.810499103992574	1.5625\\
0.780667316108635	1.75\\
0.803112029099236	2.0625\\
0.742798362164697	1.25\\
0.7611644813305	1.8125\\
0.864812291460509	1.8125\\
0.795737426384813	2.25\\
0.867093215899583	1.8125\\
0.809995119887225	2.5625\\
0.726632569629162	1.25\\
0.771235273422648	3.75\\
0.714889200706624	1.75\\
0.779729039201879	2.5625\\
0.800567278587956	3.375\\
0.898497476354633	3.1875\\
0.759557369568431	1.875\\
0.855932447204916	1.5\\
0.885478512206288	4.6875\\
0.839724781888114	3.5625\\
0.796258738041414	3.625\\
0.885455207569335	4.5\\
0.811381328827383	1.625\\
0.926783434392456	4.75\\
0.794457714709143	4.125\\
0.849136181739135	2.1875\\
0.846571634909818	1.6875\\
0.843865006425315	2.875\\
0.817610541088395	2\\
0.867211339306074	2.0625\\
0.900581066413645	2.1875\\
0.863683853441975	3.8125\\
0.880998916976687	3.5\\
0.891923337435877	2.6875\\
0.741305389856448	2.125\\
0.909037798532252	5\\
0.801848932198432	3\\
0.737831105728853	1.4375\\
0.750963144696944	1.25\\
0.720927569627724	1.875\\
0.712110055133747	1.625\\
0.719955919526993	1.625\\
0.805047274600099	2.375\\
0.732951719162251	2.0625\\
0.8013354413436	1.625\\
0.771416311262194	2.0625\\
0.716239173039696	1.125\\
0.806510682240356	2.75\\
0.736166514560984	1.4375\\
};
\end{axis}
\end{tikzpicture}%
 & \hspace*{-0.2in}
%
%
\begin{tikzpicture}

\begin{axis}[%
width=0.95092\figurewidth,
height=\figureheight,
at={(0\figurewidth,0\figureheight)},
scale only axis,
every outer x axis line/.append style={black},
every x tick label/.append style={font=\color{black}},
xmin=-0.2,
xmax=1.2,
xlabel={Metric Score},
every outer y axis line/.append style={black},
every y tick label/.append style={font=\color{black}},
ymin=1,
ymax=5,
ylabel={MOS},
title={SROCC 0.7858 KROCC 0.6178 PLCC 0.8687},
axis x line*=bottom,
axis y line*=left,
title style={font=\scriptsize},xlabel style={font=\scriptsize},ylabel style={font=\scriptsize}
]
\addplot[only marks,mark=o,mark options={},mark size=1.5000pt,color=blue] plot table[row sep=crcr,]{%
0.261072573921285	1.3125\\
0.324608397354791	2.25\\
0.296837423987065	1.6875\\
0.386152758975591	2.1875\\
0.214358618713225	1.5\\
0.301234200653243	2.1875\\
0.331121896702441	1.9375\\
0.270840293792554	1.9375\\
0.238539256597915	1.9375\\
0.169596840632642	1.625\\
0.103001293549131	1.5\\
0.306436247263399	2.3125\\
1	4.9375\\
1	5\\
1	5\\
1	4.9375\\
1	5\\
1	5\\
1	5\\
1	5\\
1	5\\
1	5\\
1	5\\
1	5\\
0.218310081476094	1.4375\\
0.118440380305644	1.5625\\
0.341577070478325	2.375\\
0.154586017643543	1.3125\\
0.186582832409152	1.5\\
0.337778461049875	2.5625\\
0.219267665725198	2.0625\\
0.305298324411302	1.5\\
0.247428426832969	1.6875\\
0.0384503767240381	1.25\\
0.338762030486136	2.875\\
0.192216068535326	1.4375\\
0.210495478982184	2.5625\\
0.224222936235352	2\\
0.469577284146743	3\\
0.312403580007153	2.6875\\
0.167863480415598	1.6875\\
0.492039263707083	3.5625\\
0.157325795811539	3.9375\\
0.408575306881036	2.75\\
0.452128969775438	3.3125\\
0.130074359886633	2.0625\\
0.804075954744801	4.9375\\
0.417708647651739	3.0625\\
0.295025547734579	2.6875\\
0.381656180931132	3.1875\\
0.416441130274377	2.8125\\
0.296287070620781	1.9375\\
0.213956429488385	2.25\\
0.570484842351769	3.9375\\
0.186598000686836	4\\
0.677272367006064	4.625\\
0.449678887229943	2.75\\
0.158235222759077	2.8125\\
0.873036194501707	5\\
0.725354667678805	4.4375\\
0.148114431268234	2.6875\\
0.383455515392587	2.625\\
0.552518453844457	2.75\\
0.250319579146975	1.9375\\
0.101700687329052	1.1875\\
0.646591803749415	3.9375\\
0.249517665812797	3.25\\
0.301025392868817	2.25\\
0.33340395662489	2.5\\
0.190533529561573	2.625\\
0.788546285125454	4.3125\\
0.404362889600757	2.875\\
0.201964441219699	2.125\\
0.65483899348752	4.125\\
0.399079791754244	3.25\\
0.829230230854556	4.125\\
0.273004911374036	2.25\\
0.76410700994195	4.6875\\
0.610707835307885	3.5\\
0.757083313089005	4.5625\\
0.553367965599319	4.0625\\
0.203046351258471	2.9375\\
0.0352387747858376	1.8125\\
0.740365582372154	4.25\\
0.195344723856118	1.375\\
0.153767508900025	1.4375\\
0.137203194274165	1.6875\\
0.246029898593988	1.5\\
0.245800824473246	1.4375\\
0.369969675831573	2.6875\\
0.307302995825128	2.125\\
0.352919760292806	2.4375\\
0.262242106201373	2\\
0.155578882586397	1.625\\
0.324171587132416	2.3125\\
0.295432707264419	1.6875\\
0.172612539368012	1.625\\
0.207567570849191	1.875\\
0.281678143182337	1.875\\
0.187375572434061	1.8125\\
0.18400327056931	1.6875\\
0.296744623610057	2.375\\
0.300402523459173	3.3125\\
0.291945272044292	2.4375\\
0.27488298983046	2.3125\\
0.163839106431385	2.1875\\
0.0287213619138154	1.375\\
0.208739748129178	1.8125\\
0.185253625589421	1.5625\\
0.281249309485967	2.25\\
0.314085614026454	2\\
0.29089090188015	1.875\\
0.202894790446117	2\\
0.405870156190827	2.875\\
0.367096736349703	3.5\\
0.377751777516227	2.9375\\
0.339711058001444	2.6875\\
0.192721239809891	2.5\\
0.037133379332378	1.125\\
0.304766326739255	2.25\\
0.0315931924051937	2.375\\
0.294815626827026	1.75\\
0.433819778529099	1.75\\
0.0191416979322467	1.3125\\
0.146430232880414	1.3125\\
0.371020586425472	2.75\\
0.177173161606604	1.0625\\
0.141136754660566	2.625\\
0.191078946638524	2.375\\
0.0738086123141628	2.125\\
0.403401400048472	2.9375\\
0.138310726615663	2.25\\
0.110321564903036	2.8125\\
0.282023546361266	2\\
0.641168919224017	3.125\\
0.195185056713908	1.5\\
0.222562422427096	1.5\\
0.547327549823332	4\\
0.103969241626527	2.8125\\
0.479449514294285	3.5\\
0.561277834588227	4.4375\\
0.0979537942616853	1.5625\\
0.918599343920321	4.9375\\
0.482319342328483	4\\
0.0500134056960804	2.1875\\
0.0859557930591725	2.125\\
0.0523412735493608	3\\
-0.0381726356417086	1.1875\\
0.0110355789530837	1.8125\\
0.615098459243778	3.5625\\
-0.0292393632357295	2.6875\\
0.311653987874439	3.625\\
0.268964968635695	2.25\\
0.120278546837186	2.1875\\
0.522788617590926	3.0625\\
0.016279787650365	1.6875\\
0.20854506442028	2.625\\
0.184270534591106	2.875\\
0.29408470957776	2.1875\\
0.163808280316782	2.125\\
0.185064219874749	1.875\\
0.425193094960757	3.8125\\
0.240256053588531	2.875\\
0.342910004038541	2.875\\
0.320860688551325	2.9375\\
0.145531689454287	2.125\\
0.251968215680163	2.25\\
0.404500438077391	3.4375\\
0.0173875858886772	1.4375\\
0.0404030057448765	1.9375\\
-0.0394897923557584	1.625\\
0.0363817180465122	1.3125\\
0.142999102478088	1.3125\\
0.239946485059749	2.3125\\
0.0377852994218796	1.4375\\
0.245894826100367	1.75\\
0.207041627736769	2\\
0.0631954640899496	1.3125\\
0.217509059565445	2.875\\
0.166838052220113	1.25\\
0.1854276231018	1.8125\\
0.340203099698562	1.875\\
0.205748347666981	2.1875\\
0.242094967450059	2.5625\\
0.204127143218536	1.625\\
0.603551195597017	3.75\\
0.228323718348185	2.625\\
0.356032475103761	3.4375\\
0.386583770082941	3.0625\\
0.247154102873785	2.25\\
0.213106339209529	2.5\\
0.243226922277745	1.875\\
0.19954436717529	2.25\\
0.735714074706251	4.625\\
0.609024403365104	4.5625\\
0.66653069224713	4\\
0.25958183615349	2.0625\\
0.754893985678511	3.625\\
0.3955802226499	3.4375\\
0.736889624334389	4.5\\
0.766488662275445	4.5625\\
0.3137473471406	3.125\\
0.395209761790359	2.375\\
0.745222149048503	4.0625\\
-0.0650068212142906	1.75\\
0.418444138024289	2.9375\\
0.527739064240356	3.125\\
0.125598646933181	1.4375\\
0.0712761829223837	1.6875\\
0.63135316808316	4.125\\
0.156545943204431	2.3125\\
0.541496495302725	3.3125\\
0.328880031121605	2.75\\
0.248250474037148	2.5625\\
0.856057539726839	4.625\\
0.665163859291055	4.375\\
-0.0662902441916115	1.625\\
0.384480396796823	3.0625\\
0.456000705887932	3\\
0.207217669409959	1.6875\\
0.119202603102165	1.625\\
0.583367420104454	4\\
0.0968049669699738	2.3125\\
0.471386286115718	3.5625\\
0.305031836128544	2.5\\
0.218737125735038	2.625\\
0.698584718032301	4.5625\\
0.466073319592406	3.375\\
0.186335635264976	1.5625\\
0.154093382629735	1.75\\
0.370438501535493	2.0625\\
-0.0440057060561298	1.25\\
0.229821981334984	1.8125\\
0.395656890700023	1.8125\\
0.184892316859545	2.25\\
0.181811627056238	1.8125\\
0.256880296127372	2.5625\\
-0.0727324970270664	1.25\\
0.471320518141289	3.75\\
0.160202948866131	1.75\\
0.0375710151879287	2.5625\\
0.439103244217881	3.375\\
0.690497193836494	3.1875\\
0.325647910202063	1.875\\
0.0977205314392734	1.5\\
0.56520077227371	4.6875\\
0.168966322470857	3.5625\\
0.495937007058079	3.625\\
0.593504331657068	4.5\\
0.113502808078724	1.625\\
0.815384910830205	4.75\\
0.655707027419983	4.125\\
0.10748495002282	2.1875\\
0.0654217702394783	1.6875\\
0.393403155146402	2.875\\
0.189255201397004	2\\
0.132632806478423	2.0625\\
0.294913148373425	2.1875\\
0.227179282755271	3.8125\\
0.402984540977192	3.5\\
0.271778106431322	2.6875\\
0.122921071169871	2.125\\
0.597185032880211	5\\
0.316974452633951	3\\
0.142700792500021	1.4375\\
0.030250756486304	1.25\\
0.250543976551639	1.875\\
0.159887040695073	1.625\\
0.190407072153784	1.625\\
0.298903892976149	2.375\\
0.228464770504238	2.0625\\
0.310319161036077	1.625\\
0.241234962586939	2.0625\\
-0.0339135204537152	1.125\\
0.426555089583503	2.75\\
0.191224951652512	1.4375\\
};
\end{axis}
\end{tikzpicture}%
 & \hspace*{-0.2in}
%
%
\begin{tikzpicture}

\begin{axis}[%
width=0.95092\figurewidth,
height=\figureheight,
at={(0\figurewidth,0\figureheight)},
scale only axis,
every outer x axis line/.append style={black},
every x tick label/.append style={font=\color{black}},
xmin=-0.2,
xmax=1.2,
xlabel={Metric Score},
every outer y axis line/.append style={black},
every y tick label/.append style={font=\color{black}},
ymin=1,
ymax=5,
ylabel={MOS},
title={SROCC 0.7624 KROCC 0.5891 PLCC 0.8553},
axis x line*=bottom,
axis y line*=left,
title style={font=\scriptsize},xlabel style={font=\scriptsize},ylabel style={font=\scriptsize}
]
\addplot[only marks,mark=o,mark options={},mark size=1.5000pt,color=blue] plot table[row sep=crcr,]{%
0.025579867946817	1.3125\\
0.146522495984964	2.25\\
0.0442991926793085	1.6875\\
0.155405409824778	2.1875\\
0.0121721198752547	1.5\\
0.127551363089601	2.1875\\
0.104759388303727	1.9375\\
0.108082166623541	1.9375\\
0.0806341393478243	1.9375\\
0.0443925627919583	1.625\\
-0.0434567601735777	1.5\\
0.127621418174408	2.3125\\
1.00033105617607	4.9375\\
0.987325095237905	5\\
0.994921795027745	5\\
1.00507170529363	4.9375\\
0.998593582288323	5\\
0.994717492725298	5\\
1.00484050096248	5\\
1.00757859650075	5\\
1.0012320075507	5\\
1.0049430404337	5\\
1.0000737663856	5\\
0.999168409123068	5\\
0.118822268379697	1.4375\\
0.0417432083539945	1.5625\\
0.177029699344728	2.375\\
0.0447208474667475	1.3125\\
0.207100692950495	1.5\\
0.303751650993983	2.5625\\
0.221860273399437	2.0625\\
0.273452456125976	1.5\\
0.141670122008598	1.6875\\
0.0349389054705044	1.25\\
0.322743837753074	2.875\\
0.135953620943429	1.4375\\
0.103728411688638	2.5625\\
0.113668150066054	2\\
0.306218433678779	3\\
0.2236355761167	2.6875\\
0.178907328948544	1.6875\\
0.435005592085519	3.5625\\
0.206457875224684	3.9375\\
0.361069552700156	2.75\\
0.337322041662825	3.3125\\
0.136171288673045	2.0625\\
0.85456341993995	4.9375\\
0.317431548400729	3.0625\\
0.185307682707743	2.6875\\
0.28158757090047	3.1875\\
0.215987223912389	2.8125\\
0.244505018011497	1.9375\\
0.207891190783376	2.25\\
0.51325132219788	3.9375\\
0.148623524587818	4\\
0.713976446269571	4.625\\
0.365248321428764	2.75\\
0.147155955673447	2.8125\\
0.913382935793247	5\\
0.75010158855147	4.4375\\
0.0915377416921394	2.6875\\
0.266596107369688	2.625\\
0.331180965153837	2.75\\
0.160491655261602	1.9375\\
0.0989856003429548	1.1875\\
0.546993224802186	3.9375\\
0.229252284797604	3.25\\
0.24372158421873	2.25\\
0.186379956051255	2.5\\
0.166132508782346	2.625\\
0.790377577785281	4.3125\\
0.309234091408024	2.875\\
-0.0350533521207637	2.125\\
0.500536663489699	4.125\\
0.0639286086440363	3.25\\
0.658740359822356	4.125\\
0.0355284062874384	2.25\\
0.608050985337841	4.6875\\
0.41310844819095	3.5\\
0.644890223199089	4.5625\\
0.366939981470233	4.0625\\
0.0454620098444282	2.9375\\
-0.146891295226716	1.8125\\
0.573607521273467	4.25\\
-0.0498154491762876	1.375\\
-0.0293950419892456	1.4375\\
-0.0411353961418839	1.6875\\
0.0176807186534876	1.5\\
0.0905583390693699	1.4375\\
0.210006980191281	2.6875\\
0.161922984838325	2.125\\
0.185955245950908	2.4375\\
0.116463564065278	2\\
0.0479220395768544	1.625\\
0.235796274280559	2.3125\\
0.124799277572277	1.6875\\
-0.077510335596703	1.625\\
0.0410675178558575	1.875\\
0.0594231158751214	1.875\\
0.0042591152536086	1.8125\\
0.019626472791935	1.6875\\
0.199634812439921	2.375\\
0.163810929793482	3.3125\\
0.160247637244406	2.4375\\
0.128759197908552	2.3125\\
0.0712459991908234	2.1875\\
-0.0276261223505775	1.375\\
0.0769248277422531	1.8125\\
-0.0825686052791281	1.5625\\
0.110208969941406	2.25\\
0.0616244021403686	2\\
0.0912677849062714	1.875\\
0.0165983474032502	2\\
0.294571361657018	2.875\\
0.209312456662577	3.5\\
0.239444485678332	2.9375\\
0.179489842080942	2.6875\\
0.0836677239236198	2.5\\
-0.039667234425204	1.125\\
0.152747103092443	2.25\\
-0.019549683180515	2.375\\
0.287431286909857	1.75\\
0.42031615521757	1.75\\
0.000198429772321778	1.3125\\
0.16312701599277	1.3125\\
0.350638989176134	2.75\\
0.205673964067261	1.0625\\
0.0909871189674221	2.625\\
0.129226417835542	2.375\\
0.0882297735307115	2.125\\
0.433631122168945	2.9375\\
0.150064073694629	2.25\\
0.0366064399987986	2.8125\\
0.191541354999234	2\\
0.411935122485742	3.125\\
0.0932534464513825	1.5\\
0.232515050447562	1.5\\
0.441669805589902	4\\
0.0656928141189498	2.8125\\
0.434261768922543	3.5\\
0.42529825500716	4.4375\\
0.082895969935129	1.5625\\
0.971168712924142	4.9375\\
0.443056867628187	4\\
0.0243987206067786	2.1875\\
0.00753648230331137	2.125\\
0.0405943504908683	3\\
-0.0312469837976495	1.1875\\
-0.0262976274962166	1.8125\\
0.552668263445188	3.5625\\
-0.0662415452463837	2.6875\\
0.243884105626533	3.625\\
0.291443680025246	2.25\\
0.0811471053207806	2.1875\\
0.564278192541844	3.0625\\
-0.044290632451614	1.6875\\
0.00551636515963995	2.625\\
0.0196556186805637	2.875\\
0.100060436015654	2.1875\\
-0.030709814752096	2.125\\
0.0457939822035562	1.875\\
0.279075867494841	3.8125\\
0.106499926001046	2.875\\
0.203796450635266	2.875\\
0.151598048600026	2.9375\\
0.0257724104099627	2.125\\
0.156120733705557	2.25\\
0.250629794650286	3.4375\\
-0.0894347479790095	1.4375\\
-0.0458380134852842	1.9375\\
-0.0836709491173699	1.625\\
-0.00420481007049192	1.3125\\
0.131316926587356	1.3125\\
0.219890462437328	2.3125\\
0.107846238361681	1.4375\\
0.162506037200391	1.75\\
0.132473989690577	2\\
0.0592453689232735	1.3125\\
0.214055316087369	2.875\\
0.0957327430269746	1.25\\
-0.0776831773488399	1.8125\\
0.170281131265418	1.875\\
0.0155634061419924	2.1875\\
0.0205717236666034	2.5625\\
-0.0148906877507849	1.625\\
0.515412696101099	3.75\\
0.11506998219667	2.625\\
0.148110886658023	3.4375\\
0.263126945334234	3.0625\\
0.137479974669544	2.25\\
0.122106928564148	2.5\\
0.0396709294745478	1.875\\
-0.0562683611014883	2.25\\
0.682844323051832	4.625\\
0.377838256545073	4.5625\\
0.498866389892628	4\\
0.151401776763802	2.0625\\
0.675920404107157	3.625\\
0.285513288548155	3.4375\\
0.643506381392797	4.5\\
0.661261693723578	4.5625\\
0.17046871120308	3.125\\
0.261533195657757	2.375\\
0.598116164565337	4.0625\\
-0.0709304954946666	1.75\\
0.33049209073638	2.9375\\
0.341366752757778	3.125\\
0.108490180279542	1.4375\\
0.0941149715032565	1.6875\\
0.552431173048956	4.125\\
0.211505691568304	2.3125\\
0.491740318444949	3.3125\\
0.200457678357228	2.75\\
0.225363791528269	2.5625\\
0.878999255020989	4.625\\
0.557816595217358	4.375\\
-0.107046953118428	1.625\\
0.292447437341745	3.0625\\
0.264129871746487	3\\
0.165280079975145	1.6875\\
0.145020411951553	1.625\\
0.50479670982391	4\\
0.146330775225184	2.3125\\
0.416698504305098	3.5625\\
0.177979036732104	2.5\\
0.191722096313754	2.625\\
0.715902295235802	4.5625\\
0.352184415947894	3.375\\
0.111904722634997	1.5625\\
0.132203454861751	1.75\\
0.254335746234246	2.0625\\
-0.0788390652276283	1.25\\
0.264845472355716	1.8125\\
0.358771934548712	1.8125\\
0.224229719441625	2.25\\
0.177067412304936	1.8125\\
0.165001257251374	2.5625\\
-0.0633719948635046	1.25\\
0.496142325463489	3.75\\
0.128139389223614	1.75\\
-0.0169050418344018	2.5625\\
0.34378411161563	3.375\\
0.477097086706719	3.1875\\
0.22432793030035	1.875\\
0.108192374453433	1.5\\
0.455384512108842	4.6875\\
0.121534223274022	3.5625\\
0.429268186090386	3.625\\
0.446479701871115	4.5\\
0.0986917015925063	1.625\\
0.823923569883241	4.75\\
0.578714006041434	4.125\\
0.0232391067242073	2.1875\\
-0.0532445764166491	1.6875\\
0.179624047414667	2.875\\
0.139878604950427	2\\
0.121137987590543	2.0625\\
0.300342215836457	2.1875\\
0.203869407034401	3.8125\\
0.345331093858589	3.5\\
0.168721485868258	2.6875\\
0.100469838383931	2.125\\
0.61936688421677	5\\
0.180069646298882	3\\
0.0518935374356317	1.4375\\
-0.0352711105163689	1.25\\
0.105053917128516	1.875\\
0.0791480646328191	1.625\\
0.203164109999578	1.625\\
0.292410986065129	2.375\\
0.279852373465147	2.0625\\
0.267077093646899	1.625\\
0.138968449779501	2.0625\\
-0.0303968454743728	1.125\\
0.440489054833763	2.75\\
0.0983156232572563	1.4375\\
};
\end{axis}
\end{tikzpicture}%
		\\
		(m)\added[id=MG]{$MAE$~\cite{MAE}} & (n) Proposed $WNSS$  & (o) Proposed $sWNSS$ \\   
	\end{tabular} 
	\caption{Scatter plots for all metrics with correlation scores displayed for the VAQ database. Most of the existing metrics along with the widely used AUC based metrics like $sAUC$~\cite{borjieval}, $AUC_{Borji}$~\cite{borjieval} and $AUC_{Judd}$~\cite{Judd_2012} do not show high correlation with subjective scores. Among the non-shuffled metrics, existing metrics $CC$~\cite{borjieval} and $NSS$~\cite{borjieval} exhibit better correlation scores with subjective scores and the proposed metric $WNSS$ performs the best.  Amongst the existing shuffled metrics, the recently proposed $sNSS$~\cite{MilindSamTPAMI} metric performs well with the proposed  $sWNSS$ performing the best.}
	\label{fig:scatterplots}		
\end{figure*}
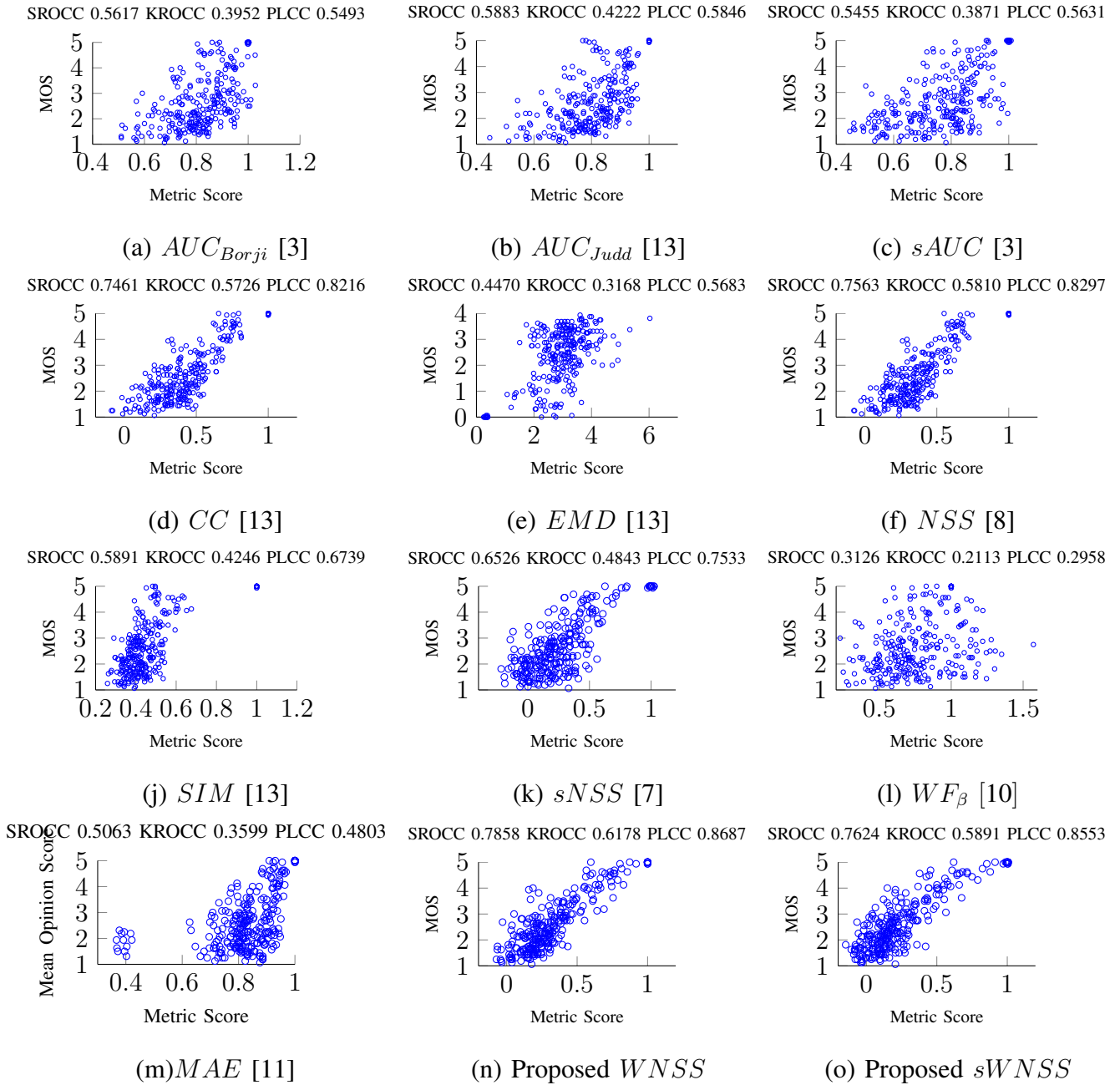

This section discusses the correlation results  between the subjective ratings and the metric scores. To evaluate how good a performance metric is, we compare the scores given by each metric to each of the considered models  with the average scores given by the subjects to the same models. To correlate the scores we use the widely used correlation measures of  Spearman Rank Order Correlation Coeffficient (SROCC), Kendall Rank Correlation Coefficient (KROCC) and Pearson Linear Correlation Coefficient (PLCC).
 \added[id=MG]{The SROCC and KROCC are rank correlation coefficients, and enable us to compare the ranking given to the VA models by a VA metric with the ranking given by the MOS scores. The PLCC is a linear correlation coefficient that measures how linear the relationship between the metric scores and MOS score is.} 
 The metric scores are normalized by the metric score obtained for the ground-truth saliency map that serves as an upper-bound on most metrics before performing the correlation.   For the $EMD$ metric for which a lower score is better and the best possible score is zero this normalization was not performed, and subjective scores were inverted by subtracting the scores from the maximum subject score of 5 to obtain positive correlation scores. For the $WF_{\beta}$ measure~\cite{margolin2014evaluate}, which requires the ground-truth to be a binary mask, we threshold the ground-truth saliency map by its standard deviation as suggested in~\cite{Moorthy_Bovik}. \added[id=MG]{For the $MAE$ metric, instead of using a binary ground-truth map as in~\cite{MAE}, we use a real-valued ground-truth saliency map since, by definition, the $MAE$ can be computed for two real-valued maps.} Table~\ref{tab:correlationresult} shows the result of the correlations for the  VAQ  database for all the existing metrics listed in Table~\ref{tab:existingmetrics} and our proposed metric (the shuffled $sWNSS$ and non-shuffled $WNSS$ versions). \added[id=MG]{The results for shuffled and non-shuffled metrics are reported separately because there is no explicit way to remove the center bias effect from the ground-truth saliency maps in the subjective study. As a result, human ratings will tend to be better for the saliency maps that boost central regions over peripheral regions. This leads to non-shuffled metrics like $NSS$ and $WNSS$ which tend to reward maps with more central than peripheral activity correlating better with human scores compared to their shuffled versions $sNSS$ and $sWNSS$.} Figure~\ref{fig:scatterplots} shows the scatter plots corresponding to the existing and proposed metrics. From the existing metrics used in the MIT saliency benchmark~\cite{mit-saliency-benchmark}, the $NSS$~\cite{NSS} and $CC$~\cite{borjieval} metrics perform significantly better than the other metrics with the $NSS$ performing the best among them. The $AUC_{Borji}$~\cite{borjieval} and its derivative $sAUC$~\cite{borjieval} metric that suffer from the most number of flaws perform the worst among the MIT benchmark metrics. The $WF_{\beta}$~\cite{margolin2014evaluate} metric also performs poorly.  The proposed  $WNSS$ metric gives the best correlation in the non-shuffled metrics and correspondingly the proposed $sWNSS$ metric gives the best performance among the shuffled metrics.


\section{Conclusion}
\label{sec:conclusion}
This paper proposes a locally weighted fixation-density based performance metric for assessing the quality of saliency predictions for VA models.  A subjective ground-truth Visual Attention Quality (VAQ) database is created to evaluate the performance of the proposed metric and other existing metrics.  Results of the evaluation show that the proposed metrics ($WNSS$ and its shuffled version $sWNSS$) outperform the widely used $sAUC$, $AUC_{Borji}$ and $AUC_{Judd}$ measures as well as other popular metrics used in the MIT Benchmark~\cite{mit-saliency-benchmark} in terms of their agreement with the subjective ratings. The subjective database is made available online to the research community as a performance metric evaluation benchmark on which future metrics can be tested.

%

%
%
%


\ifCLASSOPTIONcaptionsoff
  \newpage
\fi



\bibliographystyle{IEEEtran}
\bibliography{IEEEabrv,refs}

\newpage

\end{document}